\documentclass[final,5p,times,twocolumn]{elsarticle}
\usepackage{microtype}
\usepackage{amsmath,amssymb,amsfonts}
\usepackage{subcaption}
\usepackage{amsthm}

\usepackage{longtable}
\theoremstyle{remark}

\usepackage{graphicx}
\usepackage{hyperref}

\usepackage{algorithm}
\usepackage{algorithmic}
\usepackage{booktabs,makecell,multirow,tabularx}
\usepackage{xcolor}

\newcommand{\comment}[1]{\unskip\hfill{\scriptsize$\triangleright$~#1}}
\definecolor{planonecolor}{RGB}{180,30,30}
\definecolor{plantwocolor}{RGB}{20,120,60}
\definecolor{comparecolor}{RGB}{120,40,150}
\usepackage{array}
\newtheorem{definition}{Definition}

\begin{document}

\begin{frontmatter}
	
\title{EditSR: Enhancing Neural Symbolic Regression via Edit-based Rectification}

\author[label1]{Da Li}
\ead{dli@nenu.edu.cn}

\author[label5]{Xinxin Li}
\ead{51265500102@stu.ecnu.edu.cn}

\author[label3]{Xingyu Cui}
\ead{cuixingyu@szim.ac.cn}

\author[label3]{Jin Xu}
\ead{xujin22@gscaep.ac.cn}

\author[label2,label6]{Juan Zhang\corref{cor1}}
\ead{zhang\_juan@buaa.edu.cn}

\author[label1,label3,label4,label6]{Junping Yin\corref{cor1}}
\ead{yin\_junping@iapcm.ac.cn}

\cortext[cor1]{Corresponding authors: Junping Yin (yin\_junping@iapcm.ac.cn); Juan zhang (zhang\_juan@buaa.edu.cn) .}


\affiliation[label1]{organization={Academy for Advanced Interdisciplinary
		Studies, Northeast Normal University},
	city={Changchun},
	postcode={130024},
	state={Jilin},
	country={China}}
\affiliation[label2]{organization={Institute of Artificial
		Intelligence, Beihang University},
	city={Beijing},
	postcode={100191},
	state={Beijing},
	country={China}}

\affiliation[label3]{organization={Institute of Applied Physics and Computational Mathematics},
	city={Beijing},
	postcode={100094},
	state={Beijing},
	country={China}}
\affiliation[label4]{organization={National Key Laboratory of Computational Physics},
	city={Beijing},
	postcode={100088},
	state={Beijing},
	country={China}}
\affiliation[label5]{organization={School of Mathematical Sciences, East China Normal University},
	city={Shanghai},
	postcode={200241},
	state={Shanghai},
	country={China}}
\affiliation[label6]{organization={Shanghai Zhangjiang Institute of Mathematics},
	city={Shanghai},
	postcode={201203},
	state={Shanghai},
	country={China}}

\begin{abstract}
Neural symbolic regression models improve inference efficiency by shifting structural search to pretraining, but their one-pass autoregressive decoding is prone to error accumulation, which may lead to generating structurally incorrect expressions, especially in complex expression generation scenarios. Existing rectification strategies can alleviate this issue, but they often depend on restarting global search, thereby weakening the efficiency advantage of neural models, and remain susceptible to error accumulation. In this paper, we propose EditSR, a two-layer framework that combines a neural symbolic regression model in the first layer with an edit-based Rectifier in the second layer to achieve efficient prediction and post-hoc rectification. Instead of restarting the global search, we maintain rectification efficiency by pretraining the Rectifier. Specifically, we formulate the rectification process as a step-by-step state-transition chain starting from an incorrect expression, and develop a state-transition algorithm to construct supervised rectification chains for training the Rectifier. To ensure syntactic validity throughout rectification, each edit action is restricted to a syntactically valid space so that every edited expression remains parseable. In addition, because each edit decision is conditioned on the current state rather than the history, the Rectifier allows errors made in earlier steps to be rectified by subsequent edits, thereby reducing the risk of error accumulation. Extensive experiments and ablation studies show that EditSR substantially improves symbolic structure recovery with limited extra cost, with more pronounced gains on complex expressions, where one-pass autoregressive decoding is more susceptible to error accumulation.

\end{abstract}

	
\end{frontmatter}

\section{Introduction}
Symbolic regression aims to discover an explicit expression that maps inputs to outputs from sparse and noisy observational datasets. Unlike conventional regression, which optimizes parameters within a fixed function family, symbolic regression jointly searches over both the expression structure and the parameters, thereby offering greater degrees of freedom. The generated expressions are parseable, allowing researchers to examine symmetries and invariances, perform differentiation and integration, and directly deploy them in simulators and controllers. With these properties, symbolic regression is regarded as a powerful tool for scientific discovery~\citep{schmidt2009distilling, udrescu2020ai}.

Classical genetic programming approaches perform symbolic regression via population-based evolution, where the expressions are represented as trees and evolved through mutation and crossover~\citep{koza1994genetic,o2009riccardo}. Over the past decades, genetic programming-based models have been strengthened by advances in complexity control~\citep{Poli2008Parsimony, Poli2003Tarpeian}, evolution strategies~\citep{Luke2002LPP, Uy2011SemanticCrossover, Moraglio2012GSGP}, and numerical optimizations~\citep{Kommenda2013NLS,Burlacu2020Operon}, which help genetic programming remain a competitive paradigm on modern benchmarks~\citep{la2021contemporary, de2024srbench++}. Nevertheless, genetic programming-based models have long faced efficiency challenges because the search space is highly flexible and grows rapidly with expression depth and the size of the operator set. 

Deep learning represents a new paradigm in symbolic regression~\citep{lample2019deep}. A representative deep learning-based model is Equation Learner (EQL), which replaces activation functions with primitive operators so that the resulting network admits a symbolic form, enabling interpretability and gradient-based training~\citep{martius2016extrapolation}. However, such models are often constrained by a predetermined depth, which limits their flexibility. Another line of deep learning-based methods leverages reinforcement learning, where the constants, operators, and variables are treated as tokens, with policy gradient updates guiding the expression search~\citep{petersen2019deep, mundhenk2021symbolic}. Overall, by introducing optimizable agents, deep learning-based models can improve inference efficiency and reduce sensitivity to hyperparameters. Moreover, since the optimization process can naturally incorporate structural constraints, the expressions they generate are often simpler. However, these models still need to search from scratch and cannot accumulate knowledge from historical experience. Therefore, a trade-off between efficiency and performance remains.

Another line of research introduces large-scale pretraining into symbolic regression, drawing inspiration from machine translation. In this paradigm, a Transformer-based architecture is trained on massive corpora of dataset--expression pairs~\citep{lample2019deep, biggio2021neural}, so that symbolic regression can be formulated as a set-to-sequence task that maps datasets directly to expressions~\citep{valipour2021symbolicgpt}. As a result, much of the expensive exploration over symbolic structures is absorbed into pretraining, allowing inference to be reduced to a single forward pass~\citep{kamienny2022end, vastl2024symformer, landajuela2022unified}. Furthermore, by introducing noise during pretraining, these models also demonstrate strong zero-shot generalization to unseen expression forms and robustness to noisy datasets. 

In summary, the development of symbolic regression reflects sustained efforts to balance search capability with computational efficiency. Neural models therefore have the potential to offer a practical balance between search capability and computational efficiency. However, their one-pass autoregressive decoding paradigm is often limited by error accumulation, thus hindering their ability to generate complex expressions. Although post-hoc rectification strategies can alleviate this issue, they often incur substantially higher computational costs, potentially offsetting the speed advantage originally offered by pretraining~\citep{shojaee2023transformer, meidani2023snip}. Therefore, we investigate whether a lightweight post-hoc rectification module can rectify incorrect predictions from neural symbolic regression models without sacrificing efficiency.

In recent years, edit-based generative models have developed rapidly across multiple domains. In natural language generation, models such as Levenshtein Transformer~\citep{gu2019levenshtein} and EDITOR~\citep{xu2021editor} replace one-pass autoregressive decoding with iterative insertions, deletions, and local rewrites, demonstrating that an imperfect draft can be efficiently improved without full autoregressive regeneration. Based on the concept of discrete diffusion, subsequent studies further model the generation process itself as a sequence of multi-step edits~\citep{reid2022learning, reid2022diffuser, havasi2025edit}. A key advantage of these models is that they allow errors to be progressively rectified, analogous to iterative refinement in evolutionary search. This property is particularly attractive for symbolic regression, because many expression errors are local, while substantial subexpressions are often reusable and need not be re-explored from scratch. In addition, edit-based models support pretraining, which is important for ensuring rectification efficiency.

To achieve efficient prediction and rectification, we propose \textbf{EditSR}, which consists of two layers. The first layer is a neural symbolic regression model. We instantiate it as NeSymReS~\citep{biggio2021neural}, a representative architecture. Like other one-pass autoregressive models, NeSymReS cannot recover once an early error distorts the downstream syntactic context. Therefore, we introduce a Rectifier in the second layer that performs multi-step edit-based rectification on incorrect predictions. The Rectifier is composed of a Tagger and Editor, where the Tagger determines the edit position and the edit action, while the Editor predicts the corresponding edit content. We define the edit actions \textsc{Keep}, \textsc{Replace}, \textsc{Delete}, \textsc{Rewrite}, and \textsc{Insert}, which are executed at the subtree level rather than the token level, so that the expression remains syntactically valid after editing.

The rectification process is organized as a state-transition chain, where each state corresponds to a parseable expression and each state transition is realized by executing a single edit action. We construct a set of expressions with diverse error patterns as initial states by applying random corruptions to the target expression, inspired by the widely explored discrete diffusion models~\citep{austin2021structured, reid2022diffuser, Liu2023SNR}, and then use a state-transition algorithm to map them back to the target, thereby obtaining supervised rectification chains. The Rectifier is trained on these one-step transitions, exposing it to a wide variety of error patterns. During inference, each decision is made by conditioning on the current state rather than the history, which helps mitigate error accumulation and leaves room for later edits to repair earlier errors. 


The main contributions of our work are summarized as follows:
\begin{itemize}
	\item We revisit neural symbolic regression models from the perspective of error accumulation and examine whether incorrect predictions can be rectified through a few edits, rather than restarting the global search.
	
	\item We propose EditSR, a two-layer framework that combines a neural symbolic regression model in the first layer with a Rectifier in the second layer. EditSR provides a novel prediction-and-rectification mechanism, which preserves the inference efficiency of neural models while rectifying structurally incorrect expressions via syntax-constrained post-hoc edits, without reopening the global search process.

	\item We represent rectification as an iterative state-transition process where each state corresponds to a parseable expression, and each state transition is realized by executing a single edit action. In our setting, each rectification decision by the Rectifier is conditioned on the current state rather than the history, thereby suppressing error accumulation while allowing errors made in earlier steps to be rectified by subsequent edits.

	\item Through extensive evaluations on mainstream benchmarks and ablation studies, we show that the Rectifier is beneficial for long-expression generation, where one-pass autoregressive decoding is more prone to error accumulation. In addition, the Rectifier improves symbolic structure recovery, as its rectification objective is always directed toward the target expression, rather than merely satisfying a numerical error threshold.
\end{itemize}

The remainder of the paper is organized as follows. Section~\ref{motivation} introduces the motivation behind the proposed framework, and Section~\ref{related_work} reviews the related work. Section~\ref{method} presents the EditSR architecture, and Section~\ref{data_generation} describes the data-generation procedure. The empirical evaluation is reported in Sections~\ref{experiment_design} and~\ref{Results}. Finally, Section~\ref{sec:conclusion} presents the conclusion of our work.

\section{Motivation}
\label{motivation}

In recent years, neural symbolic regression models have become attractive because they amortize much of the structural search during pretraining, reducing inference to a single forward pass. However, their drawback lies in the mismatch between teacher-forcing training and free-running inference. Early errors then propagate through subsequent autoregressive decoding steps and cannot be self-rectified. In symbolic regression, this problem is especially costly because the generated expression represents a tree under strict arity constraints. Once an early operator or subtree is wrong, the subsequent content is no longer generated in the intended syntactic context. Therefore, the dominant error pattern is often structural rather than lexical. 

\begin{figure}[h]
	\centering
	\includegraphics[width=0.49\textwidth]{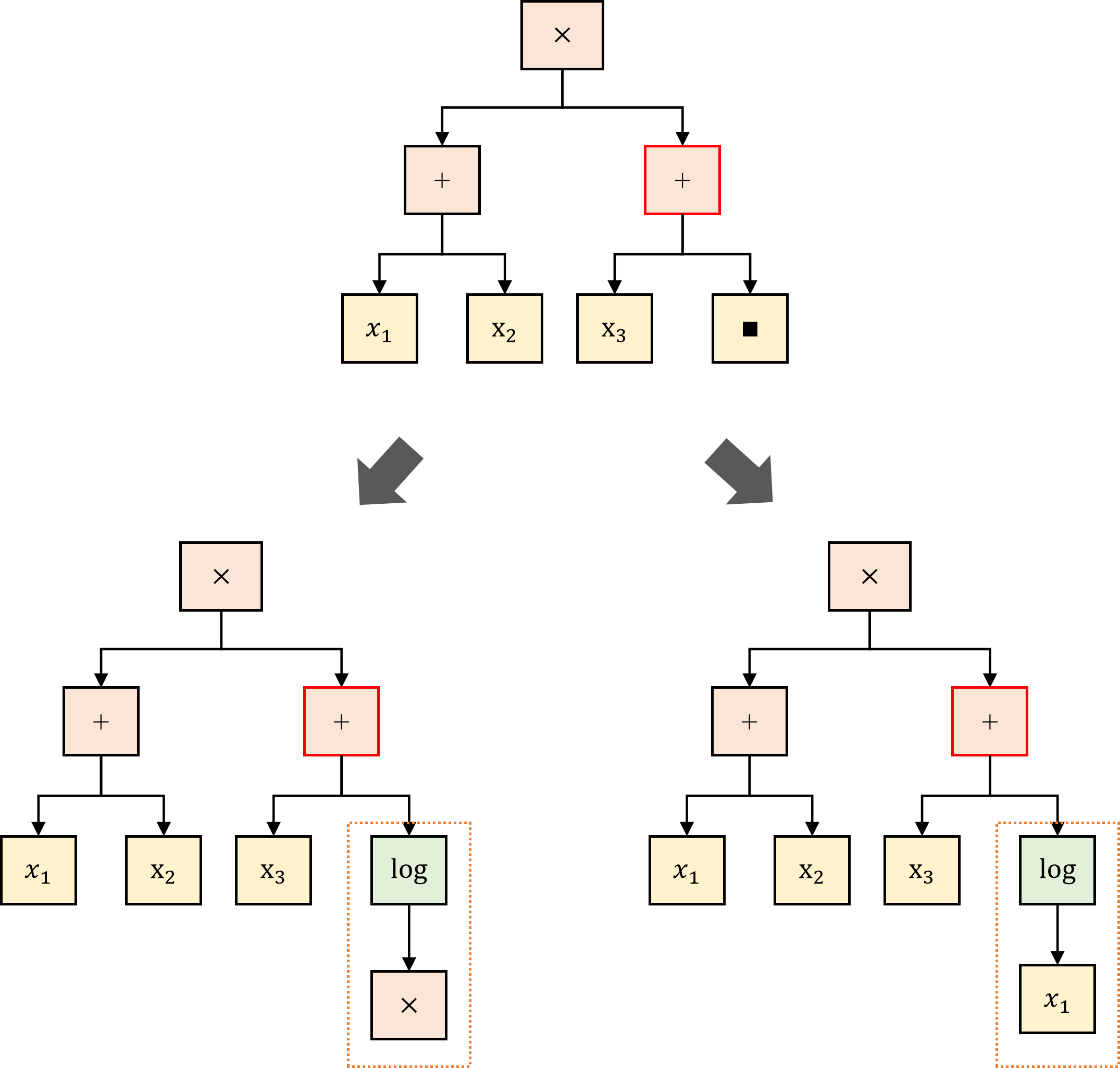}
	\caption{\textbf{Under autoregressive decoding, two possible outcomes after generating an incorrect token.} If the target expression is $(x_1+x_2)\cdot \sin(x_3)$, then once the unary operator $\sin$ is incorrectly generated as the binary operator $+$ (highlighted in red), the model either leaves the tree structurally incorrect (dashed box on the left) or is forced to close the wrong branch with an irrelevant subtree (dashed box on the right).}
	\label{fig:example}
\end{figure}

This difference can be understood by contrasting symbolic regression with natural language generation. In the latter, an early deviation from a reference sentence does not necessarily invalidate the remainder of the output, because multiple continuations may still preserve the intended meaning. For example, the sentence ``Energy cannot be created or destroyed; it can only be transformed from one form to another'' may also be expressed as ``Energy is not created or destroyed; it can only be transformed between different forms.'' Although the wording deviates from the outset, the subsequent text remains semantically correct. By contrast, in symbolic regression, an early error can alter the required tree expansion. As shown in Fig.~\ref{fig:example}, if the target expression is $(x_1+x_2)\cdot \sin(x_3)$ but the unary operator $\sin$ is incorrectly generated as a binary operator such as $+$, then the following tokens are no longer interpreted under the intended arity pattern and syntactic context. As a result, what begins as a local mistake can quickly cascade into a global structural deviation. 

Existing post-hoc rectification strategies have demonstrated that such errors can be mitigated. However, many of them rely on restarting global search, or performing latent-space optimization during inference~\citep{shojaee2023transformer, Liu2023SNR, meidani2023snip}. Therefore, the rectification process often introduces substantial additional computational overhead, which weakens the efficiency advantage that neural models originally gained through pretraining.

Our framework is motivated by a simple observation. Even when a neural model fails to recover the exact target expression, it often already captures a reusable structure (as seen in Fig.~\ref{fig:example}), such as correct variables, constants, or subtrees. This phenomenon becomes more common on complex expressions, where the mismatch is often concentrated in a limited region rather than the whole expression. Therefore, always restarting the search process from scratch can be inefficient. A more efficient strategy is to preserve the structurally correct part and rectify only the incorrect part. Recent edit-based generative models support this view. They show that an imperfect prediction can often be improved through a sequence of local insertion, deletion, and replacement operations rather than full regeneration~\citep{karras2022elucidating, austin2021structured, chang2022maskgit, reid2022diffuser}. However, in symbolic regression, the extra cost of iterative generation still makes its role relative to autoregressive prediction unclear~\citep{tymkow2025symbolic, bastiani2025diffusion}. Therefore, we retain the fast inference capability of autoregressive models and focus on a narrower question: how to perform iterative rectification once an incorrect prediction has already been generated. This motivates us to introduce the Rectifier, which performs error rectification via edit-based transitions, and maintains efficiency through pretraining. Moreover, because each rectification decision is conditioned only on the current state, the Rectifier is less vulnerable to error accumulation and can still rectify earlier errors through subsequent edits. However, applying edit-based generation to symbolic regression is challenging. Unlike natural language, expressions are governed by strict arity constraints, and the rectification objective is directed toward a much less ambiguous target rather than a diverse set of semantically acceptable sentences. Such challenges prompt us to restrict edits to a strictly syntax-constrained space and to construct a deterministic rectification chain from an incorrect expression toward the target one, rather than relying on stochastic forward corruption or reverse-path reconstruction as in diffusion-style edit models.

\section{Related work}
\label{related_work}
\textbf{Iterative generation and edit-based models.} Iterative generation has been extensively studied in recent years, spanning denoising-based generation, masked prediction, and edit-based refinement~\citep{ho2020denoising, song2020score, hoogeboom2021argmax, austin2021structured, chang2022maskgit}. A common principle underlying these methods is that an initial imperfect output can be progressively improved through a sequence of intermediate states, which has proved effective in a variety of domains. For example, in natural language generation, DiffusER factorizes the edit process into several coordinated subproblems, including edit position prediction, edit action prediction, and edit content generation, thereby enabling iterative rectification of incorrect predictions even in full noise~\citep{reid2022diffuser}. Edit Flows similarly formulates generation as a sequence of edit actions, while introducing auxiliary alignments to describe how one intermediate sequence is transformed into the next~\citep{havasi2025edit}. These studies inspire our work because they show how rectification trajectories can be prelearned and executed iteratively during inference, thereby offering an alternative to traditional online search-based rectification strategies in symbolic regression.

\textbf{Genetic programming-based symbolic regression models.} Genetic programming remains the dominant paradigm for symbolic regression, where expressions are represented as trees and evolved through mutation, crossover, and selection~\citep{koza1994genetic, o2009riccardo}. In recent years, several works have addressed more specific bottlenecks. Some replace unconstrained evolution with more structured search spaces to obtain simpler expressions and more controllable runtime~\citep{mcconaghy2011ffx, de2021interaction}. Others incorporate semantic guidance, such as MRGP~\citep{arnaldo2014multiple} and GP-GOMEA~\citep{virgolin2021improving}, to bias evolution toward more meaningful updates. For engineering applications, widely used models such as Eureqa~\citep{schmidt2009distilling}, Operon~\citep{Burlacu2020Operon}, and PySR~\citep{cranmer2024pysr} combine genetic programming with algebraic simplification, constant optimization, and efficient implementations to improve performance and efficiency. AI Feynman adopts a physics-inspired strategy, recursively reducing the effective search space by exploiting structural properties such as separability, symmetry, and invariance, and then invoking symbolic regression on the resulting subproblems~\citep{udrescu2020ai}. Overall, by combining evolutionary search with gradient-based parameter optimization, broad priors, and complexity control, modern genetic programming-based models remain competitive on mainstream benchmarks~\citep{la2021contemporary, de2024srbench++}.

\textbf{Deep learning-based symbolic regression models.} Deep learning has become an important tool in symbolic regression because it inherently possesses powerful numerical fitting capabilities and can guide the search by introducing intelligent agents. Equation Learner (EQL) replaces activation functions with primitive operators and encourages sparsity, yielding interpretable expressions~\citep{martius2016extrapolation}. EQL$^{\div}$ further improves the training stability of EQL and extends the application scope to control-oriented problems~\citep{sahoo2018learning}. However, these architectures are often less flexible because the depth and operator layout must be preset. Reinforcement learning-based models are able to explore more flexible symbolic structures, which treats expressions as sequences sampled from a policy. Deep Symbolic Regression (DSR) uses a Recurrent Neural Network with risk-seeking policy gradients and syntax-constrained sampling, making the generated expressions simpler~\citep{petersen2019deep}. Sym-Q~\citep{tian2025interactive} emphasizes expert-guided symbolic regression by introducing a co-design mechanism that allows domain experts to intervene during the equation discovery process and inject prior knowledge into the evolving expression. Recently, symbolic regression models based on Monte Carlo Tree Search (MCTS) have improved search efficiency through complementary mechanisms, including GPT-guided search policies in SR-GPT~\citep{li2025discovering}, graph-based equivalence compression and constraint encoding in GSR~\citep{xiang2025graph}, and non-local state-jumping actions such as mutation and crossover in improved MCTS variants~\citep{huang2025improving}. Beyond improving individual search paradigms, recent studies have increasingly explored hybrid frameworks. Representative models include uDSR~\citep{landajuela2022unified}, Neural-Guided Genetic Programming~\citep{mundhenk2021symbolic}, and Parallel Symbolic Enumeration (PSE)~\citep{ruan2026discovering}, which combine deep learning, genetic programming, and parallel optimization to improve robustness and generalization across diverse application scenarios.

Overall, deep learning-based symbolic regression has evolved into increasingly flexible search-guided frameworks, where neural policies, expert knowledge, MCTS, and hybrid optimization are combined to balance interpretability, search efficiency, and generalization. However, similar to genetic programming, most of these models still perform problem-specific searches from scratch, making it difficult to accumulate historical search experience and reuse prior knowledge to further improve efficiency and performance across scenarios.

\textbf{Neural symbolic regression models.} The Transformer architecture has recently been introduced into symbolic regression to address the challenge of fast inference. NeSymReS~\citep{biggio2021neural} and SymbolicGPT~\citep{valipour2021symbolicgpt} are representative models. After pretraining, they can directly generate expression skeletons from datasets, but they usually require additional constant optimization. E2E~\citep{kamienny2022end} and SymFormer~\citep{vastl2024symformer} further promote the direct mapping from datasets to parseable expressions. Multimodal approaches, such as SNIP~\citep{meidani2023snip} and MMSR~\citep{Li2024MMSR}, represent datasets and symbols as different modalities and perform modality alignment and fusion through contrastive learning to alleviate training instability caused by the lack of token-level correspondence between datasets and expressions. ViSymRe~\citep{li2024visymre} introduces the visual modality as an additional information source, thus enhancing the convergence and generalization capability. Recent efforts have incorporated large language models into symbolic regression by using them to propose equation-program skeletons informed by scientific priors, such as LLM-SR~\citep{shojaee2024llm}, or to induce reusable natural language concept libraries that guide evolutionary search, such as LaSR~\citep{grayeli2024symbolic}, thereby enriching symbolic regression with knowledge beyond data alone. However, such models may require dedicated evaluation benchmarks, since conventional datasets can be affected by memorization in large language models~\citep{shojaee2025llm}. Several recent works refine either the search objective or the search space itself. SR4MDL~\citep{yu2024symbolic} uses a learned Minimum Description Length (MDL) surrogate to reduce the mismatch between numerical fitting and symbolic structure recovery. ParFam~\citep{scholl2025parfam} translates discrete symbolic regression into continuous global optimization over structured parametric families, while its transformer-guided variant further reduces search cost. CaMo~\citep{liu2025camo} emphasizes modular reuse of high-value substructures in end-to-end symbolic regression. 

These models confirm that modern neural symbolic regression increasingly benefits from richer priors and tighter constraints. Although they differ in architecture, many share a common inference pattern, i.e., directly predicting an expression from datasets in a one-pass autoregressive manner. Therefore, provided that the interface matches, such models can naturally serve as the first layer of our framework.

\textbf{Post-hoc rectification strategies.} Post-hoc rectification has been widely studied in generative models. One representative approach is gradient-based optimization by perturbing hidden states or by optimizing continuous relaxations during decoding to steer a pretrained generator toward desired outputs, such as PPLM~\citep{dathathri2019plug} and COLD~\citep{qin2022cold}. Another approach introduces search-based refinement by explicitly exploring alternative continuations during inference, for example, with MCTS~\citep{kumagai2016human}. Related ideas have also been explored in symbolic regression. TPSR incorporates MCTS into Transformer decoding, enabling non-differentiable feedback, such as fitting error and Complexity, to guide the refinement of candidate expressions~\citep{shojaee2023transformer}. SNIP improves symbolic regression through latent space optimization~\citep{meidani2023snip}. SNR introduces a rectifiable learning framework that combines learned structural priors with reinforcement learning to iteratively adjust incorrect expression structures~\citep{Liu2023SNR}. However, compared with rectification strategies in other fields, the efficiency cost of TPSR, SNIP, and SNR is often more evident, as each refinement round may involve repeated numerical evaluations and constant optimization of candidate expressions.

\section{Methods}
\label{method}

\begin{figure*}[t]
	\centering
	\includegraphics[width=\textwidth]{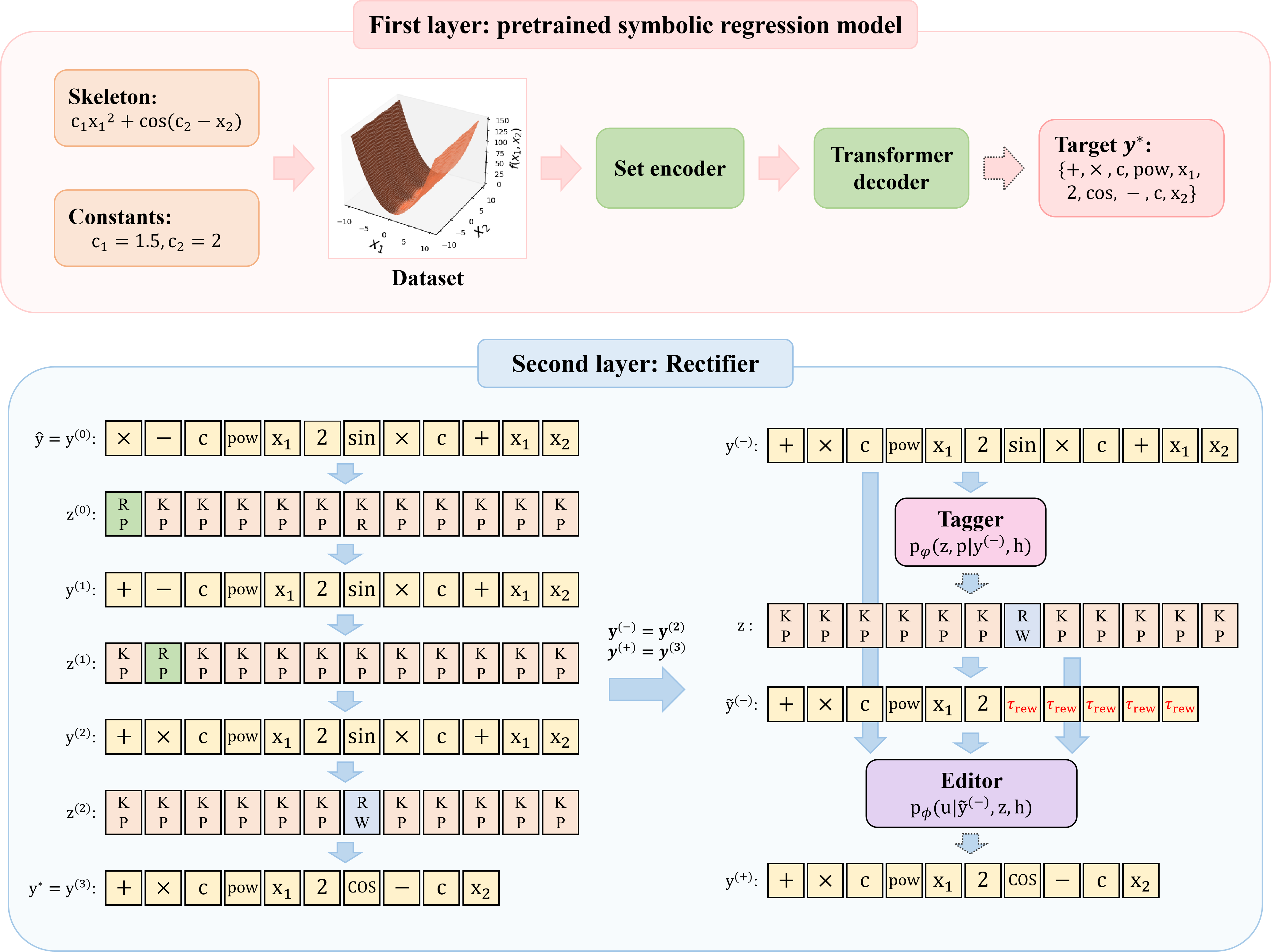}
	\caption{\textbf{Training overview of EditSR.} At the top, a neural symbolic regression model in the first layer is trained to map datasets directly to expressions, using the target expression $f^*$ for supervision. At the bottom, the Rectifier learns a rectification chain from $f^{(0)}$ to $f^*$. Here, $z^{(t)}$ denotes the edit action at step $t$, where KP, RP, and RW are abbreviations for the edit actions \textsc{Keep}, \textsc{Replace}, and \textsc{Rewrite}, respectively. $h$ denotes the dataset encoding, and $p$ denotes the position associated with the highest-confidence edit action. During training, $f^{(0)}$ is obtained by applying random corruptions to $f^*$, and a deterministic rectification chain from $f^{(0)}$ to $f^*$ is constructed using a state-transition algorithm. The Rectifier is trained on the one-step state transition $f^{(t)} \rightarrow f^{(t+1)}$. The dashed arrows in the figure indicate the supervision targets.}
	\label{fig:diffsr_train}
\end{figure*}

In this section, we introduce the architecture of EditSR as well as its optimization and inference procedures. Fig.~\ref{fig:diffsr_train} provides an overview. In EditSR, NeSymReS serves as the fast neural model in the first layer and is extended to support 10 variables, while the second layer consists of the Rectifier. We define $f^{(0)}$ as the initial state, i.e., an incorrect expression that requires rectification. Each edit action executed by the Rectifier induces a state transition. We denote by $f^{(t)}$ the state reached after $t$ rectification steps starting from $f^{(0)}$. A complete rectification process therefore forms a rectification chain
\[
f^{(0)} \rightarrow f^{(1)} \rightarrow \cdots \rightarrow f^{(T)} = f^*.
\]
For each edit, we apply strict syntactic constraints to avoid unclosed subtrees, thereby ensuring that the rectification chain always remains in the space of parseable expressions. 

In the following, we first introduce the syntax-constrained decoding rule shared by both layers, then describe the first-layer neural symbolic regression model and the second-layer Rectifier, and finally present the corresponding optimization and inference procedures. Table~\ref{tab:symbols} summarizes the main symbols used throughout this paper.

\begin{table*}[h]
	\centering
	\caption{Main symbols used in the paper.}
	\label{tab:symbols}
	\renewcommand{\arraystretch}{1.3}
	\begin{tabularx}{\textwidth}{>{\raggedright\arraybackslash}l | >{\raggedright\arraybackslash}X}
		\toprule
		\textbf{Symbol} & \textbf{Description} \\
		\midrule
		$\mathcal{D}=\{(x_n,y_n)\}_{n=1}^{N}$ & Dataset containing $N$ sampled input-output pairs. \\
		$h$ & Dataset encoding produced by the Set encoder. \\
		$f=(f_1,\ldots,f_L)$ & Prefix expression sequence. For convenience, we uniformly call it ``expression'' in the paper. \\
		$f^*$ & Target expression sequence. \\
		$\hat{f}$ & Expression predicted by NeSymReS. \\
		$f^{(t)}$ & Expression after $t$ rectification steps. \\
		$\tilde f^{(t)}$ & Editor context state at step $t$. \\
		$\mathcal{V}$ & Vocabulary of tokens, including constant placeholders, operators, and variables. \\
		$a(v)$ & Arity of token $v$, $v \in \mathcal{V}$. \\
		$L_{\max}$ & Maximum expression sequence length. \\
		$\mathcal{Z}$ & Edit action set, including $\{\textsc{Keep},\textsc{Replace},\textsc{Delete},\textsc{Rewrite},\textsc{Insert}\}$. \\
		$p,\,z,\,u$ & Edit position, edit action, edit content. \\
		$T$ & Number of rectification steps for the supervised rectification chain. \\
		$T'$ & Number of corruption steps applied on the target expression to obtain a rectification initial state. \\
		$T_{\max}$ & Maximum number of rectification steps during inference. \\
		$S$ & Placeholder budget used by the Editor. \\
		$p_\theta,\,p_{\psi},\,p_{\phi}$ & Conditional distribution of NeSymReS, the Tagger, and the Editor. \\
		$\tau_{\mathrm{rep}},\,\tau_{\mathrm{ins}},\,\tau_{\mathrm{del}},\,\tau_{\mathrm{rew}}$ & Action-specific placeholders. \\
		$\mathcal{T}$ & One-step transition function that maps the current state to the next state. \\
		$\mathcal{L}_{\mathrm{Tagger}},\,\mathcal{L}_{\mathrm{Editor}},\, \mathcal{L}_{\mathrm{Rectifier}},\,\mathcal{L}_{\mathrm{base}}$ & Tagger loss, Editor loss, Rectifier loss, and the first-layer loss. \\
		\bottomrule
	\end{tabularx}
\end{table*}

\subsection{Syntax-constrained decoding}

Neural symbolic regression models represent expressions as prefix token sequences and model them using left-to-right conditional distributions. We denote a prefix expression sequence by $f=(f_1,\ldots,f_L)$ over the vocabulary $\mathcal{V}$, which contains operator symbols, variables, and a constant placeholder token. Each $v\in\mathcal{V}$ is assigned an arity $a(v)\in\{0,1,2\}$, where $a(v)=0$ for variables or constants, $a(v)=1$ for unary operators, and $a(v)=2$ for binary operators.

Syntactic constraints are characterized by the deficit count. Starting from the root, each token consumes one pending hole and creates $a(f_i)$ new holes. We define the deficit after reading the first $i$ tokens as
\begin{equation}
	d_0=1,\qquad d_i = d_{i-1}-1 + a(f_i)\quad \text{for } i=1,\ldots,L.
	\label{eq:deficit}
\end{equation}
The sequence is syntactically valid if and only if the deficit never drops below 1 before the end and all holes are exactly filled at termination. Intuitively, $d_i$ is the number of child nodes that remain to be filled after consuming token $i$.

Based on this, we impose syntax constraints during autoregressive decoding. Let the current deficit be $d>0$ before emitting token $f_i$, and let the remaining token budget be $R=L_{\max}-i+1$. If token $v\in\mathcal{V}$ is emitted, the updated deficit is
\[
d' = d-1+a(v).
\]
To keep decoding feasible, the new deficit must admit a valid completion within the remaining $R-1$ steps. Since each future token can reduce the deficit by at most 1, we require $d' \le R-1$. Unlike fixed-length decoding, we allow the expression to terminate before exhausting the maximum length budget. Therefore, a token is feasible if the updated deficit satisfies $0 \le d' \le R-1$. If $d'=0$, the generated prefix forms a complete expression and decoding terminates immediately; otherwise, decoding continues. The feasible token set at state $(d, R)$ is therefore defined as
\begin{equation}
	\mathcal{V}'(d,R)=
	\{v\in\mathcal{V}: 0 \le d-1+a(v) \le R-1\}.
	\label{eq:feasible_set}
\end{equation}
We adopt the same syntax-constrained feasible-set restriction in both the first layer for expression prediction and the second layer for edit content generation.

\subsection{The first layer: neural symbolic regression model}

Architecturally, a standard neural symbolic regression model is divided into two modules. A permutation-invariant Set encoder (such as the standard Transformer encoder~\citep{vaswani2017attention}, Set Transformer~\citep{lee2019set}, T-net variants~\citep{qi2017pointnet}, or Mix scheme~\citep{lalande2023transformer}) maps a dataset to the dataset encoding $h$. A decoder parameterized by $\theta$ defines the factorized conditional distribution
\begin{equation}
	p_\theta(f\mid h)=\prod_{i=1}^{L} p_\theta(f_i \mid f_{<i},h),
	\label{eq:ar_factorization}
\end{equation}
where $f_{<i}$ is the generated prefix before step $i$. Generally, the decoder is a standard Transformer decoder.

Given the target expression sequence $f^*$, the training objective is
\begin{equation}
	\mathcal{L}_{\mathrm{base}}= -\sum_{i=1}^{L}\log p_\theta \left(f^*_i \mid f^*_{<i},h\right).
	\label{eq:lar_ar}
\end{equation}

During inference, the model performs left-to-right syntax-constrained decoding. The next token is decoded as
\begin{equation}
	\hat{f}_i=\arg\max_{v\in\mathcal{V}'(\hat d_{i-1},R_i)} p_\theta \left(v \mid \hat{f}_{<i},h\right).
\end{equation}
Beam search maintains multiple high-scoring expressions to better balance search quality and computational cost.

\subsection{The second layer: Rectifier}

In this section, we present the Rectifier in detail, including the roles of the Tagger and Editor, the corresponding training target, and the inference procedure.

\subsubsection{Tagger}
\label{tagger}
At step $t$, the Tagger receives the current state $f^{(t)}$ and the dataset encoding $h$. Its role is to evaluate, at every position, which edit actions are admissible and then choose one executable position--action pair for the Editor. 

We define the following action space:
{\small
	\begin{equation}
		\mathcal{Z}=\{\textsc{Keep},\,\textsc{Replace},\,\textsc{Delete},\,\textsc{Rewrite},\,\textsc{Insert}\},
		\label{eq:edit_set}
	\end{equation}
}
Each action corresponds to a specific behavior:
\begin{itemize}
	\item \textsc{Keep}: leaves the token at the selected position unchanged.
	\item \textsc{Replace}: replaces only the token at the selected position while preserving its arity.
	\item \textsc{Delete}: collapses the internal subtree rooted at the selected position into a leaf.
	\item \textsc{Rewrite}: regenerates the internal subtree rooted at the selected position.
	\item \textsc{Insert}: expands the leaf at the selected position into a newly generated internal subtree.
\end{itemize}

\begin{figure*}[t]
	\centering
	\includegraphics[width=\textwidth]{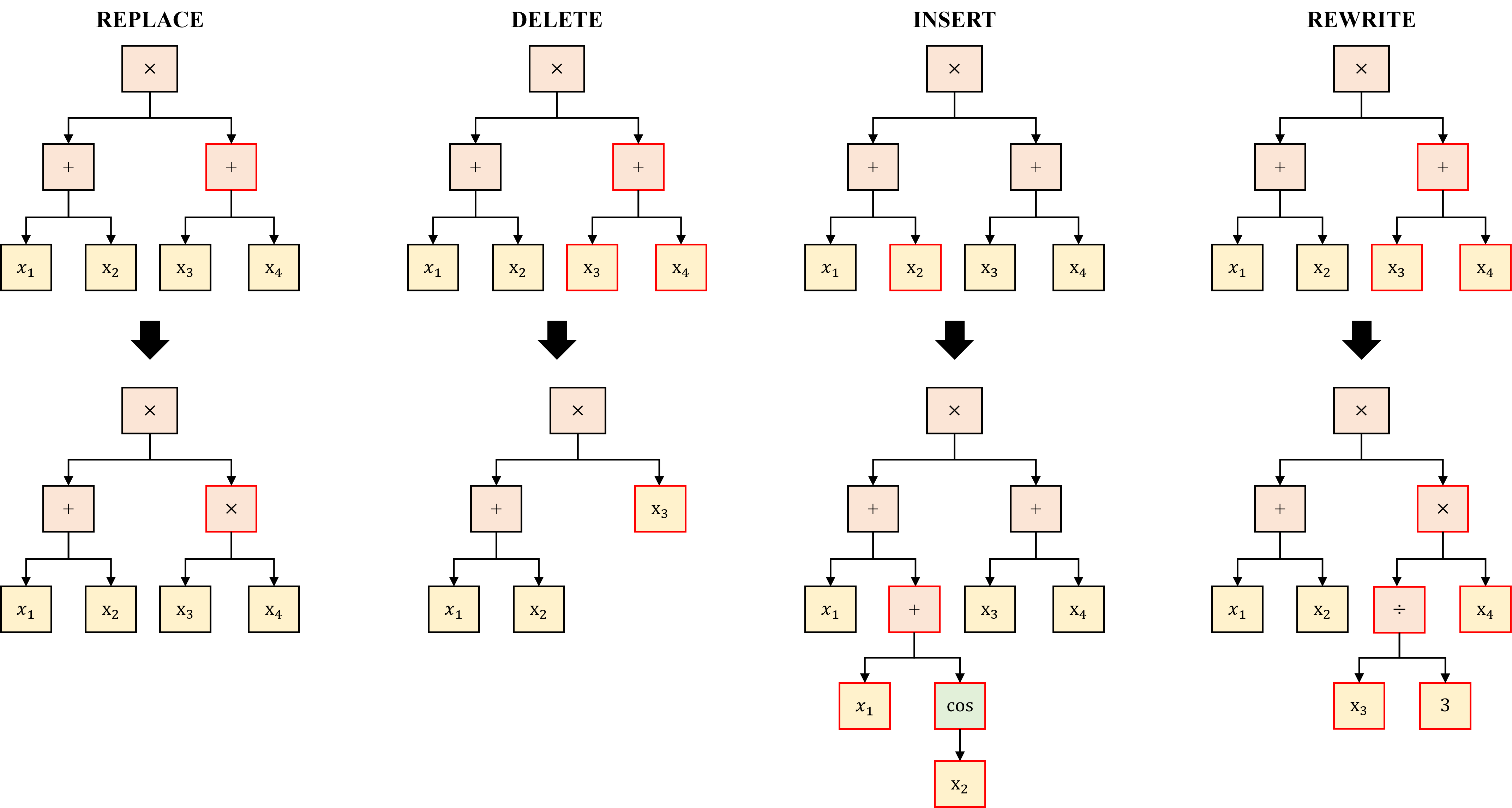}
	\caption{Illustration of the four non-trivial edit actions used by the Rectifier. From left to right, the columns show examples of \textsc{Replace}, \textsc{Delete}, \textsc{Insert}, and \textsc{Rewrite}. In each column, the top row indicates the selected edit region in the current expression (highlighted in red), and the bottom row shows the resulting expression after applying the corresponding action. \textsc{Replace} modifies only the token at the selected position while preserving its arity; \textsc{Delete} collapses the selected internal subtree into a leaf; \textsc{Insert} expands a selected leaf into a newly generated subtree; and \textsc{Rewrite} regenerates the selected subtree. The \textsc{Keep} action is not shown because it leaves the expression unchanged.}
	\label{fig:edit_actions}
\end{figure*}

Not every action is admissible at every position. Let $\mathcal{Z}^{(t)}_i$ denote the admissible action set at position $i$ of $f^{(t)}$. We define
{\footnotesize
	\begin{equation}
		\mathcal{Z}^{(t)}_i
		=
		\begin{cases}
			\{\textsc{Keep},\textsc{Replace},\textsc{Insert}\}, & a\!\left(f_i^{(t)}\right)=0, \\[2pt]
			\{\textsc{Keep},\textsc{Replace},\textsc{Delete},\textsc{Rewrite}\}, & a\!\left(f_i^{(t)}\right)>0,
		\end{cases}
		\label{eq:tagger_mask}
	\end{equation}
}
where $a(\cdot)$ denotes the arity. To provide a more intuitive understanding of the edit actions, Fig.~\ref{fig:edit_actions} illustrates four representative non-trivial edits on an example expression tree. 

The Tagger predicts a position-wise conditional distribution
$p_{\psi}(z_i=z\mid f^{(t)},h)$ over admissible actions
$z\in\mathcal{Z}^{(t)}_i$. For each position, we retain the admissible action with the highest confidence
\begin{equation}
	\hat z_i^{(t)}
	=
	\arg\max_{z\in\mathcal{Z}^{(t)}_i}
	p_{\psi}(z_i=z\mid f^{(t)},h).
	\label{eq:frontier_score}
\end{equation}
If $\hat z_i^{(t)}=\textsc{Keep}$ for all positions, the current rectification terminates. Otherwise, among all position-action pairs whose optimal action is not \textsc{Keep}, we select the one with the highest confidence score:
\begin{equation}
	p^{(t)}
	=
	\arg\max_{\substack{1\le i\le |f^{(t)}|\\ \hat z_i^{(t)}\neq \textsc{Keep}}}
	p_{\psi}\!\left(z_i=\hat z_i^{(t)} \mid f^{(t)},h\right),
	\qquad
	z^{(t)}=\hat z_{p^{(t)}}^{(t)}.
	\label{eq:tagger_select}
\end{equation}
Overall, the Tagger consists of two stages. It first produces position-wise action predictions for the current state and subsequently retains the optimal position-action pair to pass to the Editor. Algorithm~\ref{alg:tagger_step} summarizes this process.

\begin{algorithm}[h]
	\caption{Tagger prediction algorithm}
	\label{alg:tagger_step}
	\begin{algorithmic}[1]
		\REQUIRE Current state $f^{(t)}$, dataset encoding $h$
				
		\STATE {\itshape Stage I: predicting actions position-by-position.}
		\FOR{$i=1,\dots,|f^{(t)}|$}
		\STATE Build $\mathcal{Z}_i^{(t)}$ by Eq.~\eqref{eq:tagger_mask} \comment{Construct the syntax-admissible action set}
		
		\STATE Compute $p_{\psi}(z_i=z\mid f^{(t)},h)$ for all $z\in\mathcal{Z}_i^{(t)}$
		
		\STATE $\hat z_i^{(t)} \gets \arg\max_{z\in\mathcal{Z}_i^{(t)}} p_{\psi}(z_i=z\mid f^{(t)},h)$ \comment{Select the action for each position}
		\ENDFOR
		\IF{All $\hat z_i^{(t)}$ are \textsc{Keep}}
		\RETURN $(\varnothing,\textsc{Keep})$ \comment{Terminate and return no edit}
		\ENDIF
		\STATE {\itshape Stage II: determining position-action pair}
		\STATE $p^{(t)} \gets \arg\max_{\substack{1\le i\le |f^{(t)}|\\ \hat z_i^{(t)}\neq \textsc{Keep}}}
		p_{\psi}\!\left(z_i=\hat z_i^{(t)} \mid f^{(t)},h\right)$ \comment{Determine the edit position}
		
		\STATE $z^{(t)} \gets \hat z_{p^{(t)}}^{(t)}$		\comment{Determine the edit action}
		
		\RETURN $(p^{(t)}, z^{(t)})$
	\end{algorithmic}
\end{algorithm}

\subsubsection{Editor}
\label{editor}

Once the Tagger outputs the edit position $p^{(t)}$ and the non-\textsc{Keep} edit action $z^{(t)}$, the Editor is used to predict the edit content $u^{(t)}$. Since \textsc{Insert}, \textsc{Rewrite}, and \textsc{Delete} operate on subtrees, we denote by $I^{(t)}(p^{(t)})$ the closed subtree of $f^{(t)}$ rooted at position $p^{(t)}$, and by $|I^{(t)}(p^{(t)})|$ its span length. 

Each execution of an edit action corresponds to a state transition from $f^{(t)}$ to $f^{(t+1)}$. We define a one-step transition function $\mathcal{T}$:
\begin{equation}
	f^{(t+1)}=\mathcal{T}\!\left(f^{(t)},p^{(t)},z^{(t)},u^{(t)}\right).
	\label{eq:transition_function}
\end{equation}
This function specifies how the current state $f^{(t)}$ is transformed to $f^{(t+1)}$ by executing action $z^{(t)}$ at position $p^{(t)}$. Specifically,
{\small
\begin{equation}
	\mathcal{T}\!\left(f,p,z,u\right)
	=
	\begin{cases}
		(f_{<p};u;f_{\ge p+1}),
		& z\in\{\textsc{Replace},\textsc{Insert}\},\\[3pt]
		(f_{<p};u;f_{\ge p+|I(p)|}),
		& z\in\{\textsc{Delete},\textsc{Rewrite}\},\\[3pt]
		f,
		& z=\textsc{Keep},
	\end{cases}
	\label{eq:transition_function_cases}
\end{equation}
}
where $(\,;\,)$ denotes sequence concatenation. For \textsc{Insert} and \textsc{Rewrite}, $u$ denotes a syntactically closed subtree. For \textsc{Replace} and \textsc{Delete}, $u$ degenerates to a single token. 

The Editor is implemented as an infilling decoder that generates tokens only at masked positions. For each non-\textsc{Keep} action, we construct a context state $\tilde f^{(t)}$ by replacing the region that required edit in $f^{(t)}$ with one or multiple action-specific placeholders:
\begin{itemize}
	\item \textsc{Replace:} We construct $\tilde f^{(t)}$ by replacing $f^{(t)}_{p^{(t)}}$ in $f^{(t)}$ with one $\tau_{\mathrm{rep}}$ placeholder.
	
	\item \textsc{Delete:} We construct $\tilde f^{(t)}$ by replacing $I^{(t)}(p^{(t)})$ with one $\tau_{\mathrm{del}}$ placeholder.
	
	\item \textsc{Rewrite:} We construct $\tilde f^{(t)}$ by replacing $I^{(t)}(p^{(t)})$ with $S$ consecutive $\tau_{\mathrm{rew}}$ placeholders starting at $p^{(t)}$.
	
	\item \textsc{Insert:} We construct $\tilde f^{(t)}$ by replacing $f^{(t)}_{p^{(t)}}$ with $S$ consecutive $\tau_{\mathrm{ins}}$ placeholders starting at $p^{(t)}$.
\end{itemize}

To match the two infilling patterns described above, the Editor uses two output heads: a single-token classification head for \textsc{Replace} and \textsc{Delete}, and a sequence generation head for \textsc{Rewrite} and \textsc{Insert}. For the actions \textsc{Replace} and \textsc{Delete}, the edit content is predicted as
\begin{equation}
	u^{(t)} =
	\begin{cases}
		\arg\max_{a(v)=a(f^{(t)}_{p^{(t)}})}
		p_{\phi}\!\left(v \mid \tilde f^{(t)}, h\right),
		& z^{(t)}=\textsc{Replace}, \\[6pt]
		\arg\max_{a(v)=0}
		p_{\phi}\!\left(v \mid \tilde f^{(t)}, h\right),
		& z^{(t)}=\textsc{Delete}.
	\end{cases}
	\label{eq:editor_single}
\end{equation}
For the actions \textsc{Rewrite} and \textsc{Insert}, the edit content is generated autoregressively. The $i$-th token is decoded as
\begin{equation}
	u_i^{(t)}
	=
	\arg\max_{v\in\mathcal{V}'(d_{i-1}^{(t)},R_i^{(t)})}
	p_{\phi}\!\left(v \mid u_{<i}^{(t)},\tilde f^{(t)},h\right).
	\label{eq:editor_decode}
\end{equation}
The resulting subtree distribution factorizes as
\begin{equation}
	p_{\phi}\!\left(u^{(t)} \mid \tilde f^{(t)},h\right)
	=
	\prod_{i=1}^{|u^{(t)}|}
	p_{\phi}\!\left(u_i^{(t)} \mid u_{<i}^{(t)},\tilde f^{(t)},h\right).
	\label{eq:editor_factorization}
\end{equation}
Once the generated tokens form a closed subtree, the generation process terminates. The remaining placeholder slots are then discarded.

We implement the Editor with bidirectional and infilling attention, where non-hole positions attend bidirectionally to all non-hole tokens, while hole positions attend to all non-hole tokens and to earlier hole tokens only. This architecture provides bidirectional context from both sides while maintaining a causal order inside the hole. Algorithm~\ref{alg:editor_decode} details how the Editor generates $u^{(t)}$.

\begin{algorithm}[h]
	\caption{Editor decoding algorithm.}
	\label{alg:editor_decode}

	\begin{algorithmic}[1]
		\REQUIRE Current state $f^{(t)}$, dataset encoding $h$, edit position $p^{(t)}$, edit action $z^{(t)}$, Editor distribution $p_{\phi}$, vocabulary $\mathcal{V}$, placeholder budget $S$
		
		\IF{$z^{(t)}=\textsc{Replace}$}
		\STATE Set $\tilde f^{(t)} \gets f^{(t)}$; Set $\tilde f^{(t)}_{p^{(t)}} \gets \tau_{\mathrm{rep}}$ \comment{Open a hole at the selected position}
		
		\STATE Set $\mathcal{A} \gets \{v\in\mathcal{V}: a(v)=a(f^{(t)}_{p^{(t)}})\}$ \comment{Restrict the replacement token to the same arity}
		
		\STATE Set $u^{(t)} \gets \arg\max_{v\in\mathcal{A}} p_{\phi}(u=v \mid \tilde f^{(t)},h)$
		
		\ELSIF{$z^{(t)}=\textsc{Delete}$}
		\STATE Set $\tilde f^{(t)} \gets (f^{(t)}_{<p^{(t)}};\tau_{\mathrm{del}};f^{(t)}_{\ge p^{(t)}+|I^{(t)}(p^{(t)})|})$ \comment{Remove the source subtree and replace it with one hole}
		
		\STATE Set $\mathcal{A} \gets \{v\in\mathcal{V}: a(v)=0\}$ \comment{Restrict the output to leaf tokens}
		
		\STATE Set $u^{(t)} \gets \arg\max_{v\in\mathcal{A}} p_{\phi}(u=v \mid \tilde f^{(t)},h)$
		
		\ELSIF{$z^{(t)}\in\{\textsc{Rewrite},\textsc{Insert}\}$}
		\IF{$z^{(t)}=\textsc{Rewrite}$}
		\STATE Set $\tilde f^{(t)} \gets (f^{(t)}_{<p^{(t)}};\underbrace{\tau_{\mathrm{rew}},\ldots,\tau_{\mathrm{rew}}}_{S};f^{(t)}_{\ge p^{(t)}+|I^{(t)}(p^{(t)})|})$ \comment{Open a multi-token hole at the current subtree}
		\ELSE
		\STATE Set $\tilde f^{(t)} \gets (f^{(t)}_{<p^{(t)}};\underbrace{\tau_{\mathrm{ins}},\ldots,\tau_{\mathrm{ins}}}_{S};f^{(t)}_{\ge p^{(t)}+1})$ \comment{Open a multi-token hole at the selected position}
		\ENDIF
		
		\STATE Set $u \gets [\;]$, $d \gets 1$
		
		\FOR{$j \gets 1$ \TO $S$}
		\STATE Set $R \gets S-j+1$ \comment{Track the remaining decoding budget}
		
		\STATE Set $u_j \gets \arg\max_{v\in\mathcal{V}'(d,R)} p_{\phi}(u_j=v\mid u_{<j},\tilde f^{(t)},h)$ \comment{Perform syntax-constrained autoregressive decoding}
		
		\STATE $u.\mathrm{append}(u_j)$
		
		\STATE Update $d \gets d-1+a(u_j)$ \comment{Update the local deficit}
		
		\IF{$d=0$}
		\STATE \textbf{Break} \comment{Stop when the generated subtree closes}
		\ENDIF
		\ENDFOR
		
		\STATE Set $u^{(t)} \gets u$
		\ENDIF
		\STATE \textbf{return} $u^{(t)}$
	\end{algorithmic}
\end{algorithm} 

\subsubsection{Training target for Tagger and Editor}
\label{sec:train_targets}

The objective of the Rectifier is to learn a state-transition chain from a given initial state $f^{(0)}$ to the target $f^*$. Therefore, each sampled intermediate state $f^{(t)}$ is associated with a one-step supervision label $\bigl(p^{*(t)},\, z^{*(t)},\, u^{*(t)}\bigr)$, which specifies the edit position, the edit action, and the edit content required to transform $f^{(t)}$ to $f^{(t+1)}$.

During training, ensuring the robustness and stable convergence of the Rectifier requires addressing two issues. One is how to construct initial states $f^{(0)}$ with diverse error patterns. The other is, given $f^{(0)}$, how to construct a deterministic step-by-step rectification chain that moves it toward $f^*$ through fixed edit positions and edit actions, rather than random edits.

Inspired by discrete diffusion models~\citep{reid2022diffuser, havasi2025edit}, which simulate error distributions by randomly corrupting the target, we sample a corruption depth $T'$ and then apply $T'$ random edit actions on the target expression to obtain $f^{(0)}$. Each edit action is randomly drawn from the predefined action space $\mathcal{Z}$ with equal probability and is required to satisfy the action admissibility constraints at the selected position.

Next, we describe how to construct the deterministic rectification chain from $f^{(0)}$ to $f^*$. The core question is how to determine the edit position and edit action to define the state transition from $f^{(t)}$ to $f^{(t+1)}$. To keep the notation consistent, for a position $p$ in $f^{(t)}$, we denote by $p'$ the aligned position in $f^*$. In Section~\ref{editor}, we have defined $I^{(t)}(p)$ as the closed subtree of $f^{(t)}$ rooted at $p$, so the corresponding target subtree in $f^*$ is denoted by $I^*(p')$. We propose a state-transition algorithm (Algorithm~\ref{alg:recursive_select}). It first aligns $f^{(t)}$ with $f^*$ according to the tree structure. The admissible edit actions at $p$ in the state-transition algorithm follow the same constraints as those of the Rectifier:
\begin{itemize}
	\item \textsc{Replace} is admissible if and only if both $f^{(t)}_p$ and $f^*_{p'}$ are leaves, or they have the same arity and all their children already match.
	
	\item \textsc{Delete} is admissible if and only if $f^{(t)}_p$ induces a subtree $I^{(t)}(p)$, while the aligned $f^*_{p'}$ is a leaf.
	
	\item \textsc{Insert} is admissible if and only if $f^{(t)}_p$ is a leaf, while the aligned $f^*_{p'}$ induces a subtree $I^*(p')$, and  $|I^*(p')|$ does not exceed the generation budget $S$ of the Editor.
	
	\item \textsc{Rewrite} is admissible if and only if $f^{(t)}_p$ and $f^*_{p'}$ induce two different subtrees, $I^{(t)}(p)$ and $I^*(p')$, and $|I^*(p')|$ does not exceed the generation budget $S$ of the Editor.
\end{itemize}
Whenever an action is admissible, we define a cost as follows:
{\small
	\begin{equation}
		\begin{aligned}
			\omega(z;p,p')
			&=
			\begin{cases}
				0, & z=\textsc{Replace}\ \text{and}\ f^{(t)}_{p}=f^{*}_{p'},\\
				1, & z=\textsc{Replace}\ \text{and}\ f^{(t)}_{p}\neq f^{*}_{p'},\\
				1+\lambda\,\lvert I^{(t)}(p)\rvert, & z=\textsc{Delete},\\
				1+\lambda\,\lvert I^*(p')\rvert, & z\in\{\textsc{Insert},\textsc{Rewrite}\}.
			\end{cases}
		\end{aligned}
		\label{eq:frontier_cost_34}
\end{equation}}
Here, \textsc{Rewrite} and \textsc{Insert} have the same cost. They operate on the root node and leaf node respectively, and therefore do not compete with each other. $\lambda$ is a subtree length penalty, which together with the Editor generation budget $S$ constitutes a constraint to encourage the Rectifier to learn multi-step local rectification, rather than greedily attempting to solve a large discrepancy in one-step \textsc{Rewrite} or \textsc{Insert}.  Once a subtree exceeds the budget $S$, we introduce an intermediate subtree $\tilde{I}^*(p')$, where $|\tilde{I}^*(p')|$ does not exceed $S$. It is constructed by retaining the root of $I^*(p')$ and expanding it with as many of its children as permitted by the budget. The omitted children are replaced by a leaf so that the subtree remains syntactically closed. The corresponding cost is defined as the bridge cost from $I^{(t)}(p)$ to $\tilde{I}^*(p')$ plus the remaining cost from  $\tilde{I}^*(p')$ to $I^*(p')$.

The state-transition algorithm operates recursively. Starting from the root position pair $(p=1, p'=1)$ of $f^{(t)}$ and $f^*$, the algorithm determines the edit position $p^{*(t)}$ and edit action $z^{*(t)}$ to handle the difference between $I^{(t)}(p)$ and $I^*(p')$ by comparing two strategies. The first is to execute a one-step admissible action at $p$ immediately. The second is to defer the edit at $p$ and examine deeper aligned descendant nodes to find a lower-cost scheme. Note that the latter strategy requires $f_p^{(t)}$ and $f_{p'}^*$ to have the same arity. If this condition holds but the symbols differ, a unit \textsc{Replace} cost is accumulated at $p$.

\begin{algorithm}[H]
	\caption{State transition algorithm.}
	\label{alg:recursive_select}
		\small
	\begin{algorithmic}[1]
		\REQUIRE Current state $f^{(t)}$, target state $f^*$, position pair $(p,p')$, Editor generation budget $S$, subtree penalty $\lambda$, transition function $\mathcal{T}$, cost function$\omega$ (Eq.~\ref{eq:frontier_cost_34})
		\STATE \textsc{StateTransition}$(f^{(t)},f^*,p,p')$
		\IF{$I^{(t)}(p)=I^*(p')$}
		\STATE \textbf{return} $(0,\varnothing,\varnothing,\varnothing)$ \comment{Return because no edit is needed}
		\ENDIF
		
		\STATE $\mathrm{Cost}_{\mathrm{I}}^*\gets+\infty$, $(p_{\mathrm{I}}^*,z_{\mathrm{I}}^*,u_{\mathrm{I}}^*)\gets(\varnothing,\varnothing,\varnothing)$
		\STATE Set $\mathcal{A}(p,p') \subseteq Z_p^{(t)}$   \comment{Build an admissible action set at position pair $(p,p')$}
		\STATE {\itshape Stage I: execute one admissible action directly at $p$.}
		\FOR{$z \in \mathcal{A}(p,p')$ \textbackslash $\textsc{Keep}$}
		\IF{$z\in\{\textsc{Replace},\textsc{Delete}\}$}
		\STATE $u\gets f^*_{(p')}$, $c\gets\omega(z;p,p')$ \comment{Use the aligned target token}
		\ELSIF{$z\in\{\textsc{Insert},\textsc{Rewrite}\}$}
		\IF{$|I^*(p')|\le S$}
		\STATE $u\gets I^*(p')$, $c\gets\omega(z;p,p')$ \comment{Use the target subtree directly}
		\ELSE
		\STATE $u\gets \tilde I^*(p')$ \comment{Use the budget-limited subtree}
		\STATE $(c_{\mathrm{rem}},\_,\_,\_)\gets \textsc{StateTransition}(\mathcal{T}(f^{(t)},p,z,u),f^*,p,p')$ \comment{Compute the remaining cost recursively}
		\STATE $c\gets\omega(z;p,p')+c_{\mathrm{rem}}$
		\ENDIF
		\ENDIF
		
		\IF{$c<\mathrm{Cost}_{\mathrm{I}}^*$}
		\STATE $\mathrm{Cost}_{\mathrm{I}}^*\gets c$, $(p_{\mathrm{I}}^*,z_{\mathrm{I}}^*,u_{\mathrm{I}}^*)\gets(p,z,u)$ \comment{Keep the best direct action}
		\ENDIF
		\ENDFOR
		
		\STATE {\itshape Stage II: defer the edit at $p$, recursively process the aligned child nodes.}
		\STATE $\mathrm{Cost}_{\mathrm{II}}^*\gets+\infty$, $(p_{\mathrm{II}}^*,z_{\mathrm{II}}^*,u_{\mathrm{II}}^*)\gets(\varnothing,\varnothing,\varnothing)$
		
		\IF{$a(f^{(t)}_{(p)})=a(f^*_{(p')})>0$}
		\IF{$f^{(t)}_{(p)}\neq f^*_{(p')}$}
		\STATE $c\gets\omega(\textsc{Replace};p,p')$, $(\hat p,\hat z,\hat u)\gets(p,\textsc{Replace},f^*_{(p')})$ \comment{Use \textsc{Replace} as the fallback}
		\ELSE
		\STATE $c\gets0$, $(\hat p,\hat z,\hat u)\gets(\varnothing,\varnothing,\varnothing)$
		\ENDIF
		
		\FOR{$j\gets1$ \TO $a(f^{(t)}_{(p)})$}
		\STATE $(c_j,\tilde p,\tilde z,\tilde u)\gets\textsc{StateTransition}(f^{(t)},f^*,p_j,p'_j)$
		\STATE $c\gets c+c_j$ \comment{Accumulate the recursive cost}
		\IF{$\tilde p\neq\varnothing$ \AND $(\hat p=\varnothing$ $\OR$ $\hat p=p)$}
		\STATE $(\hat p,\hat z,\hat u)\gets(\tilde p,\tilde z,\tilde u)$
		\ENDIF
		\ENDFOR
		
		\STATE $\mathrm{Cost}_{\mathrm{II}}^*\gets c$, $(p_{\mathrm{II}}^*,z_{\mathrm{II}}^*,u_{\mathrm{II}}^*)\gets(\hat p,\hat z,\hat u)$ 
		\ENDIF
		
		\STATE {\itshape Stage III: Compare the best direct-edit plan and the best deferred-edit plan.}
		\IF{$\mathrm{Cost}_{\mathrm{II}}^*\le\mathrm{Cost}_{\mathrm{I}}^*$}
		\STATE  $\mathrm{Cost}^*\gets\mathrm{Cost}_{\mathrm{II}}^*$, $(p^*,z^*,u^*)\gets(p_{\mathrm{II}}^*,z_{\mathrm{II}}^*,u_{\mathrm{II}}^*)$
		\ELSE
		\STATE $\mathrm{Cost}^*\gets\mathrm{Cost}_{\mathrm{I}}^*$, $(p^*,z^*,u^*)\gets(p_{\mathrm{I}}^*,z_{\mathrm{I}}^*,u_{\mathrm{I}}^*)$
		\ENDIF
		\STATE \textbf{return} $(\mathrm{Cost}^*,p^*,z^*,u^*)$
	\end{algorithmic}
\end{algorithm}

Given the selected edit position $p^{*(t)}$, the corresponding edit content $u^{*(t)}$ is defined according to the selected edit action $z^{*(t)}$ as follows:
\begin{itemize}
	\item If $z^{*(t)}=\textsc{Replace, Delete}$, then $u^{*(t)}=f^*_{(p'^{*(t)})}$.
	
	\item If $z^{*(t)}=\textsc{Insert, Rewrite}$, then $u^{*(t)}$ is the aligned target subtree at $p^{*(t)}$. Specifically, if $\lvert I^*(p'^{*(t)})\rvert \le S$, then $u^{*(t)}=I^*(p'^{*(t)})$; otherwise, $u^{*(t)}=\tilde I^*(p'^{*(t)})$, i.e., the budget-constrained intermediate subtree.
\end{itemize}

Note that the state-transition algorithm does not aim to transform $f^{(t)}$ into $f^*$ in a single step. Instead, at each non-terminal state, it determines one supervised triplet $\bigl(p^{*(t)}, z^{*(t)}, u^{*(t)}\bigr)$, which induces a one-step state transition from $f^{(t)}$ to $f^{(t+1)}$ through $f^{(t+1)}=\mathcal{T}\!\left(f^{(t)},p^{*(t)},z^{*(t)},u^{*(t)}\right)$. Repeating this process yields a rectification chain
\[
f^{(0)} \to f^{(1)} \to \cdots \to f^{(t)} \to \cdots \to f^*,
\]
with the total cost minimized under Eq.~\ref{eq:frontier_cost_34}. Algorithm~\ref{alg:build_chain} summarizes this process.

\subsection{Optimization}
\label{Optimization}
The Rectifier loss consists of the Tagger loss and the Editor loss, as follows:
\begin{equation}
	\mathcal{L}_{\mathrm{Rectifier}}
	=
	\mathcal{L}_{\mathrm{Tagger}}+\mathcal{L}_{\mathrm{Editor}}.
	\label{eq:lrect}
\end{equation}
Here, the Tagger is trained with a position-wise multi-class cross-entropy where the selected position $p^{*(t)}$ is supervised by the target action $z^{*(t)}$, while all other positions are supervised as \textsc{Keep}:
\begin{equation}
	\begin{aligned}
		\mathcal{L}_{\mathrm{Tagger}}^{(t)}
		={}&
		-\log p_{\psi}\!\left(z_{p^{*(t)}}=z^{*(t)} \mid f^{(t)},h\right) \\
		&\;-\sum_{i\ne p^{*(t)}}
		\log p_{\psi}\!\left(z_i=\textsc{Keep} \mid f^{(t)},h\right),
	\end{aligned}
\end{equation}
and the total loss on the entire rectification chain is
\begin{equation}
	\mathcal{L}_{\mathrm{Tagger}}
	= \sum_{t=0}^{T-1}\mathcal{L}_{\mathrm{Tagger}}^{(t)}.
	\label{eq:loss_tokenizer}
\end{equation}

The Editor is supervised directly by the target edit content $u^{*(t)}$, with the one-step loss
\begin{equation}
	\mathcal{L}_{\mathrm{Editor}}^{(t)}
	=
	-\mathbf{1}\!\left[z^{*(t)}\neq \textsc{Keep}\right]
	\sum_{j=1}^{|u^{*(t)}|}
	\log p_{\phi}\!\left(
	u_j^{*(t)} \mid u_{<j}^{*(t)},\tilde f^{(t)},h
	\right),
\end{equation}
and the total loss on the entire rectification chain is
\begin{equation}
	\mathcal{L}_{\mathrm{Editor}}
	=
	\sum_{t=0}^{T-1}\mathcal{L}_{\mathrm{Editor}}^{(t)}.
	\label{eq:loss_editor}
\end{equation}
For actions \textsc{Replace} and \textsc{Delete}, the loss function degenerates into a classification loss.

\begin{algorithm}[h]
	\caption{Rectification chain construction algorithm.}
	\label{alg:build_chain}
	\begin{algorithmic}[1]
		\REQUIRE Initial state $f^{(0)}$, target $f^*$
		
		\STATE $t\gets 0$, \ $\mathcal{C}\gets[\;]$ 
		
		\WHILE{$f^{(t)}\neq f^*$}
		
		\STATE $(\mathrm{Cost}^*,p^{*(t)},z^{*(t)},u^{*(t)})\gets \textsc{StateTransition}(f^{(t)},f^*,1,1)$  	\comment{Invoke Algorithm~\ref{alg:recursive_select} to determine the edit position, action, and content}
		
		\STATE  $\mathcal{C} \gets (\mathcal{C}; (p^{*(t)},z^{*(t)},u^{*(t)}))$
		
		\STATE $f^{(t+1)}\gets \mathcal{T}\!\left(f^{(t)},p^{*(t)},z^{*(t)},u^{*(t)}\right)$ \comment{Execute the state transition}
		
		\STATE $t\gets t+1$
		\ENDWHILE
		
		\STATE \textbf{return} $(f^{(0)},f^{(1)},\ldots,f^{(T)})$, $\mathcal{C}$ 
		
	\end{algorithmic}
\end{algorithm}

We train EditSR in two stages. First, we optimize only the first layer using $\mathcal{L}_{\mathrm{base}}$ (Eq.~\ref{eq:lar_ar}) to obtain a strong first-layer neural model. After the first layer converges, we freeze its parameters and optimize only the Rectifier by minimizing $\mathcal{L}_{\mathrm{Rectifier}}$ (Eq.~\ref{eq:lrect}).

In the second stage, we mainly train the Rectifier on artificially constructed rectification chains. After that, we perform a short fine-tuning, in which the outputs of the first layer are used as initial states, so that the Rectifier is better aligned with the error patterns encountered during inference. 

\subsection{Inference}

During inference, we apply beam search to the first layer to obtain a set of candidates $\{\hat{f}_i\}_{i=1}^{B}$, where $B$ is the beam size. Any candidate that already satisfies the stopping criterion is accepted directly. Only candidates that fail this check are treated as independent initial states for rectification.


We adopt a greedy rectification process rather than beam search. In our framework, evaluating competing states is challenging because different edit actions produce sequences of varying lengths and differing potentials for subsequent rectification. Furthermore, unlike autoregressive decoding, edit-based models are not tied to irrevocable decisions since content can be freely rewritten at later steps. Previous research, such as Levenshtein Transformer~\citep{gu2019levenshtein}, also demonstrates that beam search yields marginal benefits in such scenarios. Consequently, we employ greedy inference to maintain high computational efficiency.

Starting from the initial state $f^{(0)}$, the Rectifier applies one edit action at each iteration to transform the current state $f^{(t)}$ to the next state. Then, we evaluate the numerical error. The rectification process stops immediately once the error threshold is met. Otherwise, rectification proceeds until either \textsc{Keep} is selected or the step budget $T_{\max}$ is exhausted. Algorithm~\ref{alg:greedy_rectify} outlines a complete rectification process for a given initial state.

\begin{algorithm}[h]
	\caption{Rectification algorithm.}
	\label{alg:greedy_rectify}
	\begin{algorithmic}[1]
		\REQUIRE Dataset $\mathcal{D}=\{(x_n,y_n)\}_{n=1}^{N}$, Dataset encoding $h$; initial state $f^{(0)}$; Tagger distribution $p_{\psi}$; Editor distribution $p_{\phi}$; maximum rectification steps $T_{\max}$; stopping threshold $\mathrm{MSE}_{\mathrm{stop}}$
		
		\STATE $\mathcal{P}\gets\{f^{(0)}\}$
		
		\FOR{$t=0,1,\ldots,T_{\max}-1$}
		\STATE Compute $(p^{(t)},z^{(t)})$ via Algorithm~\ref{alg:tagger_step}
		
		\IF{$z^{(t)}=\textsc{Keep}$}
		\STATE \textbf{break} \comment{Terminate rectification early}
		\ENDIF
		
		\STATE Compute $u^{(t)}$ via Algorithm~\ref{alg:editor_decode}
		
		\STATE $f^{(t+1)} \gets \mathcal{T}\!\left(f^{(t)},p^{(t)},z^{(t)},u^{(t)}\right)$ \comment{Execute the state transition}
		
		\STATE $\mathcal{P}\gets \mathcal{P}\cup\{f^{(t+1)}\}$
		
		\STATE Apply BFGS to $f^{(t+1)}$ and compute $\mathrm{MSE}^{(t+1)}$ \comment{Evaluate the error}
		
		\IF{$\mathrm{MSE}^{(t+1)}\le \mathrm{MSE}_{\mathrm{stop}}$}
		\STATE \textbf{return} $f^{(t+1)}$
		\ENDIF
		\ENDFOR
		
		\STATE \textbf{return} $\arg\min_{f\in\mathcal{P}} \mathrm{MSE}(f)$
	\end{algorithmic}
\end{algorithm}


\section{Data generation}
\label{data_generation}

In this section, we present the method used for generating training data. Each sample is generated in three stages: first, sampling an expression skeleton via a skeleton generator; second, instantiating the skeleton by sampling constants to obtain a parseable expression; and finally, sampling variable values and evaluating the expression to form a dataset. To improve the model's generalization, the training set contains 100 million randomly generated skeletons.

\subsection{Skeleton generation}
\label{subsec:skeleton_generation}

Following common practice in symbolic regression~\citep{biggio2021neural}, we represent an expression skeleton as a binary tree, where internal nodes are operators and leaf nodes are variables or constants. To prevent overly large or degenerate structures and to control the diversity of the generated skeletons, we enforce hard limits on the input dimension, the total number of nodes, and the counts of unary operators, binary operators, and constants. Detailed parameters are presented in Table~\ref{tab:equation_generator}.

\subsection{Constant and dataset sampling}

For each skeleton, we generate multiple datasets by resampling both the constants and the variables. Specifically, constants are drawn independently from the distribution in Table~\ref{tab:equation_generator}, yielding a parseable expression. After that, we sample $N$ values for each variable and compute the corresponding outputs by evaluating the expression. To make the generated data more robust across different magnitudes, we use a hybrid sampling approach that integrates both linear and logarithmic scales.
Concretely, we first sample two scalars from $\mathcal{U}(-10,10)$ and sort them to obtain the interval $[a,b]$. Then, for each variable, we select one of the following two sampling schemes with equal probability:
\begin{itemize}
	\item \textbf{Linear-uniform sampling:} $x \sim \mathcal{U}(a,b)$.
	\item \textbf{Log-scale sampling:} if $a$ and $b$ have the same sign, we sample the magnitude on a logarithmic scale and keep the sign fixed, i.e.,
	$|x| \sim \text{Log-}\mathcal{U}(|a|,|b|)$ and $\mathrm{sign}(x)=\mathrm{sign}(a)$.
\end{itemize}

During expression evaluation, numerical instabilities may arise from operations such as division, logarithms, or square roots. If a non-finite value ($NaN$ or $Inf$) is produced for any sampled point, we replace it with $0$ to keep the generated datasets numerically stable.

\section{Experimental design}
\label{experiment_design}
In this section, we describe the experimental design used to evaluate EditSR, including training and testing details, benchmarks, baselines, and metrics. Together, these settings provide a broad evaluation of EditSR.

\subsection{Training details}

The parameters during training are provided in Table~\ref{tab:rectifier_hparams}. In addition to the standard setting, we also randomly add 0--3 distractor variables, as well as noise with levels following $\mathcal{U}(0, 0.1)$.  

\subsection{Testing details}
During testing, EditSR predicts only the expression skeleton, while the constants are optimized afterward by BFGS. This setting is consistent with the inference pipeline of most neural symbolic regression models (especially NeSymReS). For large datasets, we adopt the bagging strategy following~\citep{kamienny2022end}. Specifically, each bag is a subset containing 200 sampled points. For each test problem, we perform at most 10 bagging runs during evaluation. This strategy reduces the computational cost on large datasets while preserving the robustness of the test procedure. Detailed parameter settings are reported in Table~\ref{tab:rectifier_hparams}. 
\subsection{Benchmarks}
\label{subsec:benchmarks}

We evaluate EditSR on three benchmark categories:
\begin{itemize}
	\item \textbf{Standard benchmarks:}
	We summarize several commonly used benchmarks, including Constant, Koza, Nguyen, Keijzer, Korns, Livermore, Neat and Jin. The problems in these benchmarks involve at most three variables. 
	
	\item \textbf{SRBench 1.0:} SRBench 1.0~\citep{la2021contemporary} is a commonly used benchmark family for symbolic regression. We utilize two of its widely studied benchmarks:
	\begin{itemize}
		\item \textbf{Feynman:} A collection of first-principles physics equations~\citep{udrescu2020ai} with sampled datasets, widely used to evaluate symbolic structure recovery and robustness to noise.
		
		\item \textbf{ODE-Strogatz:} Nonlinear dynamical-system datasets derived from coupled first-order ODEs. They test recovery of simple governing equations in challenging regimes~\citep{strogatz2018nonlinear}.
	\end{itemize}
	
	\item \textbf{SRBench 2.0:} SRBench 2.0~\citep{aldeia2025call} builds upon SRBench 1.0 by introducing a broader collection of problems. The benchmark consists of two complementary components:
	\begin{itemize}
	\item \textbf{Black-box:} A curated set of black-box tasks based on SRBench 1.0~\citep{la2021contemporary}. Problems that are readily solvable using linear regression are removed so that the benchmark is not skewed toward trivial problems.
	
	\item \textbf{Phenomenological \& first-principles:} A collection of real-world scientific discovery problems constructed from experimental observations and physics-inspired formulations, with realistic levels of noise taken into account.
	\end{itemize}
\end{itemize}
Except for the standard benchmarks, for which we manually sample the datasets, all other benchmarks use the officially released datasets. In Table~\ref{tab:benchmark_statistics}, we summarize the statistics of each benchmark. For all officially released datasets, we follow the public train--test splits.

\subsection{Baselines}
\label{subsec:baselines}

We compare \textsc{EditSR} with the following representative baselines:

\begin{itemize}
	
	\item \textbf{uDSR}~\citep{landajuela2022unified}:
	A unified framework that integrates multiple symbolic regression strategies, including recursive problem simplification, risk-seeking controller, pretraining, genetic programming, and local optimization, yielding strong symbolic regression performance in generating simple expressions.
	
	\item \textbf{SR4MDL}~\citep{yu2024symbolic}:
	A search-based framework that introduces the Minimum Description Length (MDL) principle into symbolic regression, utilizing a pretrained MDLformer to guide the search process toward parsimonious mathematical forms while balancing numerical fitting and Complexity.
	
	\item \textbf{ParFam}~\citep{scholl2025parfam}:
	A parametric-family approach that translates discrete symbolic regression into continuous global optimization over structured function families. It also proposes a Transformer-guided variant (DL-ParFam) to accelerate optimization.
	
	\item \textbf{RILS-ROLS}~\citep{kartelj2023rils}:
	A metaheuristic method that performs structure search via iterated local search and estimates coefficients in linear components using ordinary least squares, iteratively refining candidate expressions.

	\item \textbf{TPSR}~\citep{shojaee2023transformer}:
	A classic post-hoc rectification strategy that augments a neural symbolic regression model with MCTS during decoding, leveraging non-differentiable rewards such as error and complexity to guide the underlying Transformer toward globally preferable expressions.

\end{itemize}
For all baselines, we follow the parameter settings summarized in Table~\ref{tab:baselines_parameters}. For TPSR and SR4MDL, we employ the same bagging strategy as EditSR.

\newcommand{\std}[1]{${\scriptstyle \pm #1}$}

\begin{table*}[h]
	\centering
	\caption{Mean $R^2$ and Complexity results over 10 runs on standard benchmarks, reported as mean $\pm$ standard deviation. The best and second-best results are marked in \textbf{bold} and \underline{underlined}, respectively.}
	\label{tab:stand_r2_Complexity_results}
	\renewcommand{\arraystretch}{1.15}
	\setlength{\tabcolsep}{4pt}
	\resizebox{\textwidth}{!}{
		\begin{tabular}{l | c | c c c c c c}
			\toprule
			\textbf{Benchmarks} & \textbf{Metric} & \textbf{EditSR} & \textbf{uDSR} & \textbf{SR4MDL} & \textbf{TPSR} & \textbf{RILS-ROLS} & \textbf{ParFam} \\
			\midrule
			\multirow{2}{*}{Constant} 
			& $R^2$ & \textbf{0.9975}\std{0.0014} & 0.9538\std{0.0458} & \underline{0.9847}\std{0.0058} & 0.9533\std{0.0389} & 0.9785\std{0.0185} & 0.9710\std{0.0270} \\
			& Complexity & \textbf{11.92}\std{0.26} & 31.71\std{1.99} & \underline{17.14}\std{4.33} & 30.62\std{0.88} & 21.71\std{1.44} & 84.21\std{8.88} \\
			\midrule
			
			\multirow{2}{*}{Jin} 
			& $R^2$ & \underline{0.9812}\std{0.0007} & 0.9475\std{0.0005} & 0.8762\std{0.0692} & \textbf{0.9845}\std{0.0269} & 0.9344\std{0.0054} & 0.9124\std{0.0225} \\
			& Complexity & \textbf{12.28}\std{0.10} & 38.17\std{1.15} & 20.67\std{1.20} & 27.22\std{2.26} & \underline{19.39}\std{0.54} & 81.67\std{16.66} \\
			\midrule
			
			\multirow{2}{*}{Keijzer} 
			& $R^2$ & \underline{0.9853}\std{0.0004} & 0.9259\std{0.0218} & 0.9550\std{0.0106} & \textbf{0.9895}\std{0.0056} & 0.9841\std{0.0053} & 0.9416\std{0.0052} \\
			& Complexity & \textbf{14.31}\std{0.22} & 18.18\std{3.05} & \underline{18.07}\std{1.39} & 32.86\std{1.30} & 19.48\std{2.16} & 74.36\std{22.65} \\
			\midrule
			
			\multirow{2}{*}{Korns} 
			& $R^2$ & \underline{0.6278}\std{0.0573} & 0.5481\std{0.0411} & \textbf{0.8716}\std{0.1572} & 0.3138\std{0.0173} & 0.5354\std{0.0337} & 0.5782\std{0.0379} \\
			& Complexity & \underline{15.82}\std{0.26} & 68.80\std{6.79} & 22.69\std{2.20} & 35.71\std{1.85} & \textbf{15.58}\std{3.52} & 113.38\std{50.49} \\
			\midrule
			
			\multirow{2}{*}{Koza} 
			& $R^2$ & \textbf{1.0000}\std{0.0000} & \textbf{1.0000}\std{0.0000} & 0.9991\std{0.0008} & 0.9974\std{0.0002} & \underline{0.9996}\std{0.0002} & \underline{0.9996}\std{0.0006} \\
			& Complexity & \underline{15.67}\std{0.58} & 18.56\std{6.31} & 22.56\std{2.50} & 20.67\std{4.67} & \textbf{14.44}\std{0.51} & 46.22\std{8.28} \\
			\midrule
			
			\multirow{2}{*}{Livermore} 
			& $R^2$ & \textbf{0.9978}\std{0.0014} & 0.9355\std{0.0091} & 0.9413\std{0.0193} & \underline{0.9573}\std{0.0236} & 0.9204\std{0.0188} & 0.9341\std{0.0101} \\
			& Complexity & \textbf{10.33}\std{0.78} & 20.29\std{0.65} & 17.85\std{1.51} & 29.83\std{0.69} & \underline{15.94}\std{0.17} & 88.08\std{27.06} \\
			\midrule
			
			\multirow{2}{*}{Neat} 
			& $R^2$ & \textbf{0.9985}\std{0.0003} & 0.8903\std{0.0044} & 0.9631\std{0.0527} & \underline{0.9975}\std{0.0008} & 0.9477\std{0.0384} & 0.8848\std{0.0209} \\
			& Complexity & \textbf{11.62}\std{0.22} & 21.04\std{4.94} & 63.83\std{20.97} & 26.58\std{1.94} & \underline{19.50}\std{0.50} & 89.29\std{14.82} \\
			\midrule
			
			\multirow{2}{*}{Nguyen} 
			& $R^2$ & \textbf{1.0000}\std{0.0000} & \textbf{1.0000}\std{0.0000} & 0.9873\std{0.0128} & 0.9992\std{0.0005} & 0.9994\std{0.0004} & \underline{0.9999}\std{2.4e-05} \\
			& Complexity & 15.00\std{0.00} & \textbf{12.06}\std{2.82} & 46.64\std{2.81} & 23.79\std{1.47} & \underline{13.48}\std{0.19} & 72.95\std{16.32} \\
			\bottomrule
		\end{tabular}
	}
\end{table*}

\subsection{Metrics}
\label{subsec:metrics}

Recent analyses of symbolic regression benchmarks emphasize that performance should be read as a multi-objective profile rather than as a single scalar score. Therefore, we report four complementary metrics to assess EditSR:

\begin{itemize}
	
	\item \textbf{Noise robustness} measures the capability of the model to reconstruct correct expressions when the datasets are contaminated by noise. We corrupt the clean target values with additive Gaussian noise as follows:
	\begin{equation}
		\tilde{y}_i = y_i + \sigma \cdot \mathrm{std}(\mathbf{y}) \cdot \epsilon_i,\qquad \epsilon_i \sim \mathcal{N}(0,1),
	\end{equation}
	where $y_i$ denotes the original target value, $\tilde{y}_i$ is the corresponding corrupted target, $\mathrm{std}(\mathbf{y})$ is the standard deviation of the target values, and $\sigma$ controls the noise intensity.
	
	\item \textbf{$R^2$} measures goodness-of-fit on the dataset,
	\begin{equation}
		R^2 = 1 - \frac{\sum_{i=1}^n (y_i - \hat{y}_i)^2}{\sum_{i=1}^n (y_i - \overline{y})^2}.
	\end{equation}
	We treat a prediction as an accurate solution when $R^2 \geq 0.999$.
	The Accuracy solution rate is defined as the proportion of problems for which the predicted expression attains $R^2 \geq 0.999$.
	
	\item \textbf{Complexity} is quantified by the total number of nodes in the expression tree. When multiple expressions achieve similar fitting performance, the one with lower complexity is considered more desirable.
	
	\item \textbf{Symbolic solution} measures structural consistency between the predicted and target expressions, following the criterion introduced in SRBench~\citep{la2021contemporary}:
	\begin{definition}
		A predicted expression $\hat{f}$ is defined as a symbolic solution to a problem with target expression $f$ if $\hat{f}$ cannot be simplified to a constant and one of the following conditions is met:
		(i) $f = \hat{f} + \alpha$ for some constant $\alpha$; or
		(ii) $f = \hat{f}/\beta$ for some nonzero constant $\beta$.
	\end{definition}
	Accordingly, the \textbf{Symbolic solution rate} denotes the percentage of problems whose predicted expressions satisfy this criterion.

\end{itemize}

\section{Results}
\label{Results}

In this section, we first report EditSR's evaluation results across benchmarks to clarify which metrics it is competitive on and whether it achieves the desired balance across them. Finally, the ablation studies analyze why the Rectifier is effective and under which conditions it is most beneficial. 

\subsection{Results on standard benchmarks}

In this section, we evaluate EditSR on the standard benchmarks. Following \citep{mcdermott2012genetic}, we sample 200 points for each problem to form the dataset. This protocol covers a diverse set of sampling settings, including uniform sampling, equally spaced sampling, narrow ranges, and wide ranges, which provide a sufficiently broad test of models' robustness. Moreover, according to~\citep{mcdermott2012genetic}, the training and test sets are sampled independently from different ranges. This means that the benchmark naturally contains both interpolation and extrapolation cases, so we do not treat extrapolation as a separate evaluation scenario.

Table~\ref{tab:stand_r2_Complexity_results} reports the mean $R^2$ and Complexity results. EditSR remains in the leading group in mean $R^2$ across most benchmarks, often ranking first or second. Concurrently, it usually generates simpler expressions. TPSR attains the highest mean $R^2$ on Jin and Keijzer, but typically does so with higher Complexity. SR4MDL also returns simple expressions on several benchmarks, which is consistent with its objective of aligning search with symbolic rather than purely numerical fitting. ParFam is flexible but often generates complex expressions when faced with difficult problems. Overall, the results demonstrate that when EditSR improves $R^2$, this benefit is usually not accompanied by an increase in Complexity.

\begin{figure}[h]
	\centering
	\begin{subfigure}[h]{\linewidth}
		\centering
		\includegraphics[width=0.9\linewidth]{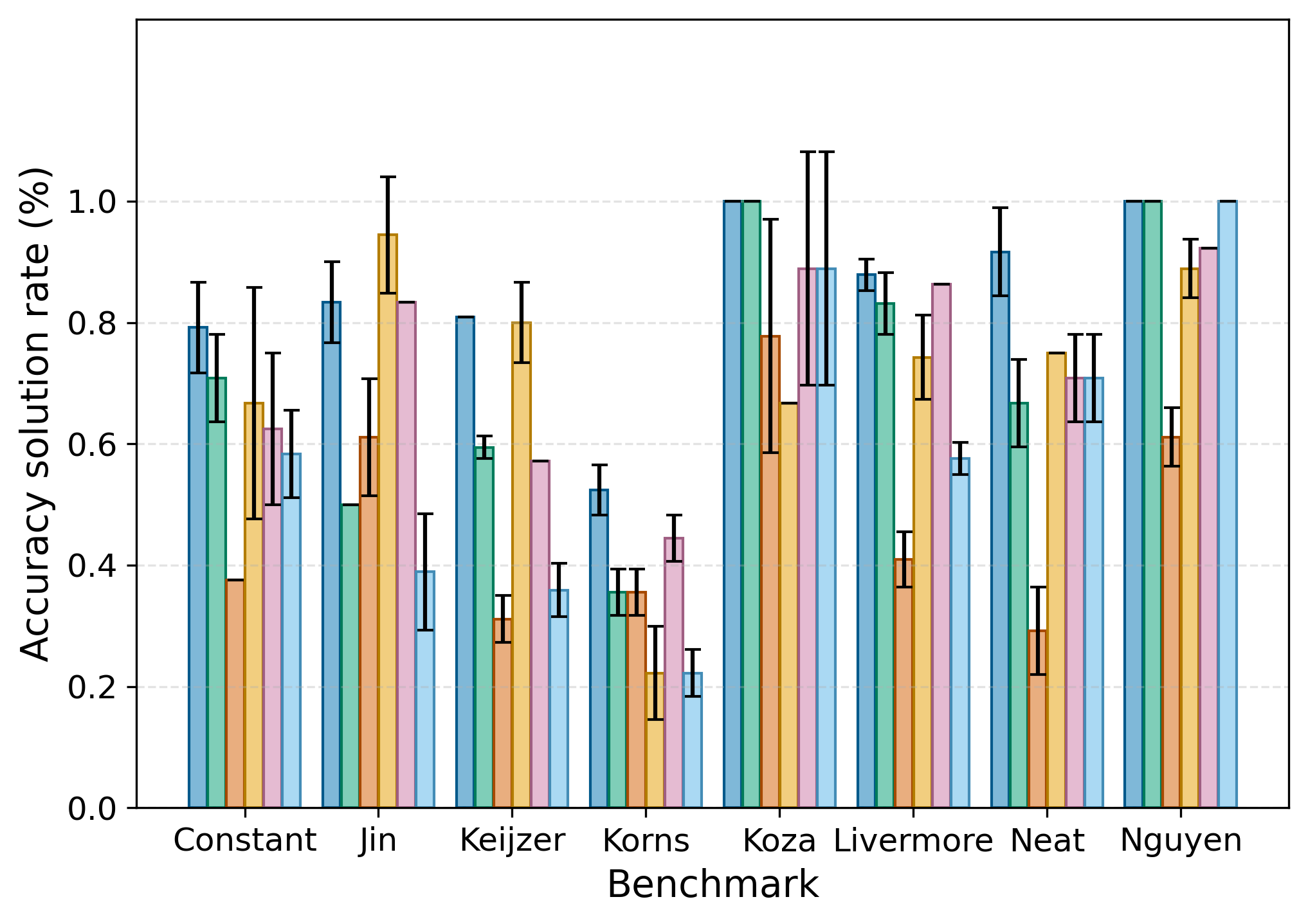}
		\caption{Accuracy solution rate}
	\end{subfigure}
	\hfill
	\begin{subfigure}[h]{\linewidth}
		\centering
		\includegraphics[width=0.9\linewidth]{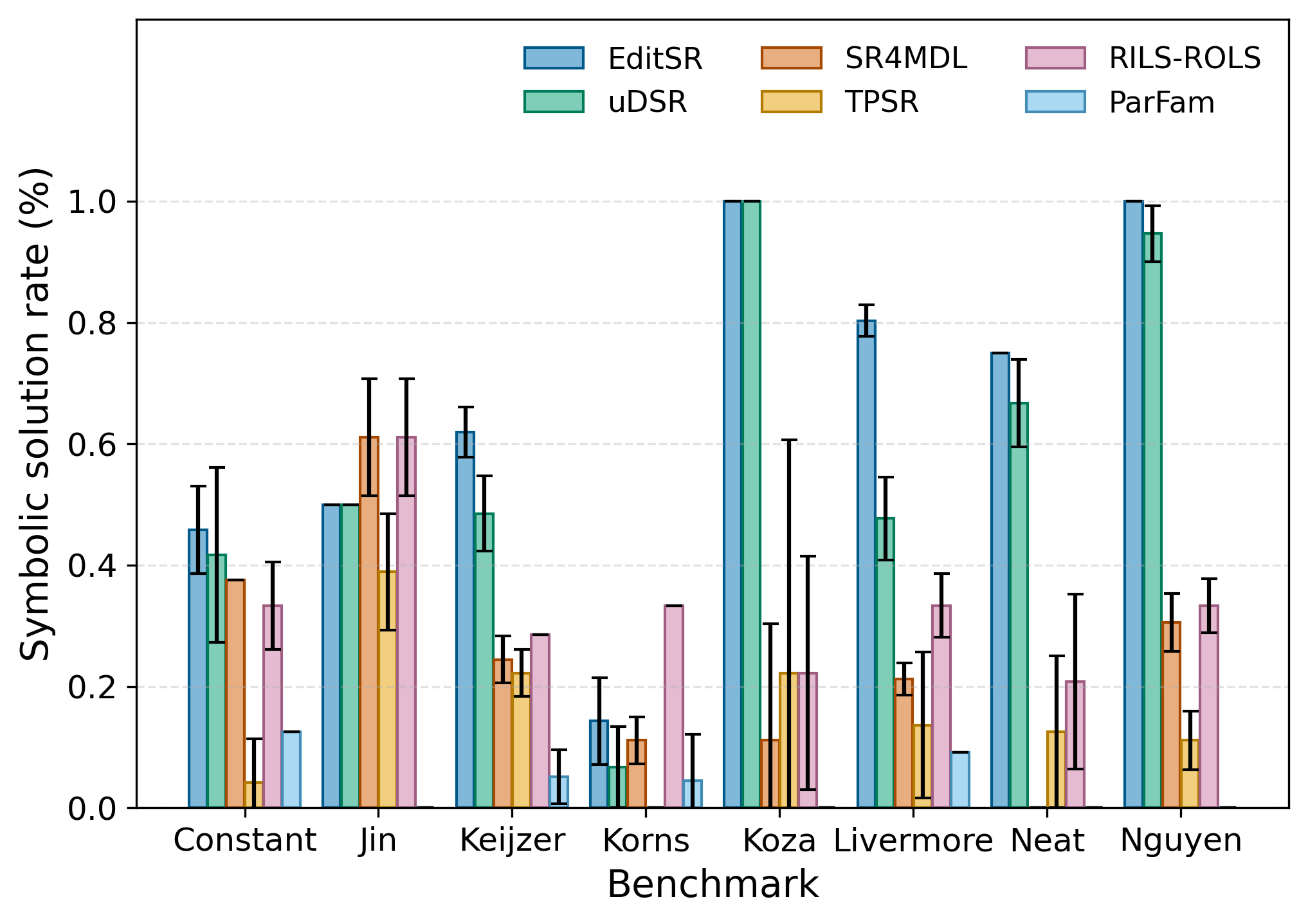}
		\caption{Symbolic solution rate}
	\end{subfigure}
	\caption{\textbf{Accuracy solution rate and Symbolic solution rate results on standard benchmarks.} The results are the mean of 10 runs, and the error bars denote the standard deviation.}
	\label{fig:stand_acc_sym_solution_rate}
\end{figure}

Fig.~\ref{fig:stand_acc_sym_solution_rate} reports the Accuracy solution rate and Symbolic solution rate on the standard benchmarks. EditSR is one of the most stable models across these benchmarks, as its Symbolic solution rate usually stays close to its Accuracy solution rate on the same benchmark. This phenomenon indicates that EditSR primarily improves structural recovery, as expected, since the Rectifier always edits an incorrect expression toward the target symbolic structure. Some baselines still achieve strong results on particular metrics and benchmarks. On the Accuracy solution rate, TPSR performs best on Jin and Keijzer, and also stays near the top on Nguyen. uDSR also remains close to EditSR on Constant, Koza, and Nguyen. On the Symbolic solution rate, uDSR is the best baseline on Constant and matches EditSR on Koza, while SR4MDL and RILS-ROLS give relatively strong results on Jin and Korns. However, for most baselines, the Symbolic solution rate falls clearly below the Accuracy solution rate, indicating that a good numerical fit does not translate into equally reliable symbolic structure recovery. By contrast, EditSR more consistently keeps the two metrics close across benchmarks.

To measure inference efficiency, we report the test time of models on the standard benchmarks. As shown in Table~\ref{tab:test_time}, EditSR has the lowest test time in this comparison, which is consistent with the design choice of shifting most rectification effort to pretraining. RILS-ROLS is the next most efficient baseline, whereas search-heavy or repeatedly optimized models require substantially more time before returning a final expression. TPSR, as another post-hoc rectification model, incurs much longer runtime than EditSR. This result highlights a practical limitation of traditional rectification strategies based on restarting global search, i.e., they dilute the efficiency advantage of the neural model, which we will discuss further in the ablation studies.

\begin{table}[h]
	\centering
	\renewcommand{\arraystretch}{1.3}
	\caption{Mean test time on standard benchmarks.}
	\label{tab:test_time}
	
	\begin{tabularx}{\columnwidth}{@{} >{\centering\arraybackslash}X >{\centering\arraybackslash}X @{}}
		\toprule
		\textbf{Model} & \textbf{Test Time (s)} \\
		\midrule
		EditSR    & 17.03 \\
		uDSR       & 77.99 \\
		SR4MDL   & 765.03 \\
		ParFam       & 367.18 \\
		RILS-ROLS       & 26.46 \\
		TPSR  & 174.24 \\
		\bottomrule
	\end{tabularx}
\end{table}

\begin{figure}[h]
	\centering
	\begin{subfigure}[t]{0.23\textwidth}
		\centering
		\includegraphics[width=\linewidth]{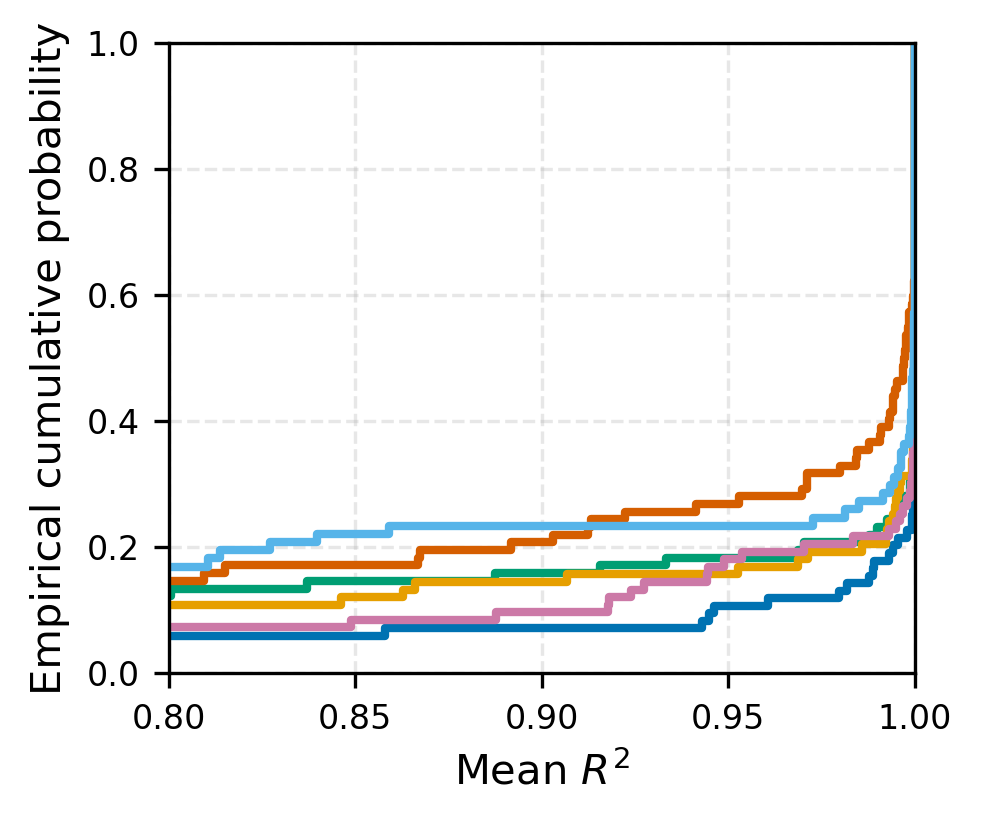}
		\caption{Noise level $0$.}
		\label{fig:stand_noise_ecdf_0}
	\end{subfigure}
	\hfill
		\begin{subfigure}[t]{0.23\textwidth}
		\centering
		\includegraphics[width=\linewidth]{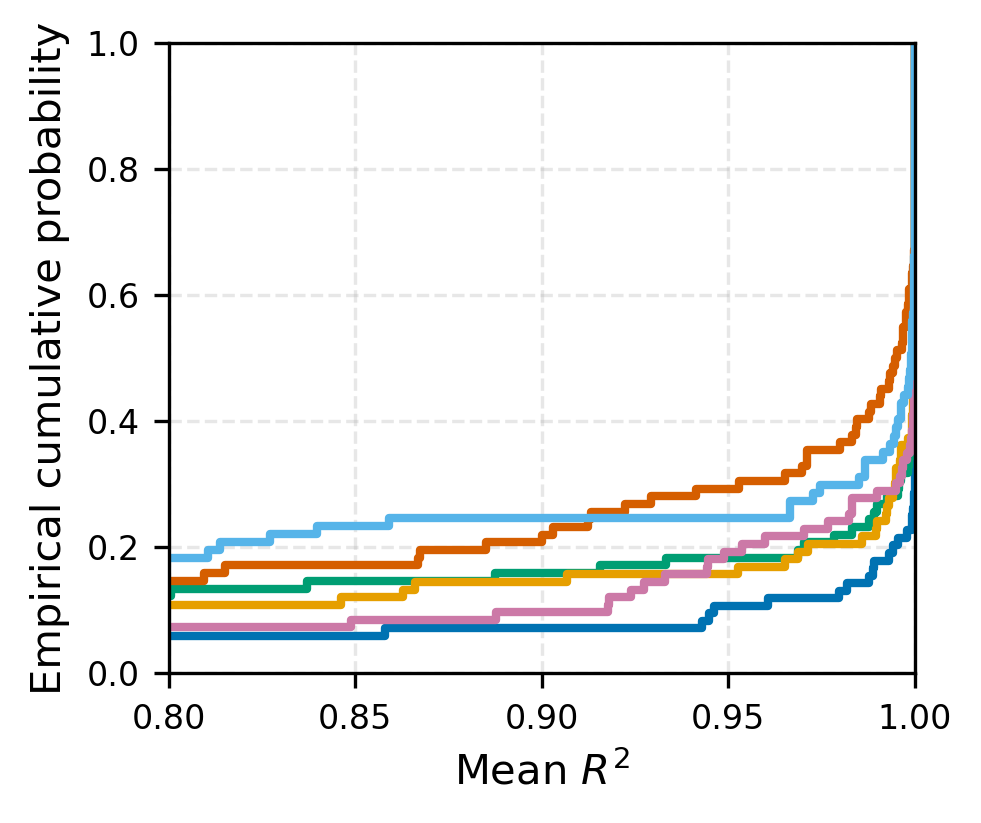}
		\caption{Noise level $0.001$.}
		\label{fig:stand_noise_ecdf_0001}
	\end{subfigure}
	\hfill
	\begin{subfigure}[t]{0.23\textwidth}
		\centering
		\includegraphics[width=\linewidth]{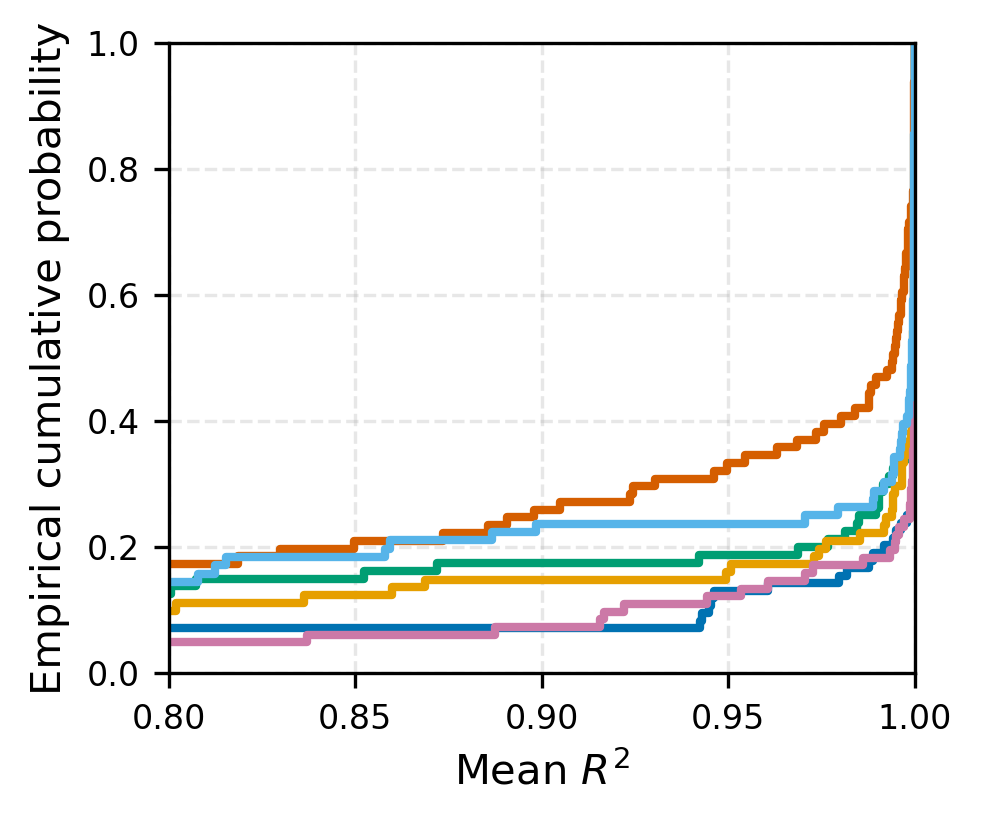}
		\caption{Noise level $0.01$.}
		\label{fig:stand_noise_ecdf_001}
	\end{subfigure}
	\hfill
	\begin{subfigure}[t]{0.23\textwidth}
		\centering
		\includegraphics[width=\linewidth]{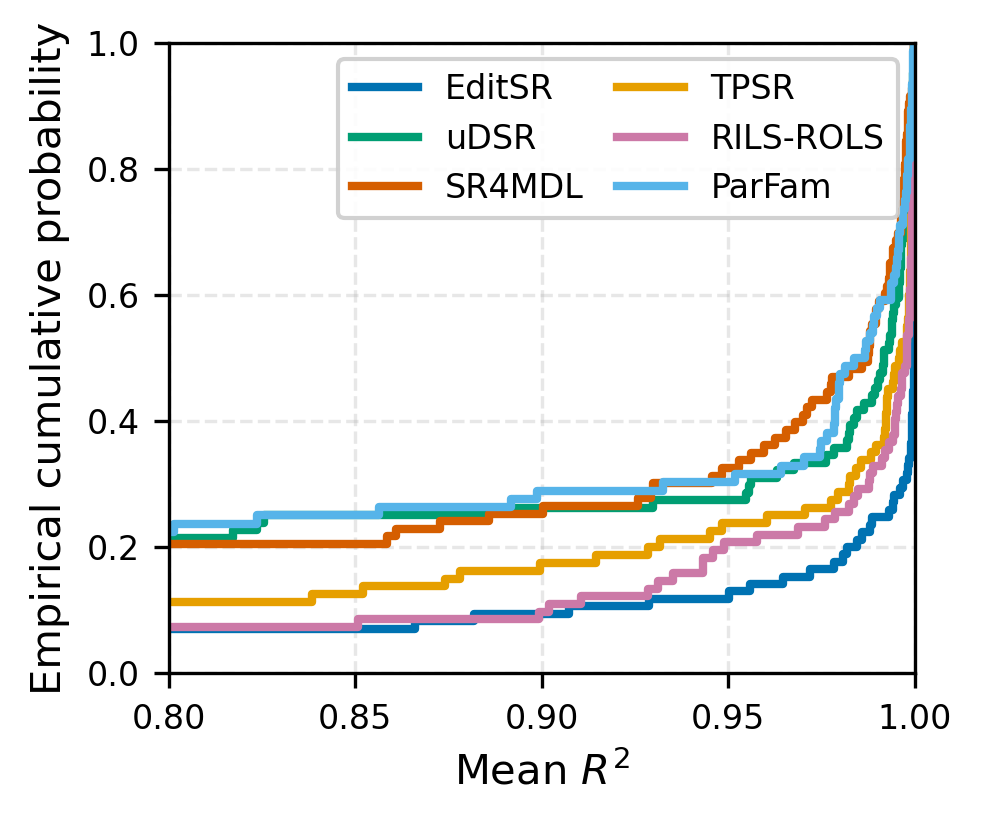}
		\caption{Noise level $0.1$.}
		\label{fig:stand_noise_ecdf_01}
	\end{subfigure}
	\caption{\textbf{ECDF of mean $R^2$ on standard benchmarks.} The results show the mean over 10 runs under Gaussian noise levels of 0, 0.001, 0.01, and 0.1, respectively.}
	\label{fig:stand_noise_ecdf}
\end{figure}

Noise robustness is a critical metric for assessing whether a symbolic regression model is suitable for real-world problems, as observed data, constrained by factors such as sensor precision and statistical errors, are rarely noise-free. Fig.~\ref{fig:stand_noise_ecdf} shows the empirical cumulative distribution function (ECDF) of $R^2$ at the 0, 0.001, 0.01, and 0.1 noise levels. Each curve shows the proportion of problems with $R^2$ below a given threshold. As expected, the curves of all models move away from the high-$R^2$ region as noise increases. Even so, EditSR remains among the more right-shifted distributions at all noise levels, which means that a relatively large proportion of problems still reaches high $R^2$. SR4MDL and ParFam place a visible proportion of problems near $R^2\!\approx\!1$ at the 0.001 noise level, but their curves rise earlier as noise increases. uDSR, TPSR, and RILS-ROLS degrade less sharply than several other baselines, but they still show a clearer leftward shift than EditSR, especially at the 0.1 noise level.

In real-world scientific discovery scenarios, observational datasets often contain measured variables unrelated to the governing equations, which poses a practical challenge for symbolic regression. Therefore, we evaluate the models' robustness to irrelevant variables on the standard benchmarks. For each problem, we append $k\in\{1,2,3\}$ distractor variables to both the training and test sets. Each distractor is sampled independently, but it reuses the marginal range of a randomly selected genuine variable from the same problem, so the perturbation remains distributionally plausible. We pool all standard benchmarks except Korns, where all models perform poorly, and report the aggregate results. Table~\ref{tab:distractor_results_redetected} reports three metrics: Accuracy solution rate, Symbolic solution rate, and distractor usage rate. The distractor usage rate measures whether the predicted expression contains at least one variable that does not appear in the target expression. As the number of distractors increases, the task overall becomes harder. TPSR and uDSR achieve strong Accuracy solution rates, especially for $k=2$ and $k=3$, but they also rely more heavily on distractors, which is accompanied by a clearer degradation in their Symbolic solution rates. By contrast, EditSR maintains the strongest Symbolic solution rate across all settings while using substantially fewer distractors than the baselines, indicating that its robustness relies less on absorbing spurious inputs into the expression. Several baselines remain competitive in terms of the Accuracy solution rate, so the distractors do not reduce the task to random search. However, EditSR more consistently preserves symbolic structure recovery under feature perturbations. In scientific discovery tasks, this behavior is desirable because a high-quality symbolic model should remain dependent on the relevant variables that actually support the governing relation rather than merely matching the observed responses numerically.

\begin{table}[h]
	\centering
	\caption{Distractor robustness results. For each model and distractor count, we report the mean and standard deviation across 10 runs of Accuracy solution rate (ASR), Symbolic solution rate (SSR), and distractor usage rate (DUR). The best and second-best results in each column are marked in \textbf{bold} and \underline{underlined}, respectively.}
	\label{tab:distractor_results_redetected}
	\setlength{\tabcolsep}{3pt}
	\renewcommand{\arraystretch}{1.3}
	\scriptsize
	\resizebox{\columnwidth}{!}{
		\begin{tabular}{l|l|c c c}
			\toprule
			\multicolumn{1}{c|}{\multirow{2}{*}{Model}} & \multicolumn{1}{c|}{\multirow{2}{*}{Metric}} & \multicolumn{3}{c}{No. distractors} \\
			\cmidrule(lr){3-5}
			& & 1 & 2 & 3 \\
			\midrule
			
			\multirow{3}{*}{EditSR} 
			& ASR (\%) & \textbf{77.94 $\pm$ {\scriptsize 3.89}} & 72.06 $\pm$ {\scriptsize 2.55} & 65.69 $\pm$ {\scriptsize 2.25} \\
			& DUR (\%) & \textbf{2.45 $\pm$ {\scriptsize 3.06}} & \textbf{4.41 $\pm$ {\scriptsize 2.94}} & \textbf{13.73 $\pm$ {\scriptsize 2.25}} \\
			& SSR (\%) & \textbf{61.76 $\pm$ {\scriptsize 5.88}} & \textbf{55.88 $\pm$ {\scriptsize 3.89}} & \textbf{48.53 $\pm$ {\scriptsize 4.41}} \\
			\midrule
			
			\multirow{3}{*}{uDSR} 
			& ASR (\%) & 71.17 $\pm$ {\scriptsize 0.78} & 70.27 $\pm$ {\scriptsize 1.35} & 64.86 $\pm$ {\scriptsize 2.70} \\
			& DUR (\%) & 29.19 $\pm$ {\scriptsize 1.35} & 38.20 $\pm$ {\scriptsize 2.81} & 45.41 $\pm$ {\scriptsize 1.35} \\
			& SSR (\%) & \underline{58.56 $\pm$ {\scriptsize 3.12}} & \underline{51.80 $\pm$ {\scriptsize 2.06}} & \underline{46.40 $\pm$ {\scriptsize 2.06}} \\
			\midrule
			
			\multirow{3}{*}{SR4MDL} 
			& ASR (\%) & 66.23 $\pm$ {\scriptsize 1.89} & 66.07 $\pm$ {\scriptsize 2.09} & 60.59 $\pm$ {\scriptsize 2.09} \\
			& DUR (\%) & 23.87 $\pm$ {\scriptsize 5.32} & 44.29 $\pm$ {\scriptsize 3.45} & 64.38 $\pm$ {\scriptsize 3.62} \\
			& SSR (\%) & 20.17 $\pm$ {\scriptsize 2.75} & 17.35 $\pm$ {\scriptsize 0.79} & 16.44 $\pm$ {\scriptsize 1.37} \\
			\midrule
			
			\multirow{3}{*}{TPSR} 
			& ASR (\%) & 73.83 $\pm$ {\scriptsize 2.81} & \textbf{76.13 $\pm$ {\scriptsize 3.90}} & \textbf{71.62 $\pm$ {\scriptsize 1.35}} \\
			& DUR (\%) & 70.50 $\pm$ {\scriptsize 0.78} & 68.65 $\pm$ {\scriptsize 0.00} & 78.20 $\pm$ {\scriptsize 0.78} \\
			& SSR (\%) & 9.91 $\pm$ {\scriptsize 4.75} & 13.06 $\pm$ {\scriptsize 2.06} & 13.06 $\pm$ {\scriptsize 0.78} \\
			\midrule
			
			\multirow{3}{*}{RILS-ROLS} 
			& ASR (\%) & \underline{74.89 $\pm$ {\scriptsize 2.85}} & \underline{72.15 $\pm$ {\scriptsize 2.85}} & \underline{70.32 $\pm$ {\scriptsize 0.79}} \\
			& DUR (\%) & \underline{7.76 $\pm$ {\scriptsize 2.09}} & \underline{8.68 $\pm$ {\scriptsize 4.81}} & \underline{18.76 $\pm$ {\scriptsize 4.40}} \\
			& SSR (\%) & 28.77 $\pm$ {\scriptsize 1.37} & 28.31 $\pm$ {\scriptsize 2.09} & 26.48 $\pm$ {\scriptsize 2.09} \\
			\midrule
			
			\multirow{3}{*}{ParFam} 
			& ASR (\%) & 66.76 $\pm$ {\scriptsize 1.35} & 62.40 $\pm$ {\scriptsize 5.26} & 55.01 $\pm$ {\scriptsize 5.57} \\
			& DUR (\%) & 44.14 $\pm$ {\scriptsize 0.78} & 22.14 $\pm$ {\scriptsize 5.68} & 55.28 $\pm$ {\scriptsize 3.90} \\
			& SSR (\%) & 13.66 $\pm$ {\scriptsize 1.56} & 7.01 $\pm$ {\scriptsize 2.61} & 6.38 $\pm$ {\scriptsize 0.12} \\
			\bottomrule
		\end{tabular}
	}
\end{table}

\subsection{Results on SRBench 1.0}

In this section, we report the results on SRBench 1.0, including Feynman and ODE-Strogatz benchmarks, to analyze whether the symbolic structure recovery advantage observed for EditSR on standard benchmarks extends to physical equations. Each experiment is repeated 10 times. In addition to the baselines defined earlier, we include several widely used models reported in SRBench~1.0 so that the comparison covers a broader set of established symbolic regression approaches. For a fair comparison, all models are evaluated under the same dataset splitting scheme. 

\begin{figure}[h]
	\centering
	\begin{subfigure}[t]{0.45\textwidth}
		\centering
		\includegraphics[width=\linewidth]{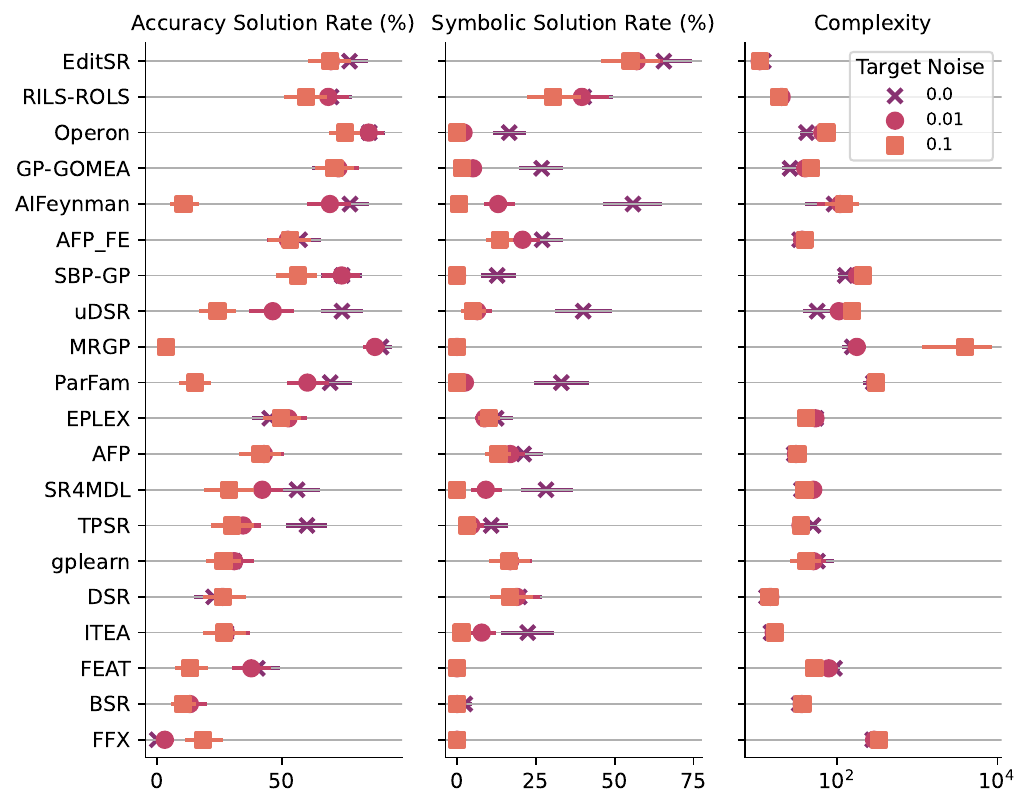}
		\caption{Feynman benchmark.}
		\label{fig:srbench1_feynman}
	\end{subfigure}
	\hfill
	\begin{subfigure}[t]{0.45\textwidth}
		\centering
		\includegraphics[width=\linewidth]{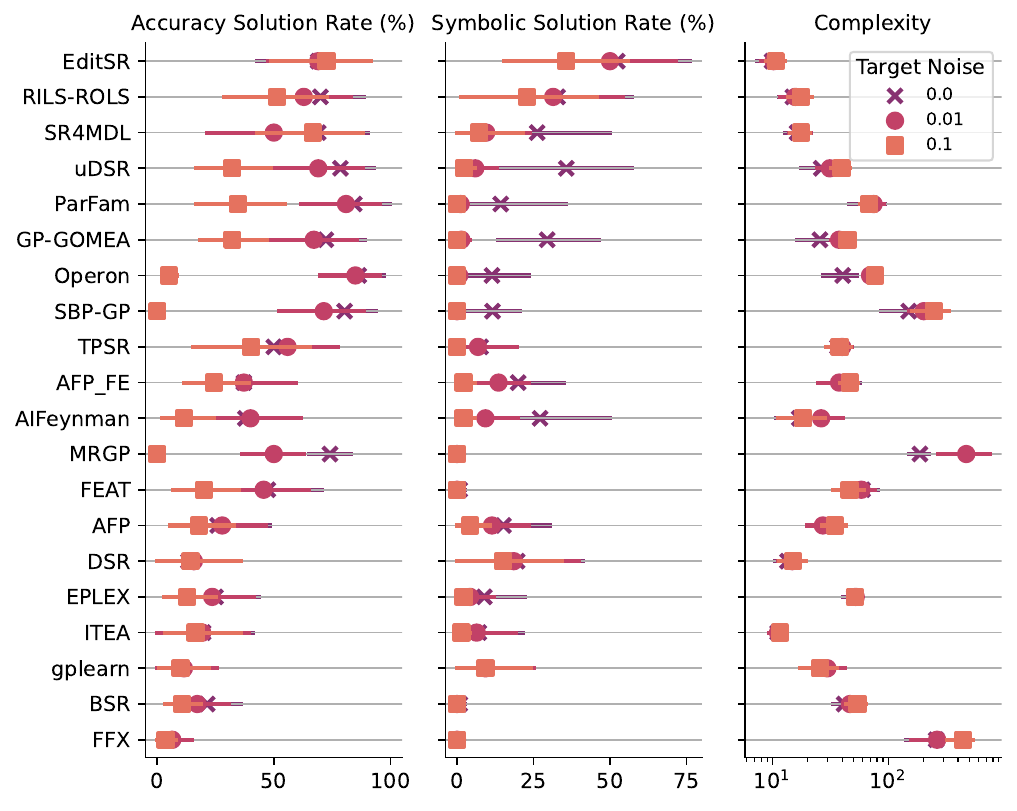}
		\caption{ODE-Strogatz benchmark.}
		\label{fig:srbench1_strogatz}
	\end{subfigure}
	\caption{\textbf{Results on Feynman and ODE-Strogatz benchmarks.} Models are ordered from top to bottom by their Accuracy solution rate and Symbolic solution rate. Each point summarizes the mean result of 10 runs, and the error bars denote 95\% bootstrap confidence intervals.}
	\label{fig:srbench1_results}
\end{figure}

The results on the Feynman benchmark under three noise levels are reported in Fig.~\ref{fig:srbench1_feynman}. For the Accuracy solution rate, EditSR remains close to 80\% at all noise levels, and shows only a limited drop at higher noise levels. Operon and RILS-ROLS are also strong on the Accuracy solution rate, demonstrating similar noise robustness. By contrast, AIFeynman, MRGP, uDSR, and ParFam decline more clearly as the noise level increases. The same phenomenon is observed for the Symbolic solution rate. EditSR exceeds 50\% consistently across noise levels. RILS-ROLS is the closest baseline, but it still falls short of EditSR. Operon, GP-GOMEA, AIFeynman, and uDSR achieve outstanding Symbolic solution rates at the 0 noise level, but decrease more as noise increases. In addition, the Complexity results show that EditSR does not achieve the above advantage by increasing expression complexity, but instead remains within a relatively low complexity range. Overall, EditSR maintains a favorable balance among the three reported metrics on the Feynman benchmark. At the same time, its performance degrades only modestly as the noise level increases, which may be related to the noise perturbations introduced during training. Such behavior is particularly relevant in scientific discovery scenarios.

\begin{figure*}[t]
	\centering
	
	\makebox[\textwidth][c]{%
		\subfloat[$R^2$ distribution on Black-box benchmark\label{fig:srbench_black_r2}]{
			\includegraphics[width=0.42\linewidth]{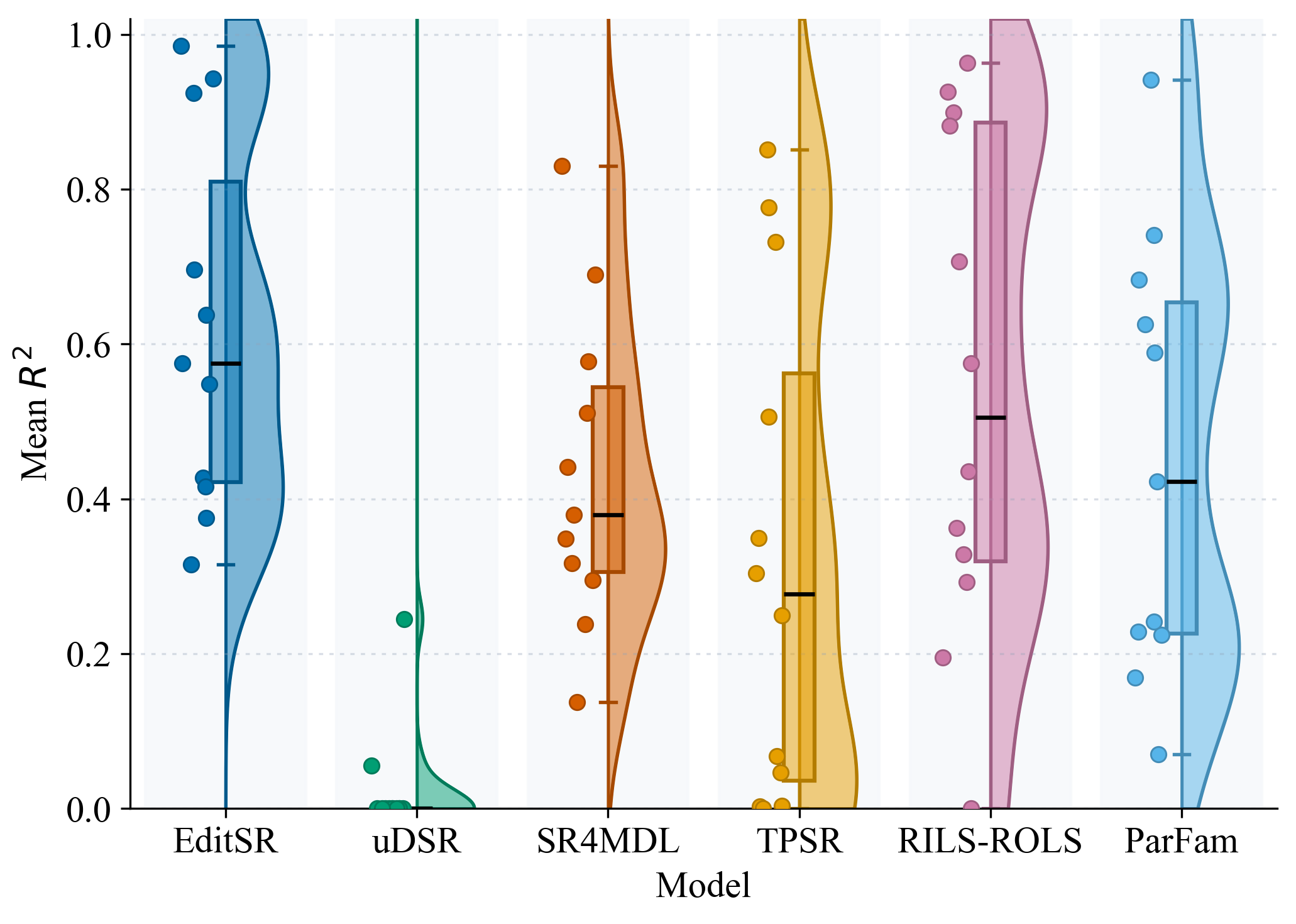}
		}
		\hspace{0.03\textwidth}
		\subfloat[Complexity distribution on Black-box benchmark\label{fig:srbench_black_complexity}]{
			\includegraphics[width=0.42\linewidth]{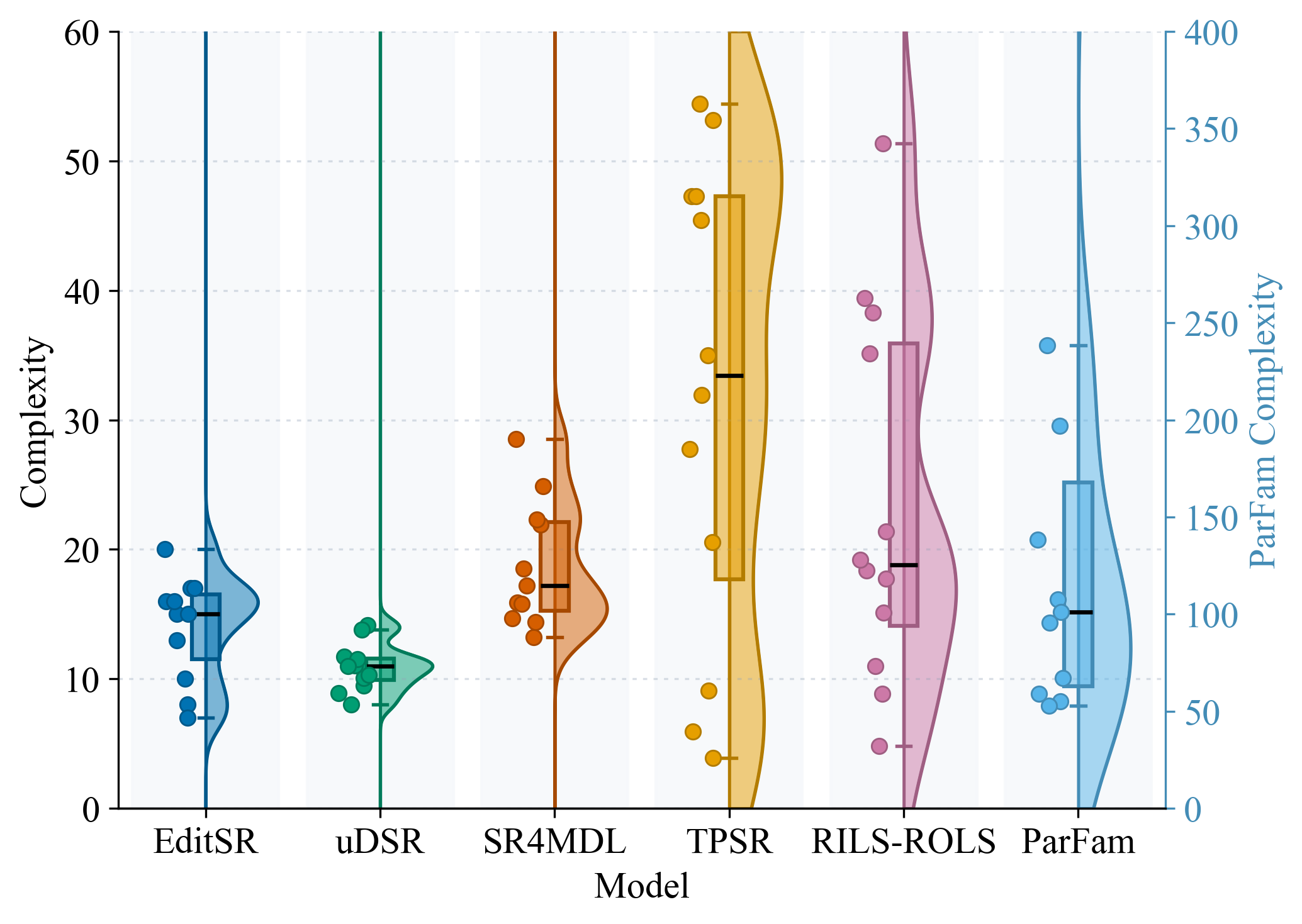}
		}
	}
	
	\vspace{1em}
	
	\makebox[\textwidth][c]{%
		\subfloat[$R^2$ distribution on Phenomenological \& first-principles benchmark\label{fig:srbench_firstprinciples_r2}]{
			\includegraphics[width=0.42\linewidth]{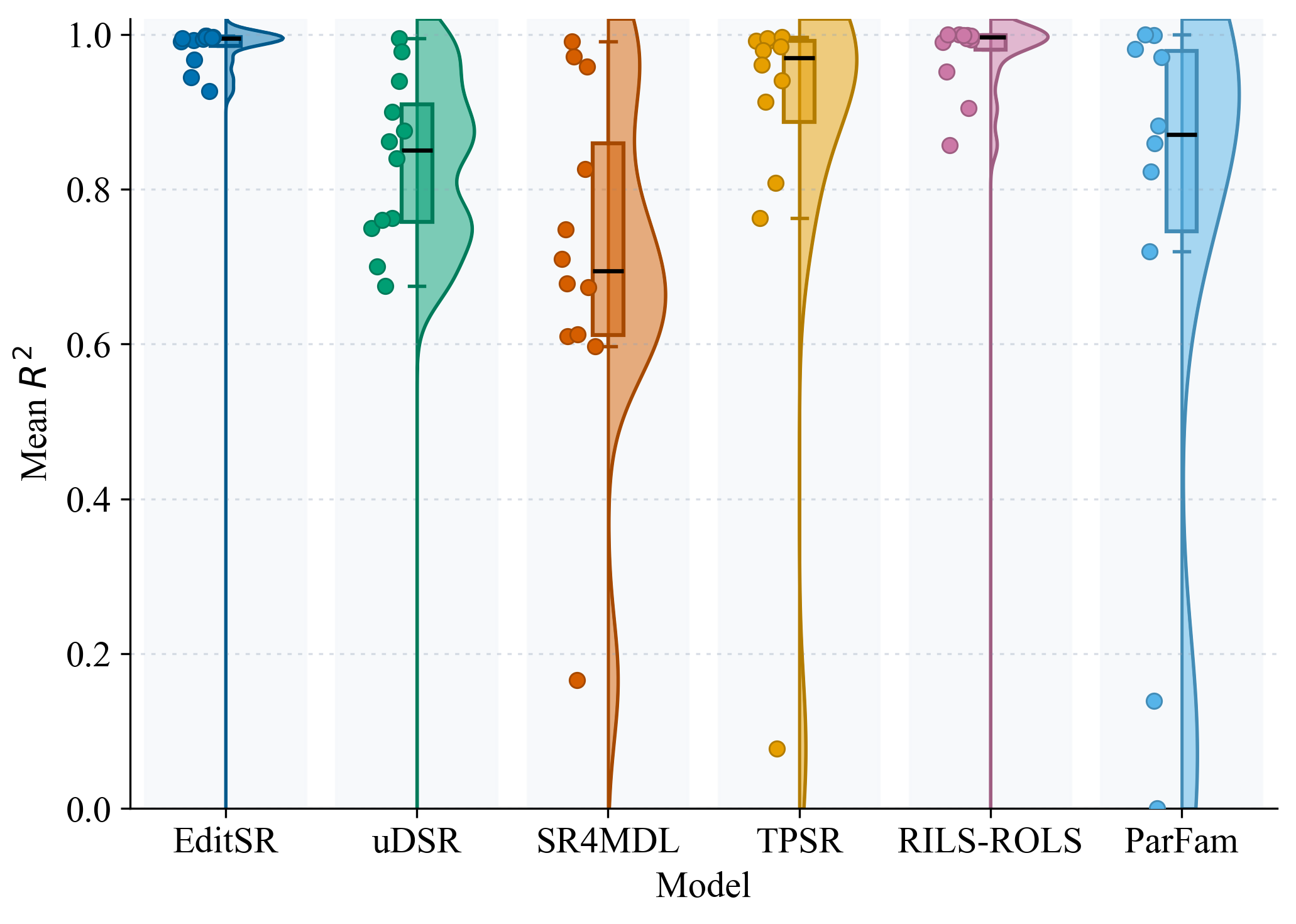}
		}
		\hspace{0.03\textwidth}
		\subfloat[Complexity distribution on Phenomenological \& first-principles benchmark\label{fig:srbench_firstprinciples_complexity}]{
			\includegraphics[width=0.42\linewidth]{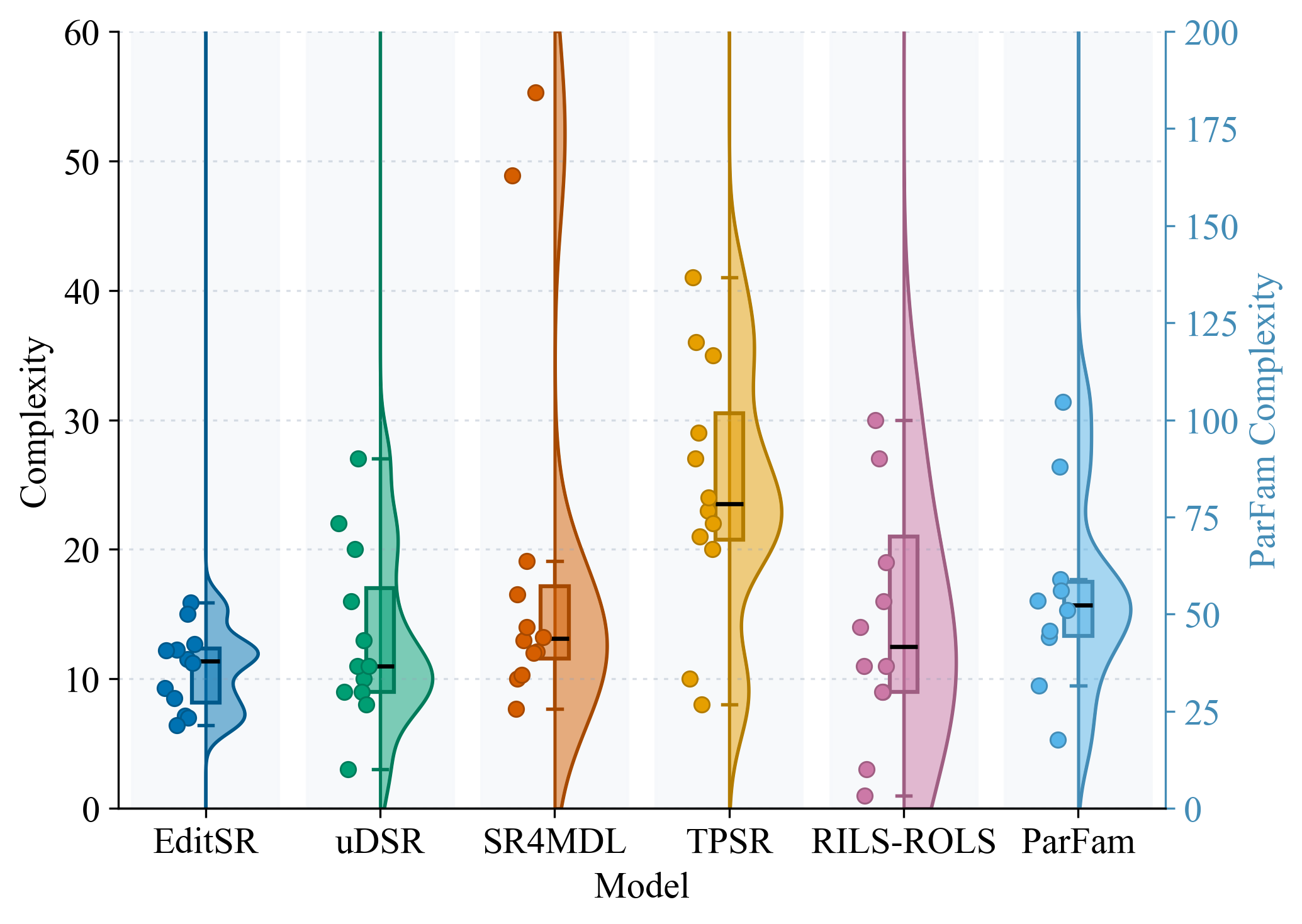}
		}
	}
	
	\caption{\textbf{Mean $R^2$ and Complexity results on Black-box and Phenomenological \& first-principles benchmarks.} Each point corresponds to the mean over 10 runs for each problem. In each rain-cloud plot, the cloud shows the distribution density, the embedded box summarizes the median and interquartile range, and the points show the individual problem results.}
	\label{fig:srbench_black_firstprinciples}
\end{figure*}

Overall, the results across these two benchmarks suggest that EditSR maintains a favorable balance among the Accuracy solution rate, Symbolic solution rate, and Complexity. Although some baselines outperform EditSR on specific metrics or noise levels, they usually incur one of three trade-offs: they produce substantially more complex expressions, their Symbolic solution rate falls more sharply than their Accuracy solution rate as noise increases, or their confidence intervals are visibly wider. By contrast, EditSR remains competitive across all three metrics and shows a comparatively stable profile across noise levels. This balance is desirable in scientific discovery scenarios because a useful expression should be accurate, structurally reasonable, simple, and stable, rather than strong on only one metric.

The results on the ODE-Strogatz benchmark (Fig.~\ref{fig:srbench1_strogatz}) are lower than those on the Feynman benchmark for most models, including EditSR. This result may be partly attributable to the fact that the ODE-Strogatz benchmark is built from dynamical systems and uses narrower sampling ranges. Even so, EditSR remains among the strongest models on both Accuracy solution rate and Symbolic solution rate. On the Accuracy solution rate, several baselines, including RILS-ROLS, uDSR, GP-GOMEA, SR4MDL, Operon, and SBP-GP, are competitive at the 0 noise level, but their results drop more as the noise level increases. In contrast, EditSR changes less across all noise levels and is among the strongest models at the 0.1 noise level. Moreover, the expressions predicted by EditSR remain simple. On the Symbolic solution rate, EditSR also achieves competitive results, with RILS-ROLS as the closest competing baseline. uDSR, GP-GOMEA, and SR4MDL all outperform most baselines at the 0 noise level, but they exhibit weaker robustness and degrade rapidly in noise scenarios. Taken together, these results suggest that EditSR remains relatively robust on this benchmark family. Even when the dataset is sparse and more challenging than the training setting, its performance remains comparatively stable.

\subsection{Results on SRBench 2.0}

In this section, we report results on the Phenomenological \& first-principles and Black-box benchmarks.  We report two metrics evaluated on SRBench 2.0, $R^2$ and Complexity, to facilitate a fair comparison. The distributions shown below are based on the mean over 10 runs. For high-dimensional problems, we adopt the feature selection strategy suggested in SRBench 2.0 and compress the dimensionality to three.

As shown in Figs.~\ref{fig:srbench_black_r2} and~\ref{fig:srbench_black_complexity}, EditSR exhibits one of the strongest trade-offs between $R^2$ and Complexity on the Black-box benchmark. In the mean $R^2$ distribution, EditSR results fall in the mid-to-high range, with many problems concentrated around 0.4--1.0. RILS-ROLS reaches similarly strong, and sometimes higher, $R^2$ on some problems, but its Complexity distribution is less concentrated and more often extends to complex expressions. By contrast, the Complexity of EditSR stays clustered in a relatively low range. Overall, the Black-box benchmark remains challenging, as none of the models show an $R^2$ distribution concentrated near perfection. However, EditSR more consistently avoids the combination of very low mean $R^2$ and very high Complexity, which makes its overall trade-off comparatively favorable on this benchmark.

Figs.~\ref{fig:srbench_firstprinciples_r2} and~\ref{fig:srbench_firstprinciples_complexity} show that, on the Phenomenological \& first-principles benchmark, EditSR can generate simple expressions for most problems while achieving a robust $R^2$ distribution. Its mean $R^2$ values are concentrated in the upper range, with most problems above 0.95. RILS-ROLS and TPSR reach higher mean $R^2$ on some problems, but they often generate longer expressions or exhibit significant performance fluctuations across different problems. Overall, this benchmark remains challenging for all models because the datasets are noisy and often extremely sparse. Under this scenario, several candidate expressions can remain numerically plausible simultaneously, thereby weakening the signal for local rectification.

\begin{figure}[h]
	\centering
	\begin{subfigure}[t]{0.23\textwidth}
		\centering
		\includegraphics[width=\linewidth]{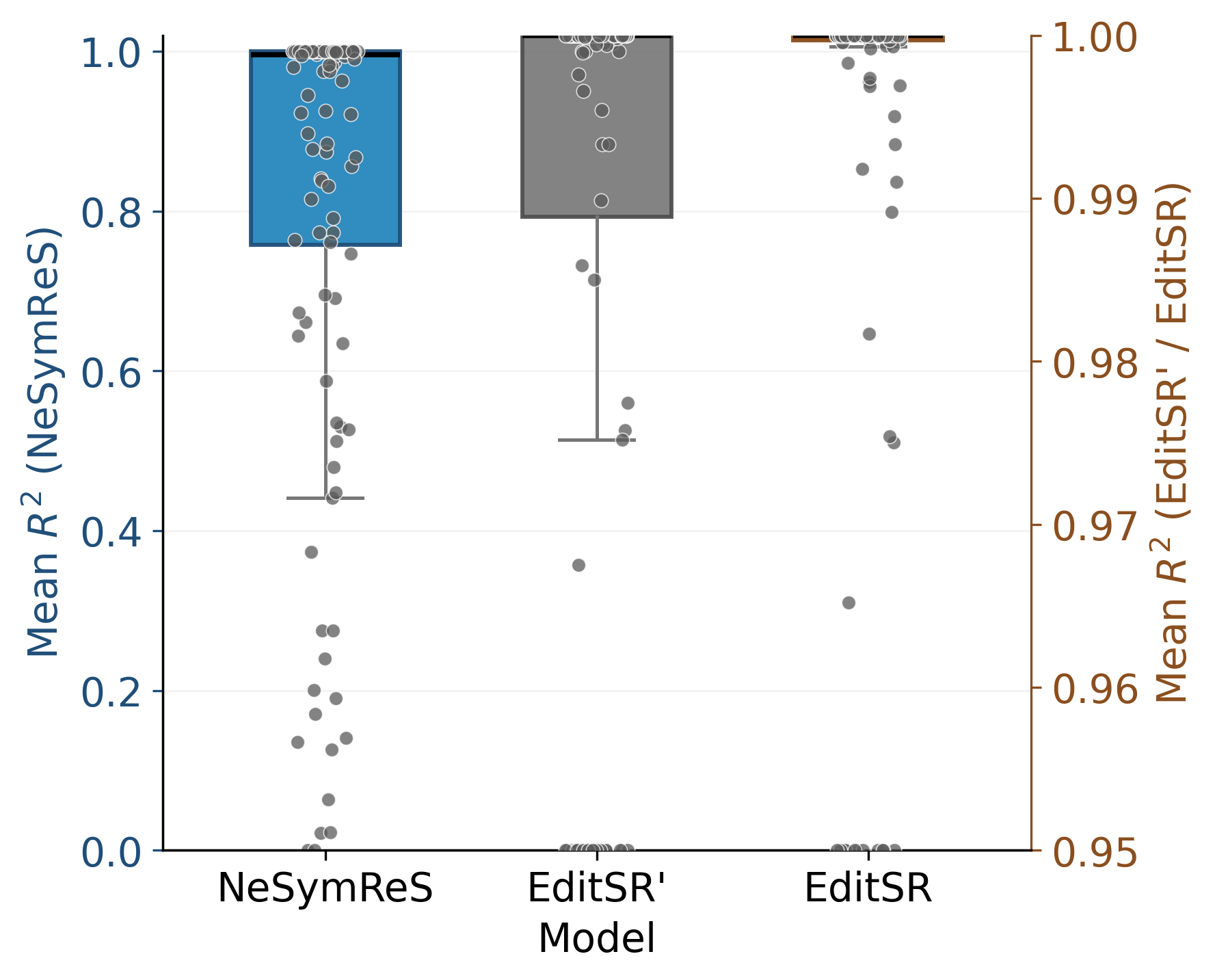}
		\caption{Mean $R^2$ distribution}
		\label{fig:ablation_r2}
	\end{subfigure}
	\hfill
	\begin{subfigure}[t]{0.23\textwidth}
		\centering
		\includegraphics[width=\linewidth]{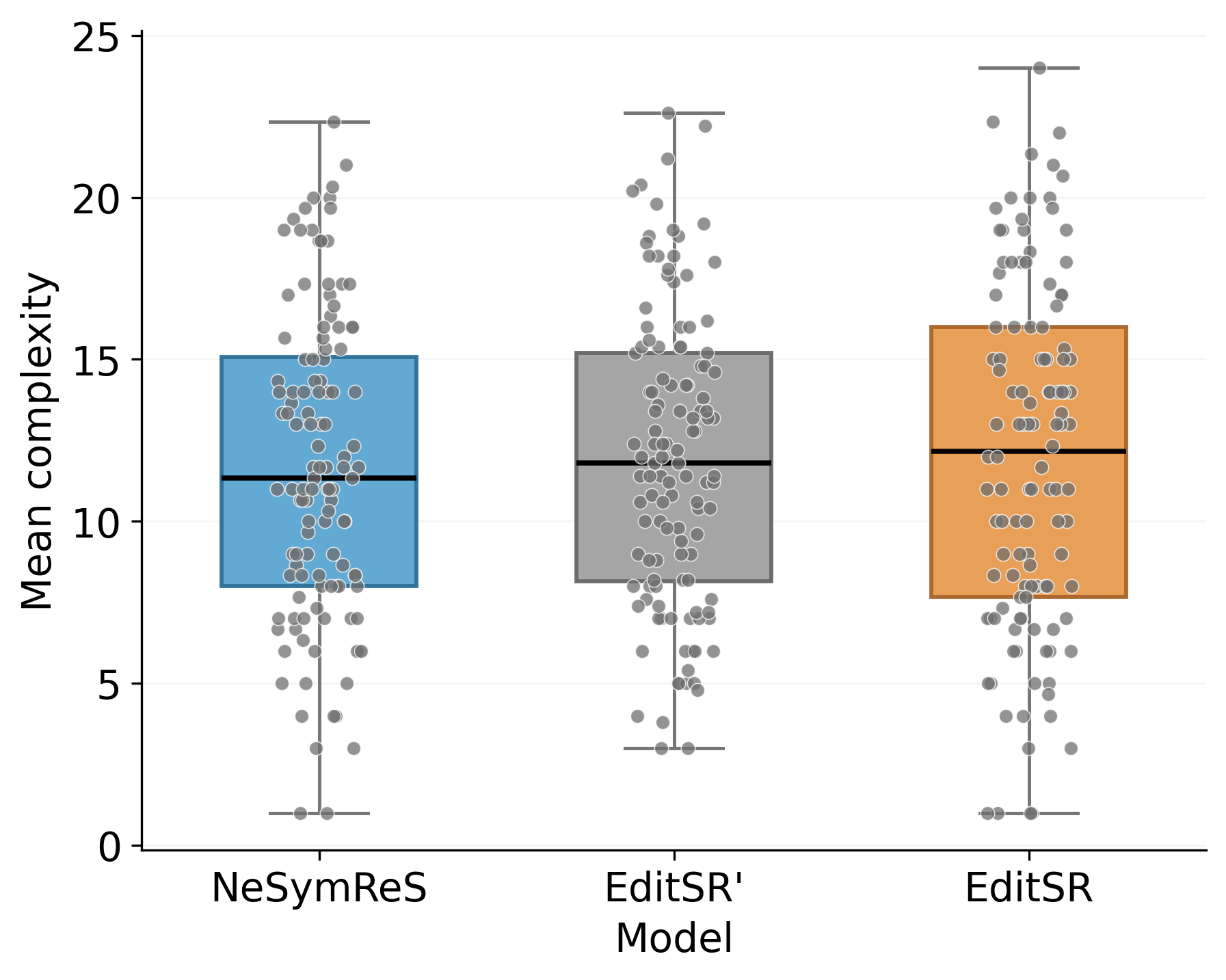}
		\caption{Mean Complexity distribution}
		\label{fig:ablation_complexity}
	\end{subfigure}
	
	\begin{subfigure}[t]{0.23\textwidth}
		\centering
		\includegraphics[width=\linewidth]{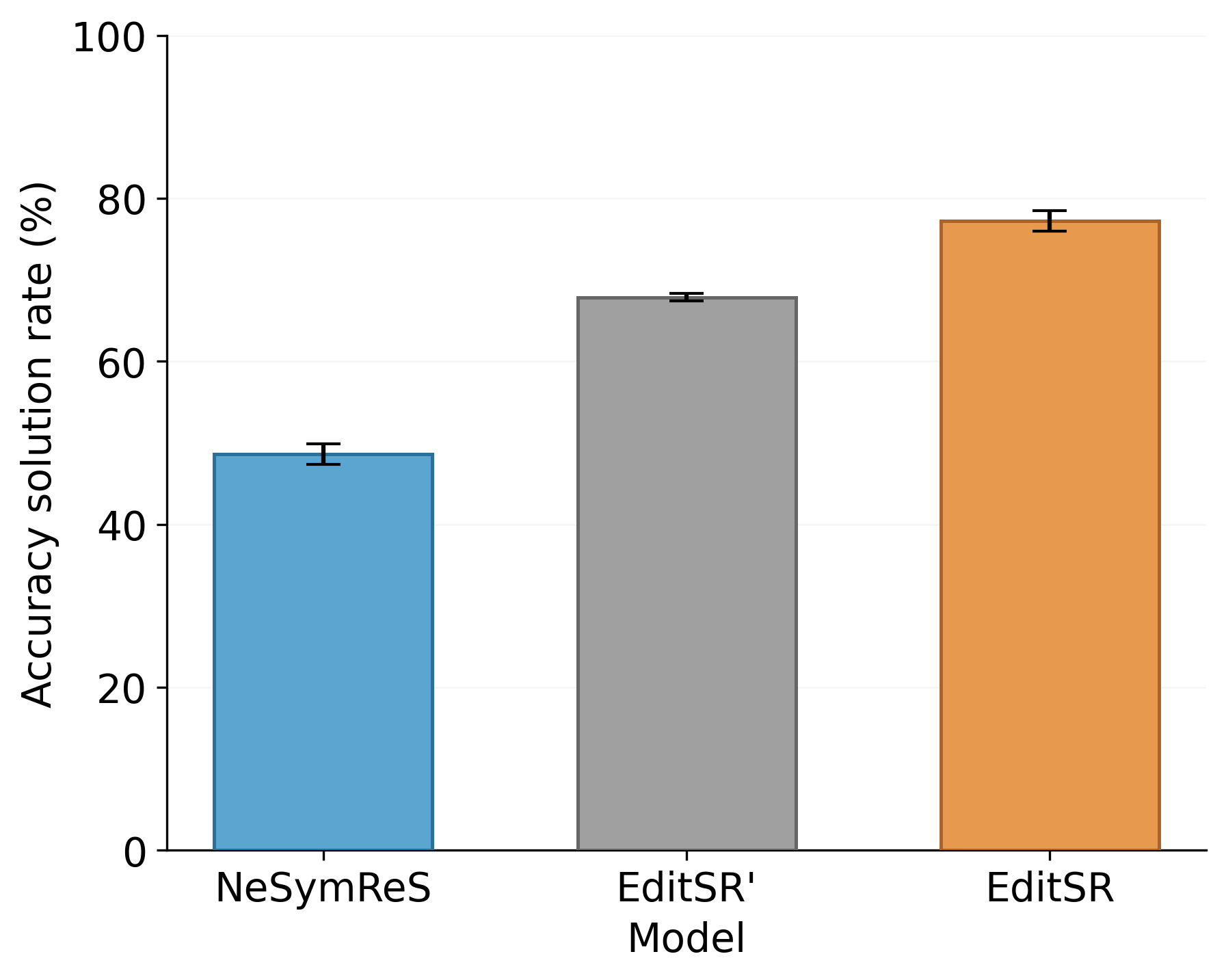}
		\caption{Accuracy solution rate}
		\label{fig:ablation_acc_rate}
	\end{subfigure}
	\hfill
	\begin{subfigure}[t]{0.23\textwidth}
		\centering
		\includegraphics[width=\linewidth]{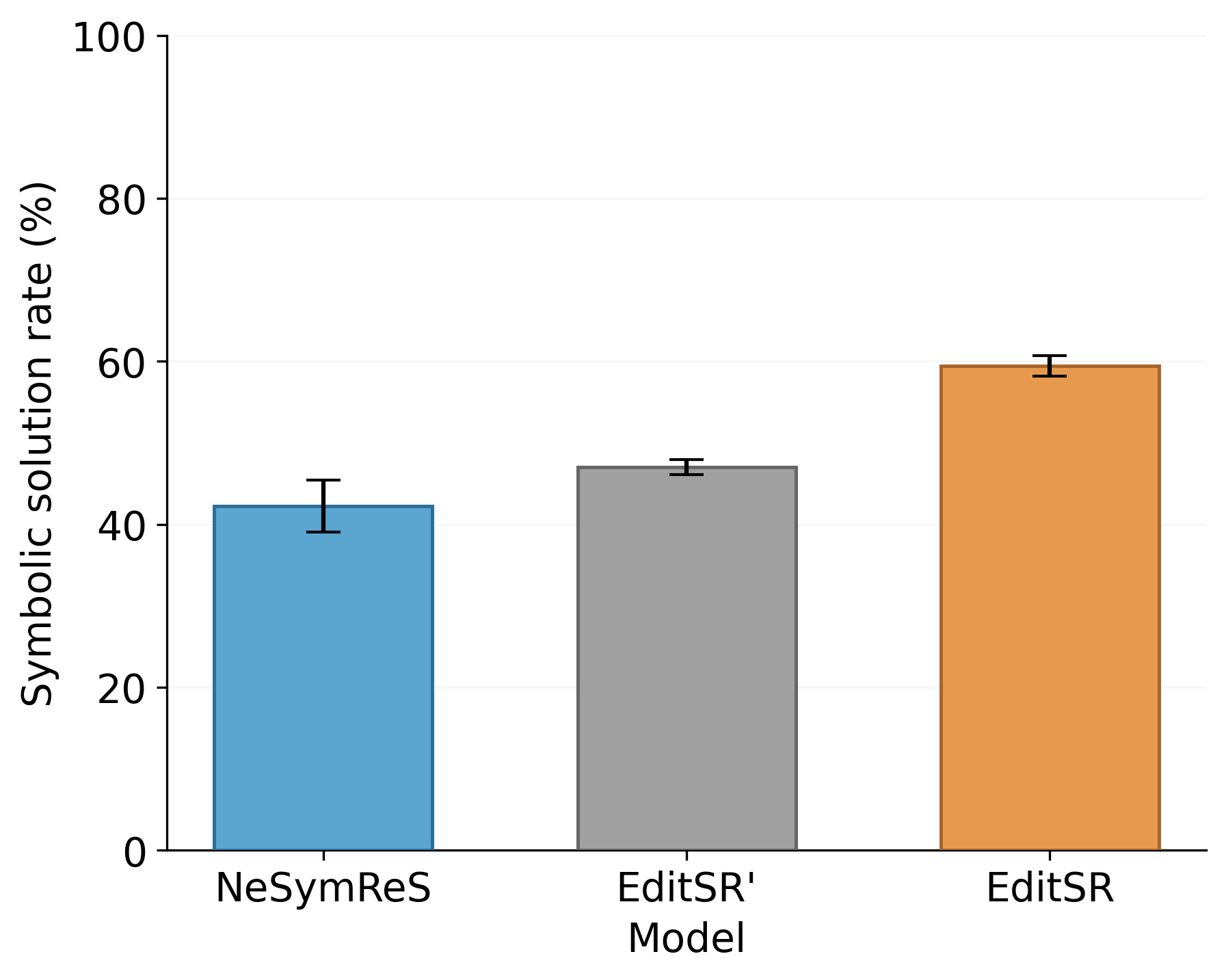}
		\caption{Symbolic solution rate}
		\label{fig:ablation_sym_rate}
	\end{subfigure}
	\caption{\textbf{Ablation results of the Rectifier on the Feynman benchmark.} Results are averaged over 5 runs, with standard deviation shown when applicable.}
	\label{fig:ablation_rectifier}
\end{figure}

\subsection{Ablation studies}

In this section, we analyze the Rectifier's role. Specifically, we investigate the sources of its performance improvements and the scenarios in which they are most significant. Finally, we analyze the Rectifier's sensitivity to the first layer to examine whether it can adapt to different error patterns through fine-tuning. We conduct the ablation experiments on the Feynman benchmark because it spans a broad range of difficulties, from simple to complex expressions and from low-dimensional to high-dimensional scenarios. Each experiment is repeated 5 times. 

\begin{figure}[h]
	\centering
	\begin{subfigure}[t]{0.45\textwidth}
		\centering
		\includegraphics[width=\textwidth]{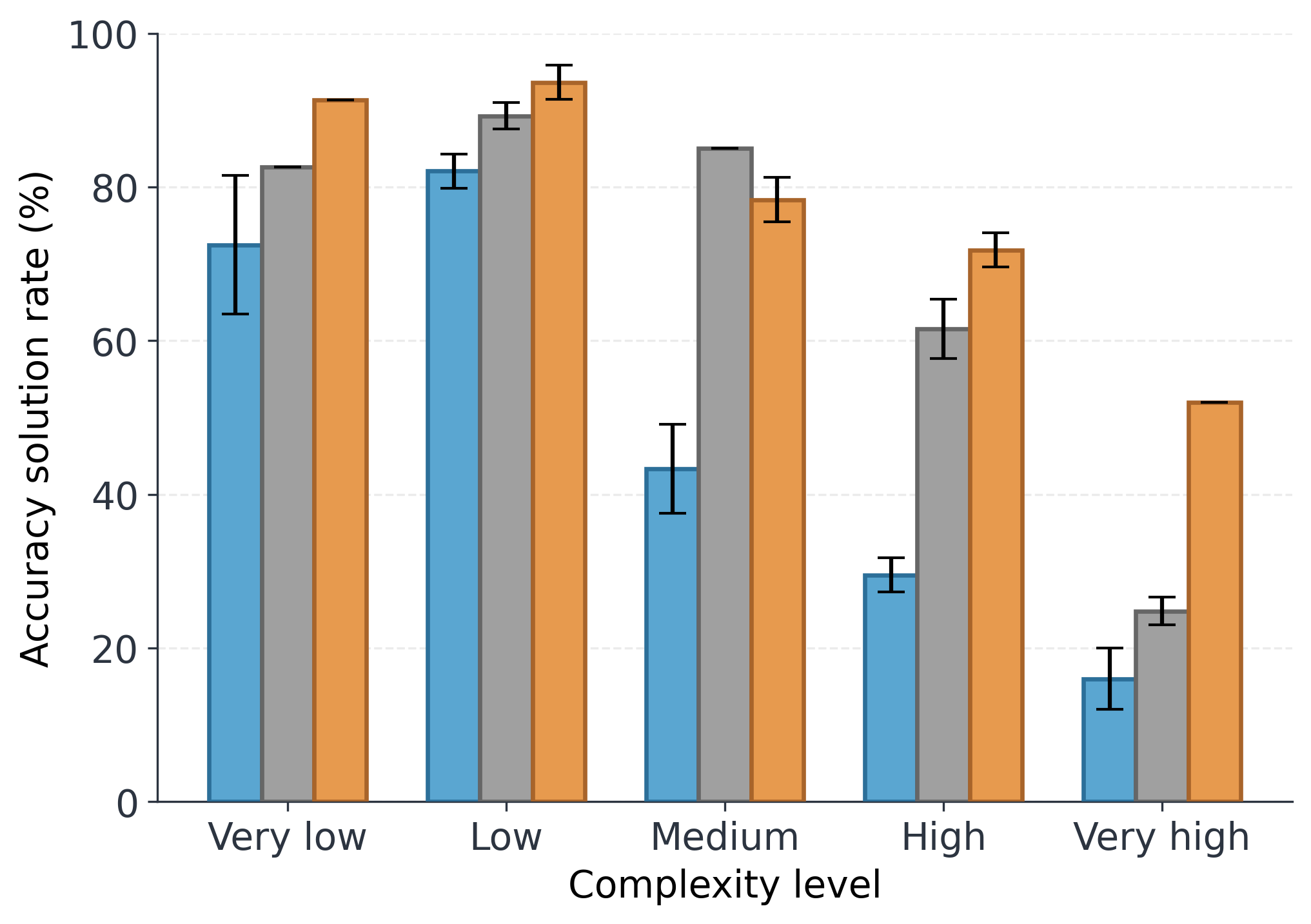}
		\caption{Accuracy solution rate.}
		\label{fig:Complexity_bins_acc}
	\end{subfigure}
	\hfill
	\begin{subfigure}[t]{0.45\textwidth}
		\centering
		\includegraphics[width=\textwidth]{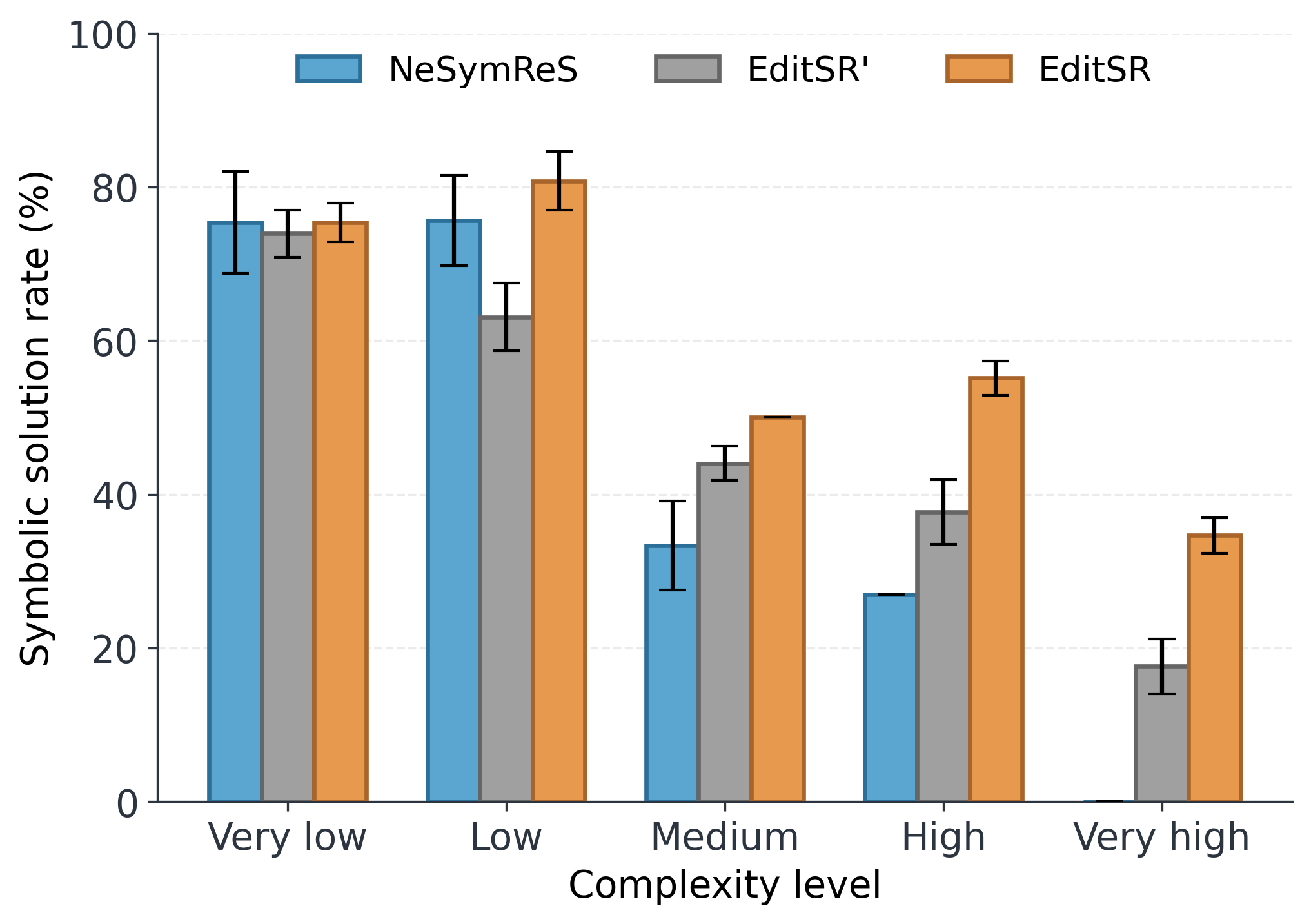}
		\caption{Symbolic solution rate.}
		\label{fig:Complexity_bins_sym}
	\end{subfigure}
	\caption{\textbf{Effect of the Rectifier across complexity levels.} Error bars denote standard deviation over 5 runs. Problems are grouped into 5 buckets according to target expression complexity.}
	\label{fig:Complexity_bins}
\end{figure}
\subsubsection{Rectifier ablation analysis}

In this section, we evaluate the benefits of the proposed Rectifier by comparing the outputs of NeSymReS with those of EditSR (NeSymReS + Rectifier). To analyze the necessity of fine-tuning, two versions of the Rectifier are developed: one fine-tuned and one not fine-tuned. The latter is designated as EditSR$'$. 

As shown in Fig.~\ref{fig:ablation_rectifier}, adding the Rectifier leads to consistent gains across the reported metrics. The largest gains occur in the Accuracy solution rate and the Symbolic solution rate. The $R^2$ and Complexity distributions further suggest that NeSymReS often returns overly simple expressions that fail to satisfy $R^2 \geq 0.999$, indicating that it may stop with structurally incomplete expressions on harder problems. Therefore, the Complexity distribution of EditSR is somewhat higher, which does not mean that EditSR is more inclined to generate complex expressions; rather, it reflects that EditSR can address cases that NeSymReS does not fully generate due to early stopping. Moreover, we observe that NeSymReS fails on many problems, as indicated by $R^2 < 0.95$, whereas EditSR noticeably reduces the number of such failures, suggesting that the Rectifier can help recover a subset of difficult cases.  Importantly, EditSR$'$ already outperforms NeSymReS on multiple metrics, indicating that training on artificially constructed rectification chains is sufficient to endow the Rectifier with a basic ability to repair local structural errors. However, the additional gains of EditSR over EditSR$'$ show that such pretraining alone is not sufficient to match the error patterns encountered during inference fully, so fine-tuning is still indispensable.

To study how the Rectifier's benefit changes with problem difficulty, we divide the benchmark into five groups based on the complexity of the target expressions. Fig.~\ref{fig:Complexity_bins} reports the Accuracy solution rate and Symbolic solution rate for NeSymReS, EditSR$'$ and EditSR in each group. Both metrics decrease as complexity increases for all models, confirming that longer expression generation remains harder. In the very low and low complexity groups, NeSymReS and EditSR are still relatively close. However, from the medium complexity group onward, NeSymReS declines more clearly, even dropping to 0\% Symbolic solution rate in the very high complexity group. In contrast, EditSR maintains higher levels of both Accuracy solution rate and Symbolic solution rate. Overall, the gap between NeSymReS and EditSR becomes progressively larger as complexity increases, which suggests that the Rectifier is particularly useful for generating long expressions. This phenomenon is consistent with the intuition that generating long expressions is more prone to errors during one-pass autoregressive decoding.

\subsubsection{Rectification step budget analysis}

In this section, we examine the internal behavior of the rectification process, focusing on the sensitivity of the rectification step budget $T_{\max}$ and how the expression evolves after being edited. These statistics help clarify how many steps are typically needed during inference.

\begin{figure}[h]
	\centering
	\includegraphics[width=0.9\linewidth]{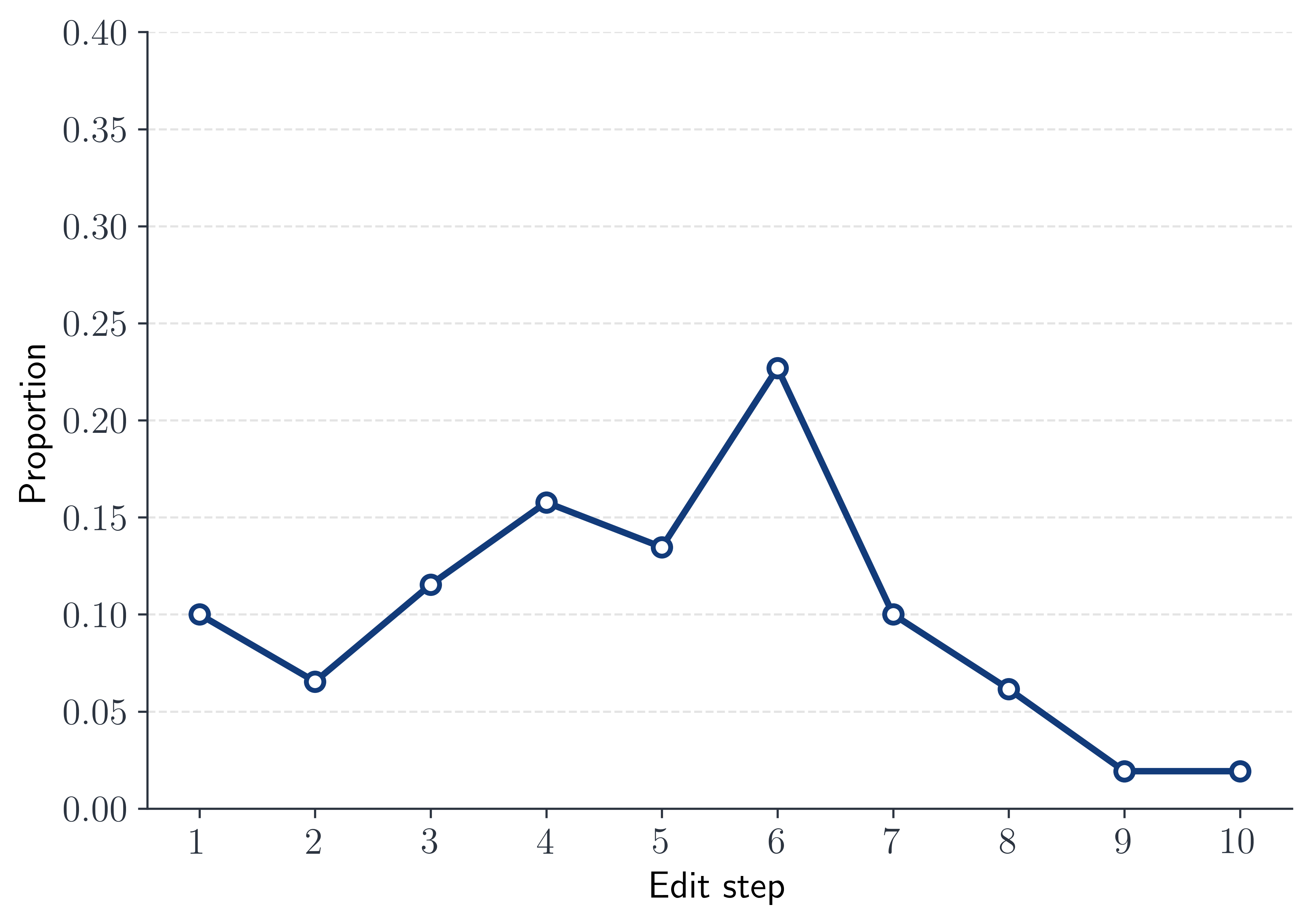}
	\caption{\textbf{Distribution of effective edit steps used by the Rectifier.}
		The results are first averaged across the beams, and then across 5 runs.}
	\label{fig:rectifier_edit_steps_ratio}
\end{figure}

Fig.~\ref{fig:rectifier_edit_steps_ratio} reports the distribution of the number of edit steps required by successful rectification cases. Successful rectification usually requires only 4--6 edit steps, whereas long edit steps are rare. This observation supports two conclusions. First, the expressions predicted by NeSymReS are often not far from the target expressions, which is consistent with our core assumption: even when the prediction is imperfect, it retains partial structural information, so only a few edits are often sufficient, rather than restarting the search process from scratch in every case. Second, the benefit of long rectification chains is limited. As the number of edit steps increases, the rectification process is more likely to deviate from the correct trajectory, thus making it harder to reach the target expression. Therefore, during inference, we typically set the rectification step budget $T_{\max}$ to 10.

\begin{figure}[h]
	\centering
	\includegraphics[width=0.9\linewidth]{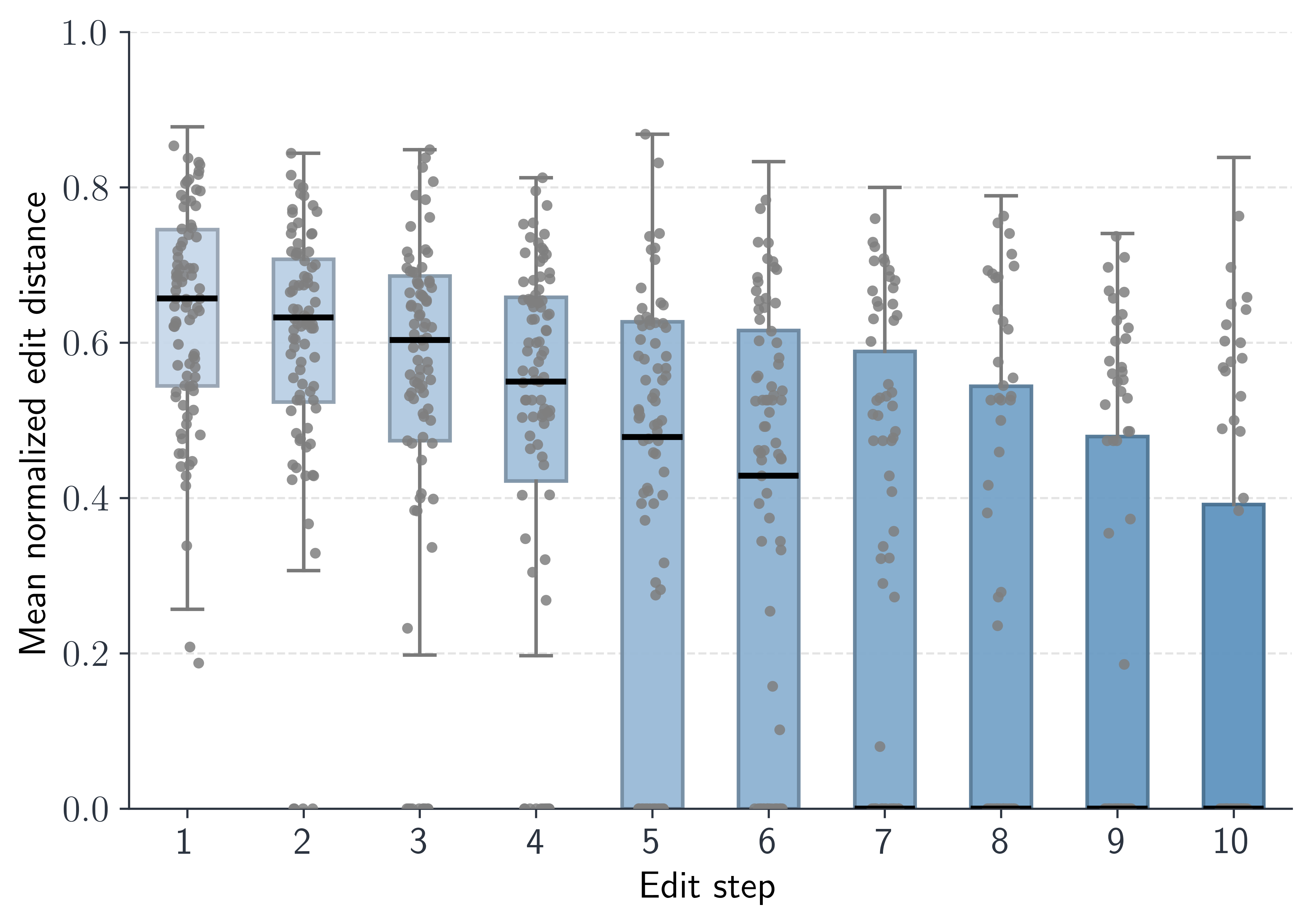}
	\caption{\textbf{Change in normalized edit distance over rectification steps.} Boxplots summarize the distribution of the mean normalized edit distance at each edit step, with 30\% of individual problems randomly sampled and overlaid as points.}
	\label{fig:rectifier_edit_distance}
\end{figure}

In Fig.~\ref{fig:rectifier_edit_distance}, we track how the normalized edit distance \footnote{The normalized edit distance was first proposed by~\citep{matsubara2022srsd} to evaluate the discrepancy between two expressions.} between $f^{(t)}$ and $f^*$ changes with the rectification steps. Here, the normalized edit distance is defined as
\begin{equation}
	ED_{\mathrm{norm}}\!\left(f^{(t)}, f^{*}\right)
	=
	\frac{\mathrm{ED}\!\left(f^{(t)}, f^{*}\right)}
	{\max\!\left(\left|f^{(t)}\right|,\left|f^{*}\right|\right)},
\end{equation}
where $\mathrm{ED}(\cdot,\cdot)$ denotes the token-level edit distance, and $|\cdot|$ denotes the sequence length. The distribution moves downward as the edit step increases, with the median distance dropping from 0.64 at step 1 to nearly 0 at step 7, and the spread narrowing thereafter. Therefore, the aggregate trend is favorable, although some problems still fail to converge. In addition, some cases still fail to converge after the rectification step budget is exhausted, and a few even move farther from the target expression, further showing that a larger rectification step budget is not always better; rather, it is a trade-off between cost and benefit. 


\begin{figure}[h]
	\centering
	\begin{subfigure}[t]{0.9\linewidth}
		\centering
		\includegraphics[width=1\linewidth]{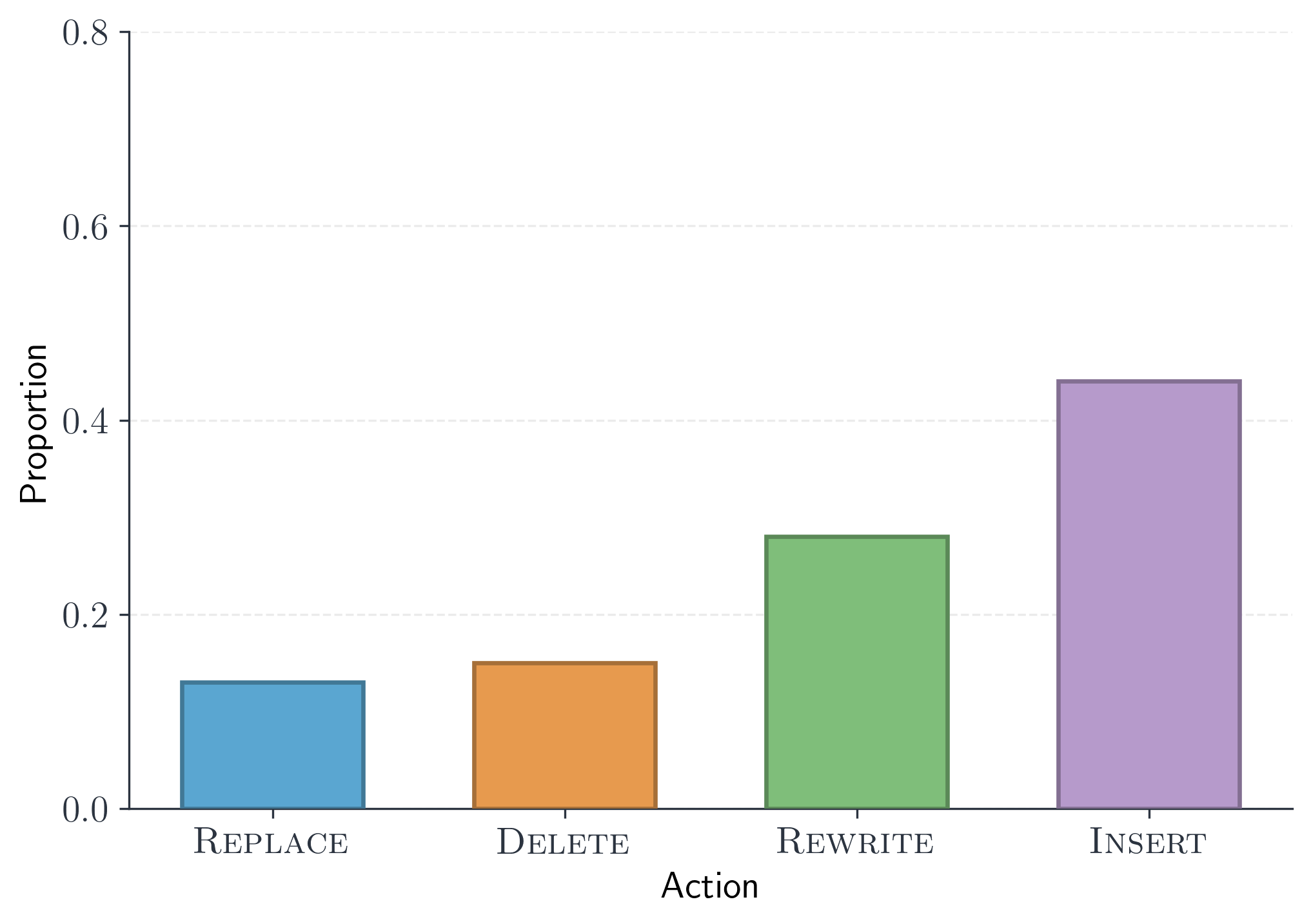}
		\caption{Frequency of different edit actions.}
		\label{fig:rectifier_action_usage}
	\end{subfigure}
	\hfill
	\begin{subfigure}[t]{0.9\linewidth}
		\centering
		\includegraphics[width=1\linewidth]{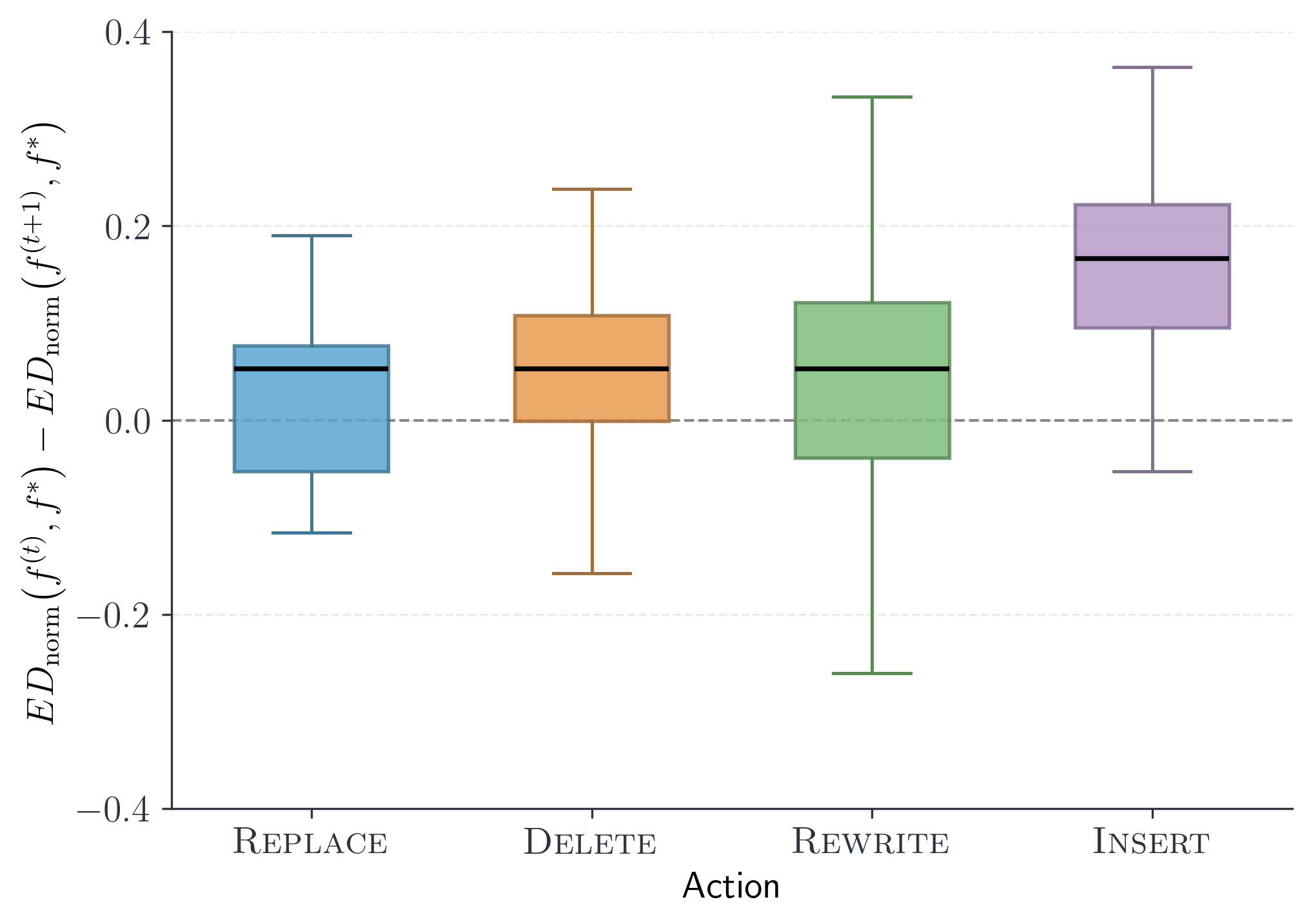}
		\caption{Distribution of the reduction in normalized edit distance resulting from different actions.}
		\label{fig:rectifier_action_effect}
	\end{subfigure}
	\caption{\textbf{Action frequency and effectiveness of the Rectifier.}
		The results treat all repeated runs as independent instances, rather than averaging them.}
	\label{fig:rectifier_action}
\end{figure}

To better understand how the Rectifier behaves in practice, we analyze both the relative frequency and the effect of different edit actions. The former indicates which edit actions are most often selected during rectification, while the latter quantifies how much each action moves $f^{(t)}$ toward $f^*$ via normalized edit distance. Fig.~\ref{fig:rectifier_action_usage} shows that \textsc{Insert} is the most frequent action, followed by \textsc{Rewrite}, whereas \textsc{Replace} and \textsc{Delete} are much less common. This phenomenon matches the error profile observed earlier: NeSymReS more often produces structurally incomplete expressions than redundant ones. As shown in Fig.~\ref{fig:rectifier_action_effect}, \textsc{Insert} has the largest median reduction in normalized edit distance, suggesting that adding missing operators, constants, variables, or local subtrees is often the most effective rectification. \textsc{Rewrite} also contributes, but with a broader distribution. By contrast, \textsc{Replace} is infrequent and has a smaller median effect, suggesting that single-token replacements are usually insufficient for the dominant error patterns. \textsc{Delete}, despite being rare, still yields a positive rectification effect. 

\subsubsection{Robustness to the first layer}

In this section, we examine whether the Rectifier remains effective when the first-layer error pattern shifts. To this end, we construct a family of NeSymReS variants by training them with different dropout rates $\{0.1, 0.2, 0.3\}$. For each converged NeSymReS variant, we fine-tune the Rectifier on rectification chains constructed from its predictions as initial states. In this controlled robustness experiment, dropout remains active during inference so that the forward passes induce additional perturbations. For each setting, we report the Complexity, Symbolic solution rate and Accuracy solution rate results. 

As shown in Table~\ref{tab:mc_dropout_rectifier}, the small performance fluctuations across dropout levels indicate that, after sufficient fine-tuning, the Rectifier is robust to these induced shifts in first-layer error patterns. Its final performance appears to depend more strongly on its own optimization quality than on differences in the upstream neural model. This observation is encouraging for practical adaptation, because it suggests that the Rectifier may be adapted to closely related neural models through interface alignment and Rectifier fine-tuning, without necessarily retraining the entire symbolic regression system. We further discuss the corresponding assumptions in~\ref{app:compatibility}.
%

\begin{table}[h]
	\centering
	\caption{\textbf{Effect of the Rectifier under different first-layer error patterns.} Results are reported as mean $\pm$ standard deviation over 5 runs.}
	\label{tab:mc_dropout_rectifier}
	\setlength{\tabcolsep}{5pt}
	\renewcommand{\arraystretch}{1.3}
	\begin{tabularx}{\linewidth}{l|>{\centering\arraybackslash}X>{\centering\arraybackslash}X>{\centering\arraybackslash}X}
		\toprule
		Dropout & {\small Accuracy solution rate (\%)} & {\small Symbolic solution rate (\%)} & Complexity \\
		\midrule
		0.1 & 79.38 $\pm$ {\scriptsize 1.84} & 60.11 $\pm$ {\scriptsize 2.03} & 11.68 $\pm$ {\scriptsize 0.21} \\
		0.2 & 76.11 $\pm$ {\scriptsize 2.10} & 56.78 $\pm$ {\scriptsize 1.27} & 11.96 $\pm$ {\scriptsize 0.41} \\
		0.3 & 78.06 $\pm$ {\scriptsize 1.73} & 59.17 $\pm$ {\scriptsize 2.50} & 11.77 $\pm$ {\scriptsize 0.52} \\
		\bottomrule
	\end{tabularx}
\end{table}

\subsubsection{Convergence analysis of fine-tuning}

We analyze the convergence behavior of EditSR during fine-tuning. Using the predictions of NeSymReS as the initial states of the supervised rectification chains, we fine-tune the Rectifier for 5 epochs and evaluate it on the validation set every 0.5 epoch. The Tagger is evaluated using action-prediction accuracy, while the Editor is evaluated by content-prediction accuracy for each edit action. Concretely, for a mini-batch of size $BS$, let $n \in \{1,\dots,BS\}$ index the samples in the batch, and let $L_n^{(t)}$ denote the length of the current expression of the $n$-th sample at rectification step $t$. The Tagger action-prediction accuracy is computed by:
\[
\mathrm{Acc}_{\mathrm{Tagger}}^{(t)}
=
\frac{
	\sum_{n=1}^{BS}\sum_{i=1}^{L_n^{(t)}}
	\mathbf{1}\!\left[\hat z_{n,i}^{(t)} = z_{n,i}^{*(t)}\right]
}{
	\sum_{n=1}^{BS} L_n^{(t)}
},
\]
where $\hat z_{n,i}^{(t)}$ and $z_{n,i}^{*(t)}$ are the predicted and target edit actions at position $i$, respectively. For the Editor, the content-prediction accuracy is computed separately for each edit action. Let
\[
\mathcal{B}_z^{(t)}=\{\,n \in \{1,\dots,BS\}: z_n^{*(t)} = z\,\}
\]
denote the subset of samples in the batch whose target edit action is $z$. Then the corresponding Editor accuracy is defined as
\[
\mathrm{Acc}_{\mathrm{Editor}}^{(t)}(z)
=
\frac{
	\sum_{n \in \mathcal{B}_z^{(t)}}
	\mathbf{1}\!\left[\hat{\mathbf{u}}_n^{(t)} = \mathbf{u}_n^{*(t)}\right]
}{
	|\mathcal{B}_z^{(t)}|
},
\]
where $\hat{\mathbf{u}}_n^{(t)}$ and $\mathbf{u}_n^{*(t)}$ are the predicted and target edit contents, respectively. For \textsc{Delete} and \textsc{Replace}, the edit content degenerates to a single token, whereas for \textsc{Insert} and \textsc{Rewrite}, it corresponds to a generated subtree.

\begin{figure}[h]
	\centering
	
	\begin{subfigure}[t]{0.48\linewidth}
		\centering
		\includegraphics[width=\linewidth]{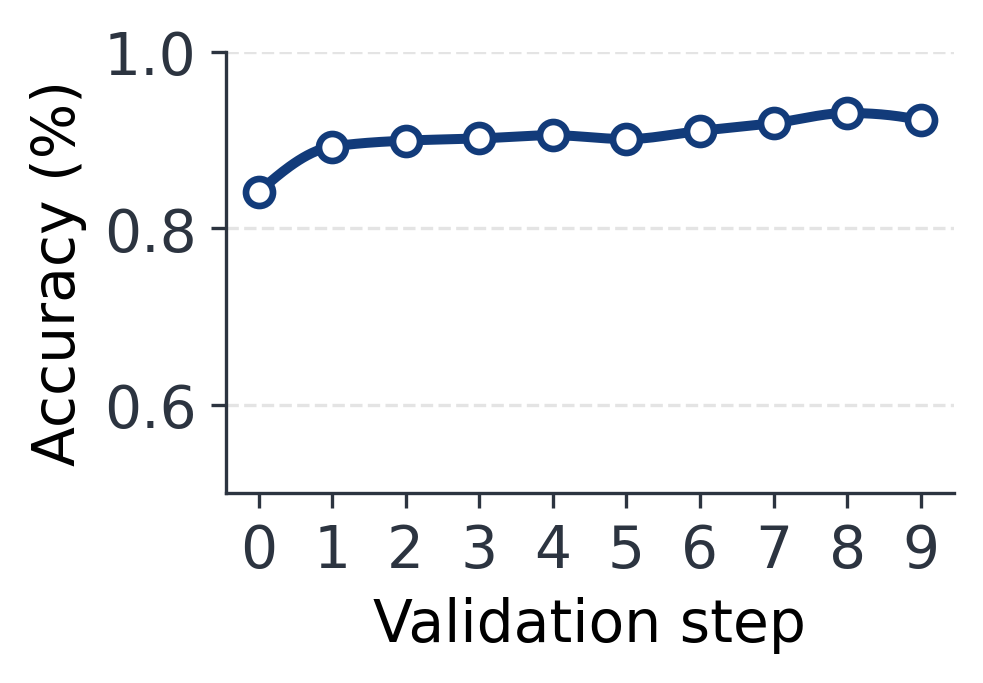}
		\caption{Tagger}
		\label{fig:tagger_val_acc}
	\end{subfigure}
	\hfill
	\begin{subfigure}[t]{0.48\linewidth}
		\centering
		\includegraphics[width=\linewidth]{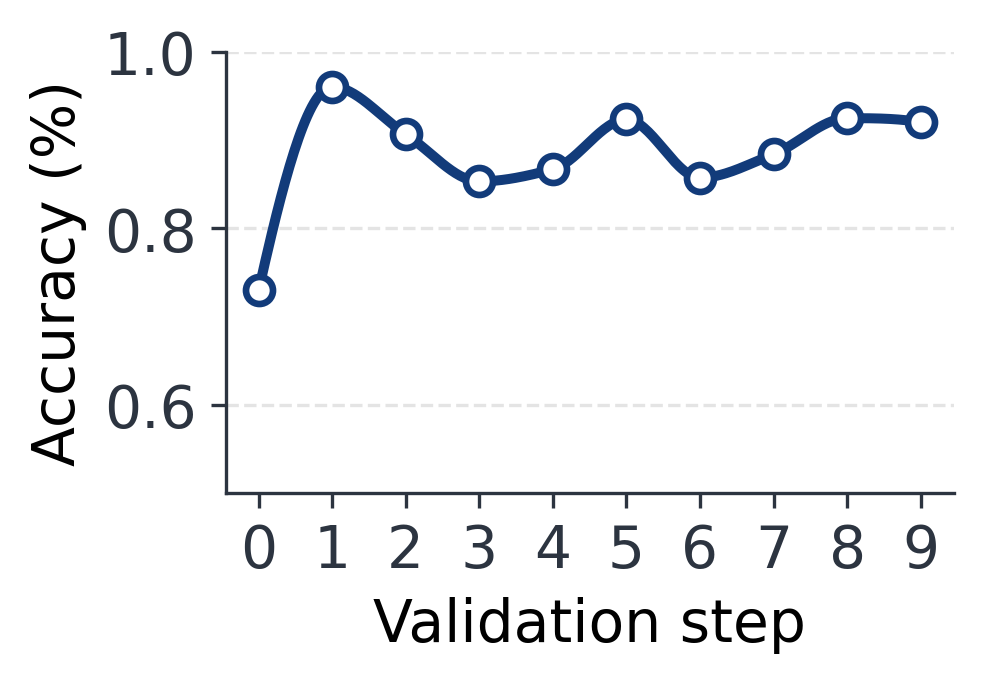}
		\caption{\textsc{Delete} of Editor}
		\label{fig:delete_val_acc}
	\end{subfigure}
	
	\vspace{0.3cm}
	
	\begin{subfigure}[t]{0.48\linewidth}
		\centering
		\includegraphics[width=\linewidth]{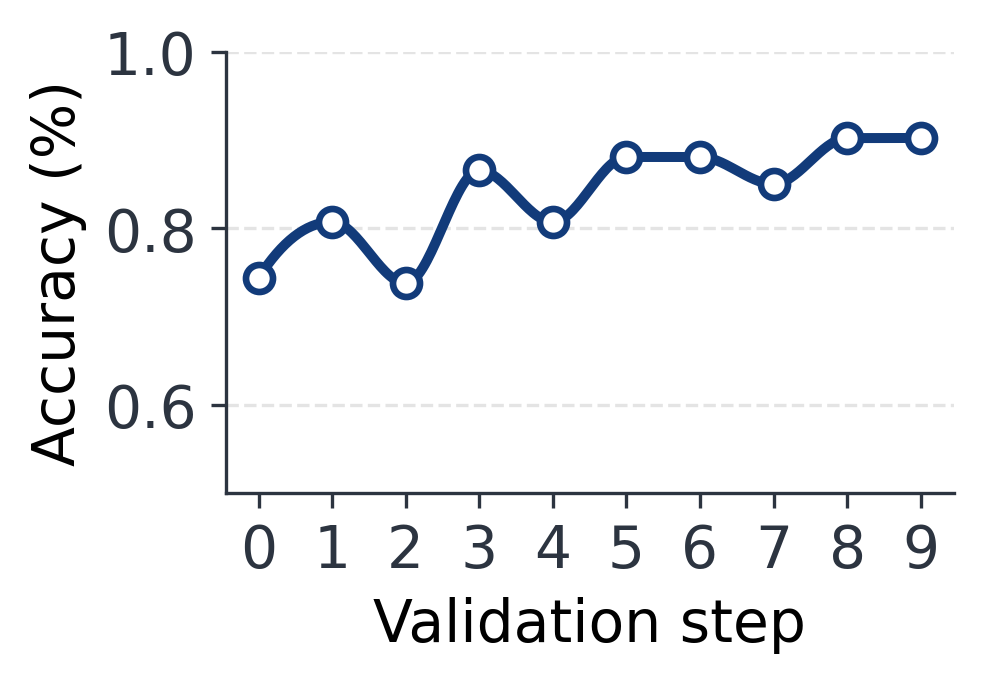}
		\caption{\textsc{Replace} of Editor}
		\label{fig:replace_val_acc}
	\end{subfigure}
	\hfill
	\begin{subfigure}[t]{0.48\linewidth}
		\centering
		\includegraphics[width=\linewidth]{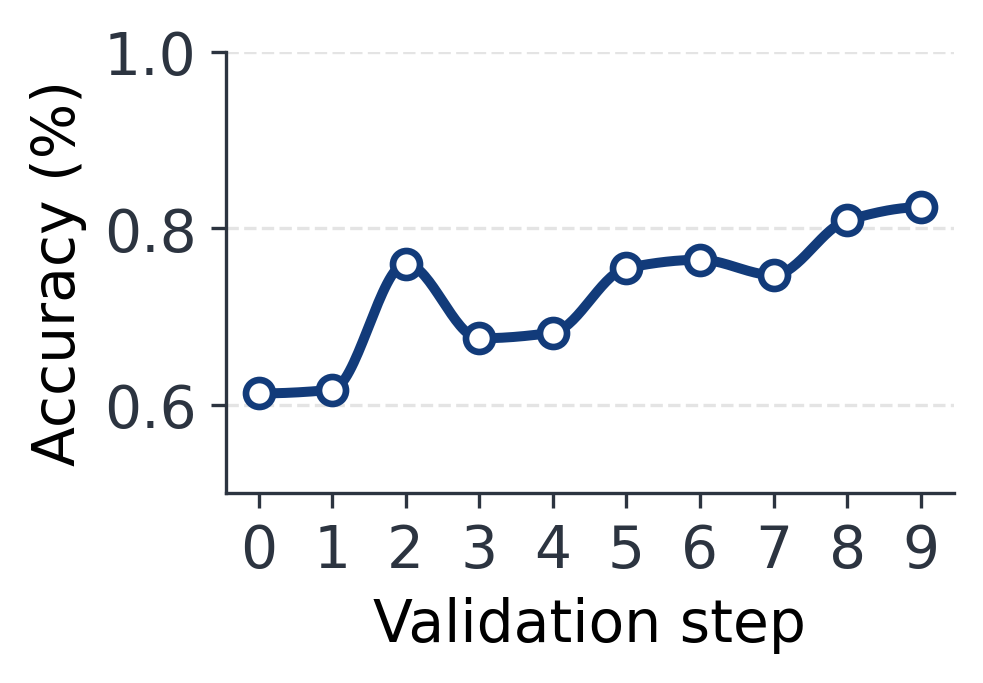}
		\caption{\textsc{Rewrite} of Editor}
		\label{fig:rewrite_val_acc}
	\end{subfigure}
	
	\vspace{0.3cm}
	
	\begin{subfigure}[t]{0.48\linewidth}
		\centering
		\includegraphics[width=\linewidth]{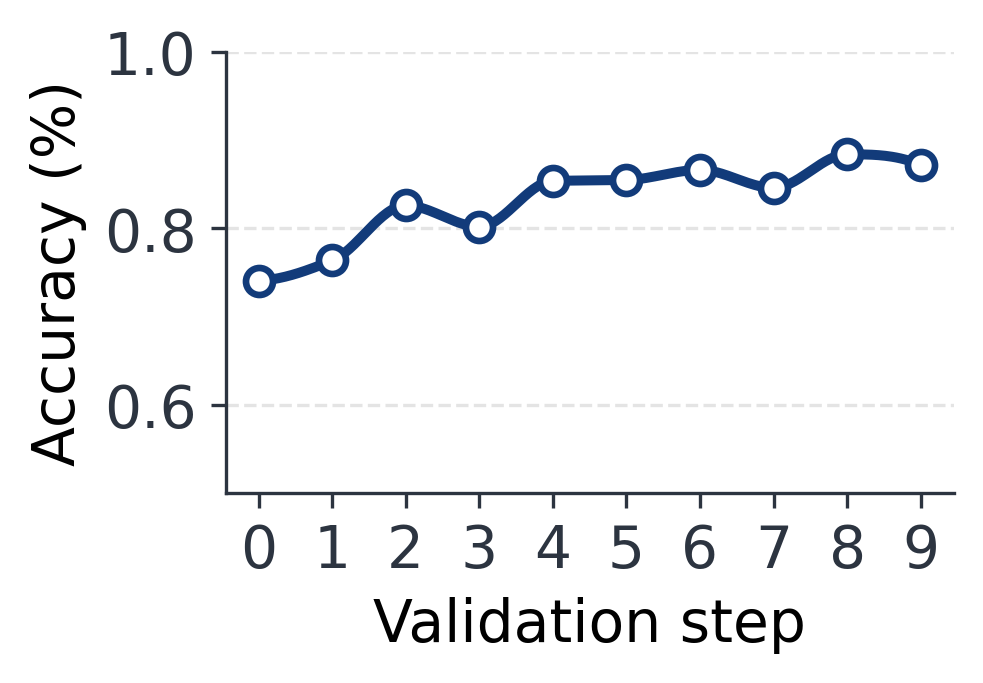}
		\caption{\textsc{Insert} of Editor}
		\label{fig:insert_val_acc}
	\end{subfigure}
	
	\caption{\textbf{Validation accuracy curves of the Rectifier during fine-tuning.} The Rectifier is fine-tuned for 5 epochs and evaluated on the validation set every 0.5 epoch.}
	\label{fig:rectifier_convergence}
\end{figure}

As shown in Fig.~\ref{fig:rectifier_convergence}, the Tagger converges smoothly and reaches more than 90\% validation accuracy in later stages. The Editor branches show a similar trend. The single-token actions, \textsc{Delete} and \textsc{Replace}, reach high accuracy quickly. The subtree-generation actions, \textsc{Insert} and \textsc{Rewrite}, are slightly harder but remain around 90\% accuracy. Overall, these curves indicate stable convergence within a few fine-tuning epochs. Moreover, we observe that all accuracy curves start above 60\%, which suggests that the large-scale pretraining of the Rectifier has already established a strong initialization and endowed it with the basic rectification capability before fine-tuning for a specific neural model.

\subsubsection{Successful and failed case analysis}
\label{case_analysis}

\begin{figure*}[h]
	\centering
	
	\makebox[\textwidth][c]{%
		\begin{subfigure}[t]{0.3\textwidth}
			\centering
			\includegraphics[width=\linewidth]{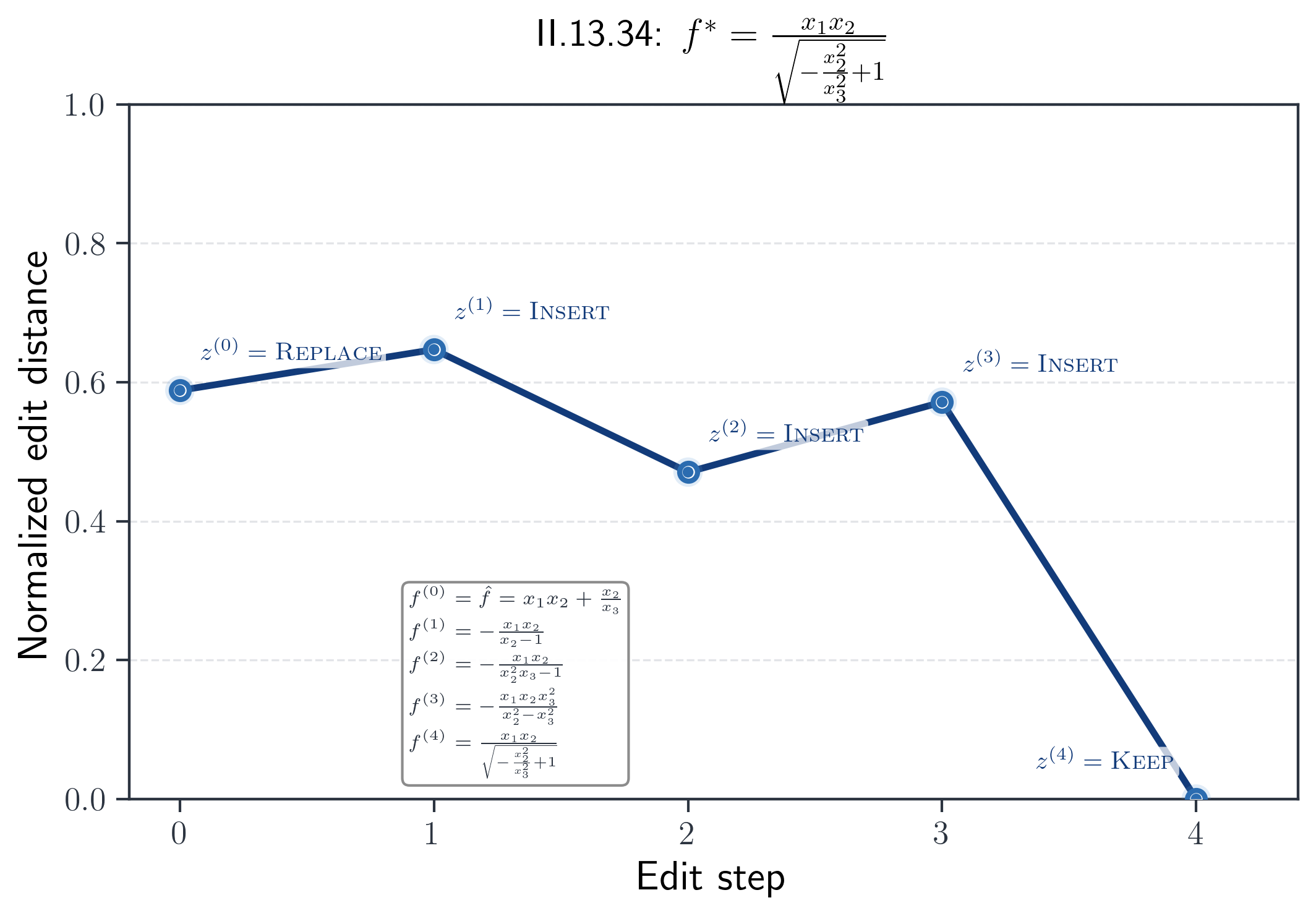}
			\label{fig:case_ii1334}
		\end{subfigure}
		\hspace{0.03\textwidth}
		\begin{subfigure}[t]{0.3\textwidth}
			\centering
			\includegraphics[width=\linewidth]{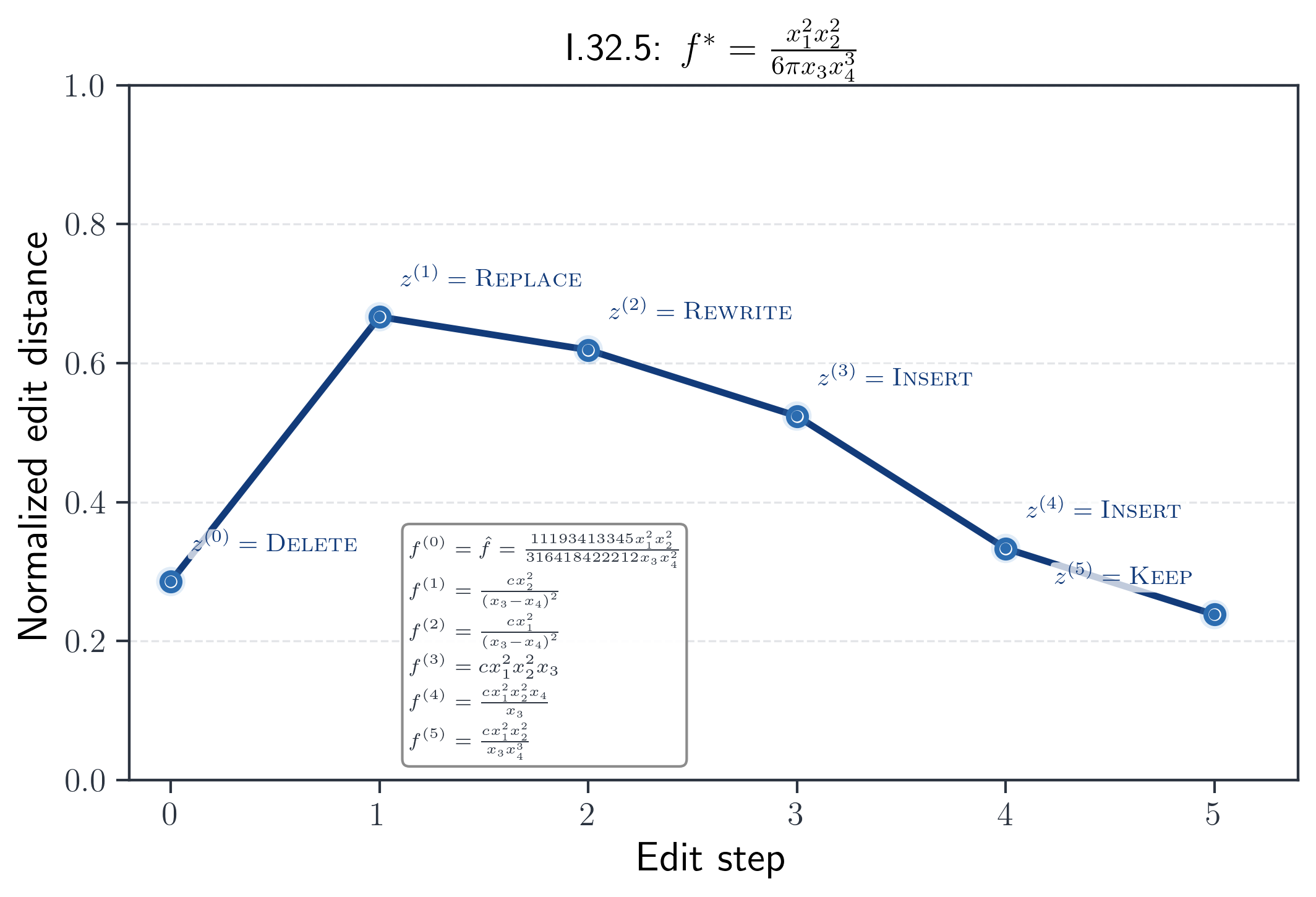}
			\label{fig:case_i325}
		\end{subfigure}
		\hspace{0.03\textwidth}
		\begin{subfigure}[t]{0.3\textwidth}
			\centering
			\includegraphics[width=\linewidth]{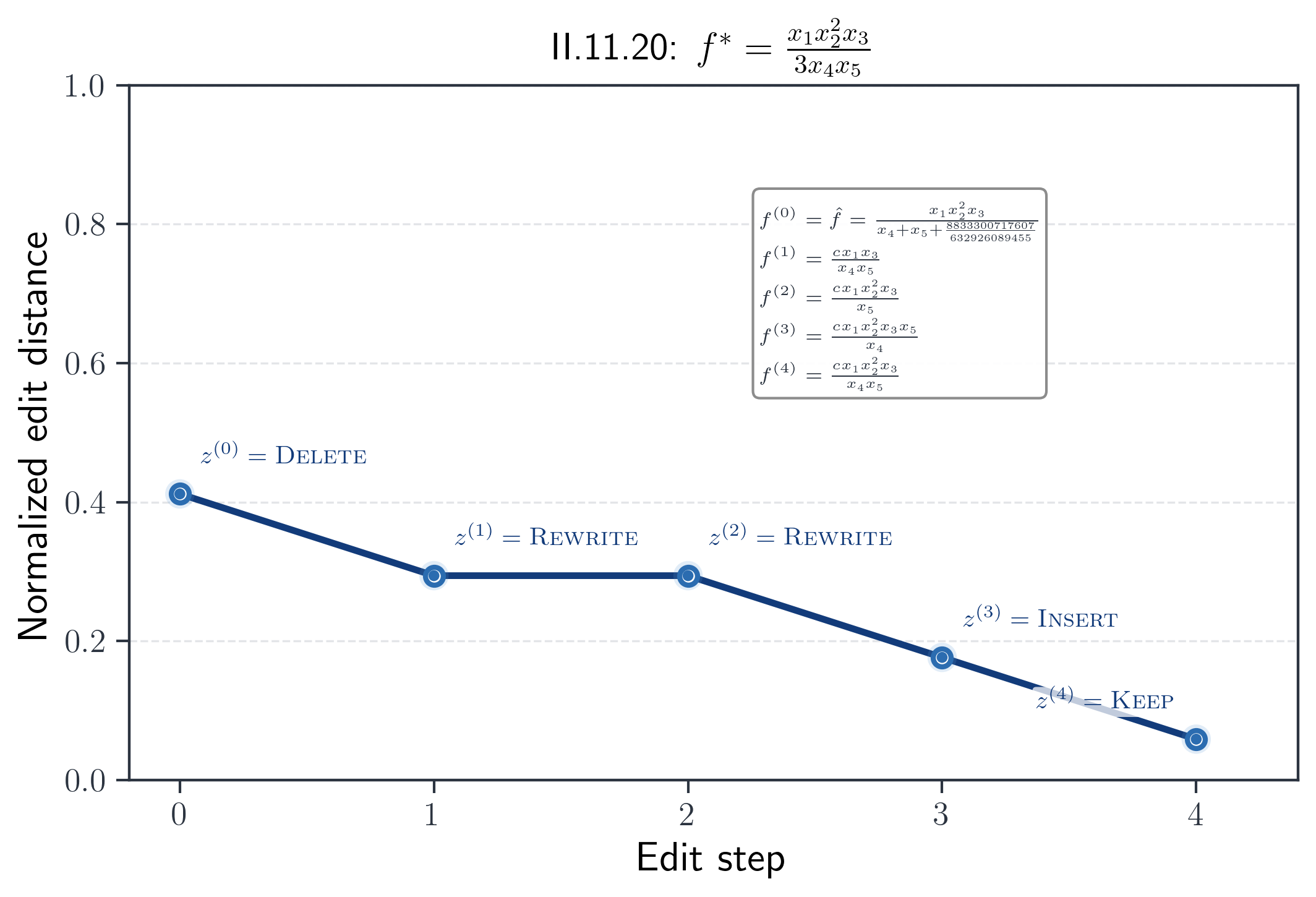}
			\label{fig:case_ii1120}
		\end{subfigure}
		
	}
	\makebox[\textwidth][c]{%
		
		\begin{subfigure}[t]{0.3\textwidth}
			\centering
			\includegraphics[width=\linewidth]{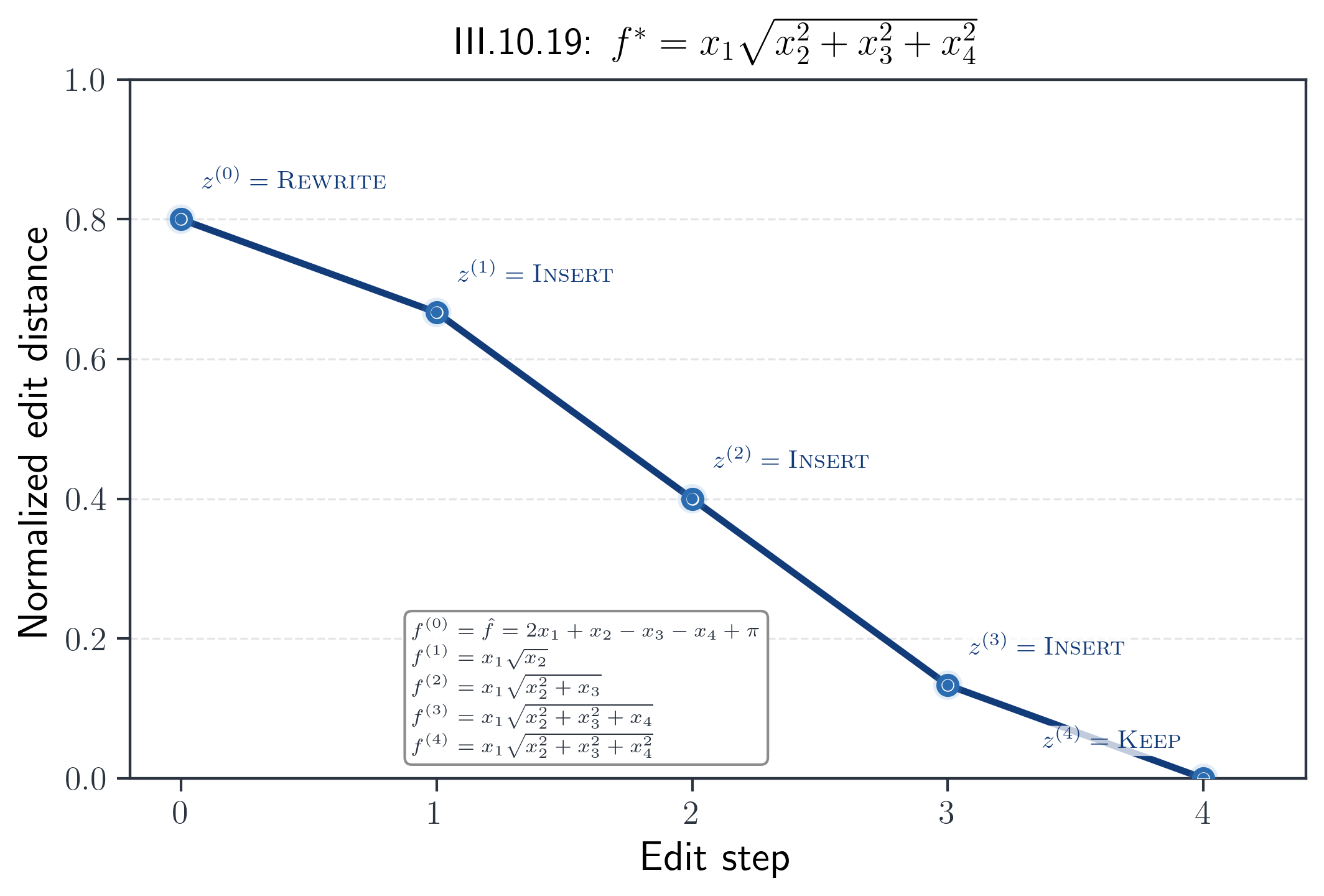}
			\label{fig:case_iii1019}
		\end{subfigure}
		\hspace{0.03\textwidth}
		\begin{subfigure}[t]{0.3\textwidth}
			\centering
			\includegraphics[width=\linewidth]{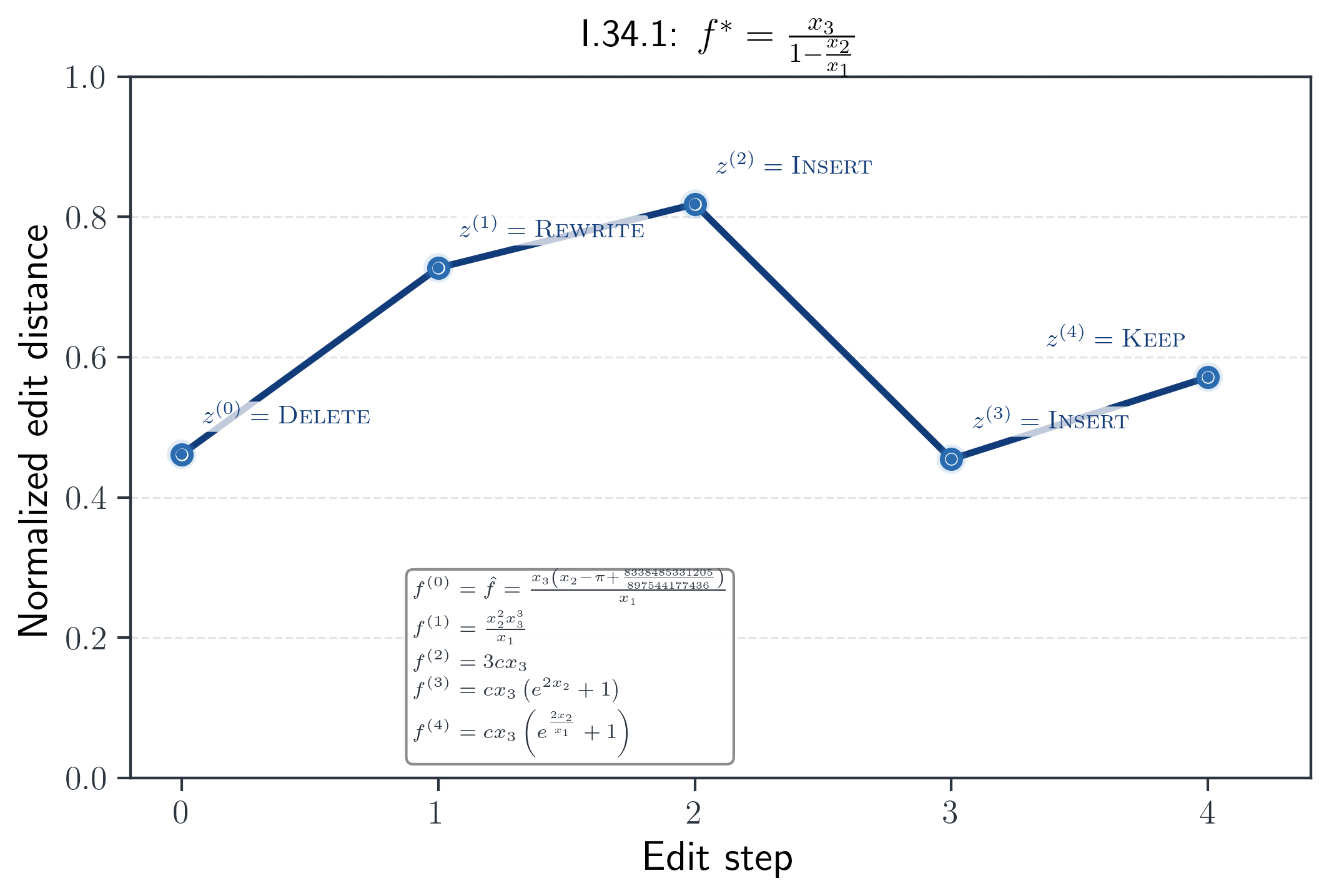}
			\label{fig:case_i341}
		\end{subfigure}
		\hspace{0.03\textwidth}
		\begin{subfigure}[t]{0.3\textwidth}
			\centering
			\includegraphics[width=\linewidth]{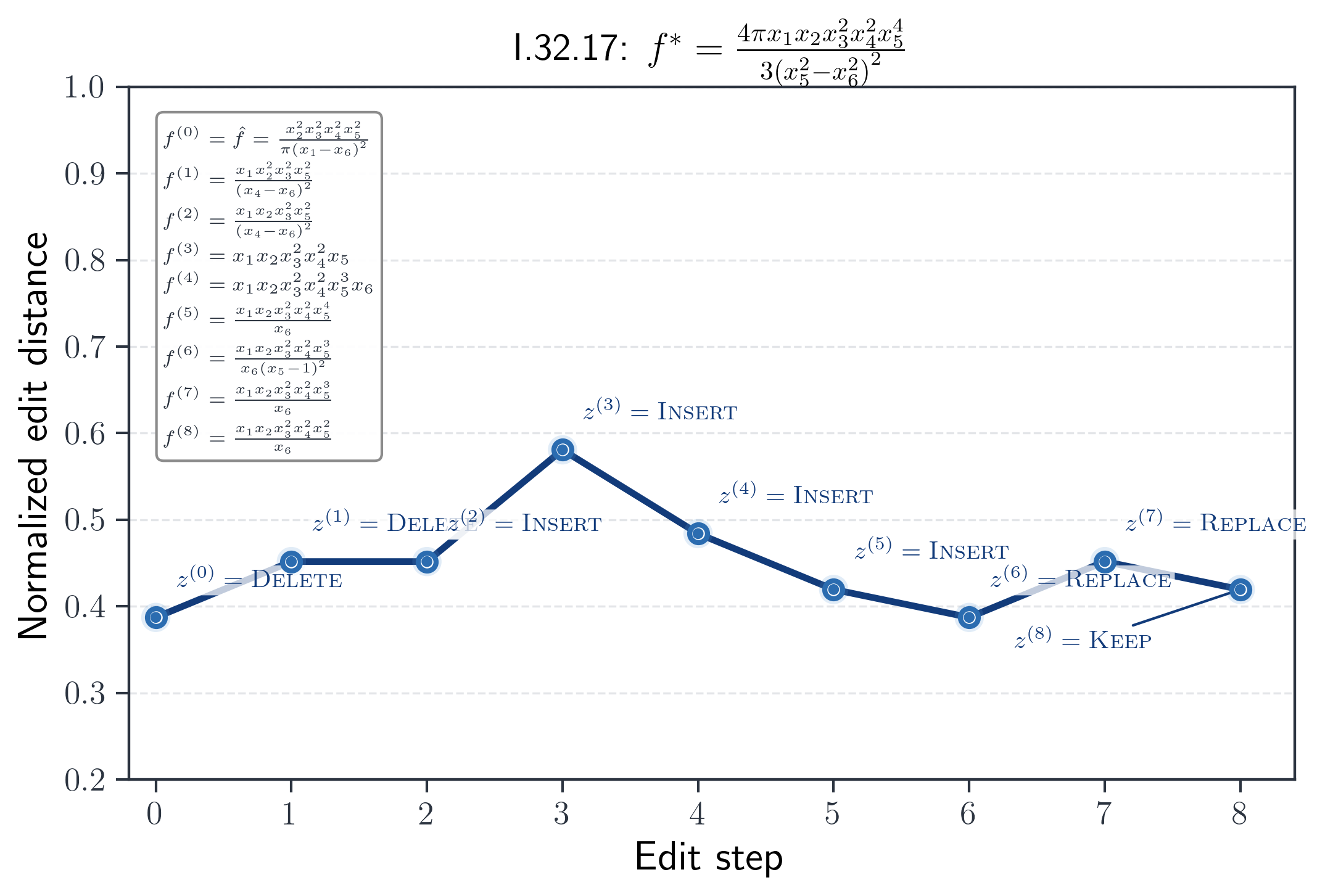}
			\label{fig:case_I3217}
		\end{subfigure}
	}
	
	\caption{\textbf{Representative successful and failed rectification cases of EditSR.}
		Each subplot shows the normalized edit distance between $f^{(t)}$ and $f^*$ across rectification steps, together with the selected edit action and the corresponding intermediate state. The first four cases show the dominant regime in which the first-layer prediction already provides a plausible structural scaffold, and the Rectifier recovers missing powers, factors, denominator terms, or function arguments through a short sequence of bounded edits. The last two cases illustrate the applicability boundary of bounded rectification. Although some edits still reduce part of the discrepancy, the large distance between the first-layer prediction and the target expression limits the Rectifier’s capability to fully recover the correct symbolic form within the available rectification step budget.}
	\label{fig:rectifier_case_studies}
\end{figure*}

Fig.~\ref{fig:rectifier_case_studies} contrasts four successful rectification trajectories with two failure cases to help clarify the regime where EditSR is most beneficial. In successful cases, the target expressions are structurally nontrivial, and the initial predictions already contain the main variable scaffold or part of the symbolic structure. Rectification is not strictly monotonic at the level of individual edits. Some intermediate actions temporarily increase the normalized edit distance. This behavior is natural for multi-step structural rectifications. Since each decision is conditioned on the current state rather than the whole edit history, the Rectifier is not forced to preserve a myopic locally best trajectory at every step. Instead, it can use an intermediate edit to expose a more suitable context for the subsequent step, and then recover the target structure through later rectifications. To more clearly analyze the scenarios where the Rectifier demonstrates advantages, we present additional successful rectification cases in~\ref{app:successful_rectification_cases}.

In failure cases, the first-layer prediction is already too far from the target expression,  so the remaining discrepancy is no longer concentrated in a local region. Under this condition, bounded rectification within a limited step budget may improve part of the structure but is less likely to recover the target expression fully. Even so, these cases remain consistent with our assumption: EditSR is most beneficial when the first-layer prediction is structurally plausible but locally incomplete, which is also the regime where restarting the global search would be unnecessarily expensive.

\subsection{EditSR vs TPSR}

In this section, we compare EditSR with TPSR on the Feynman benchmark, where TPSR is a representative post-hoc rectification strategy that enhances neural models using MCTS. Our purpose is to evaluate, based on the same NeSymReS backbone, the rectification effectiveness and runtime cost of the two methods.

\begin{figure}[h]
	\centering
	\includegraphics[width=0.9\linewidth]{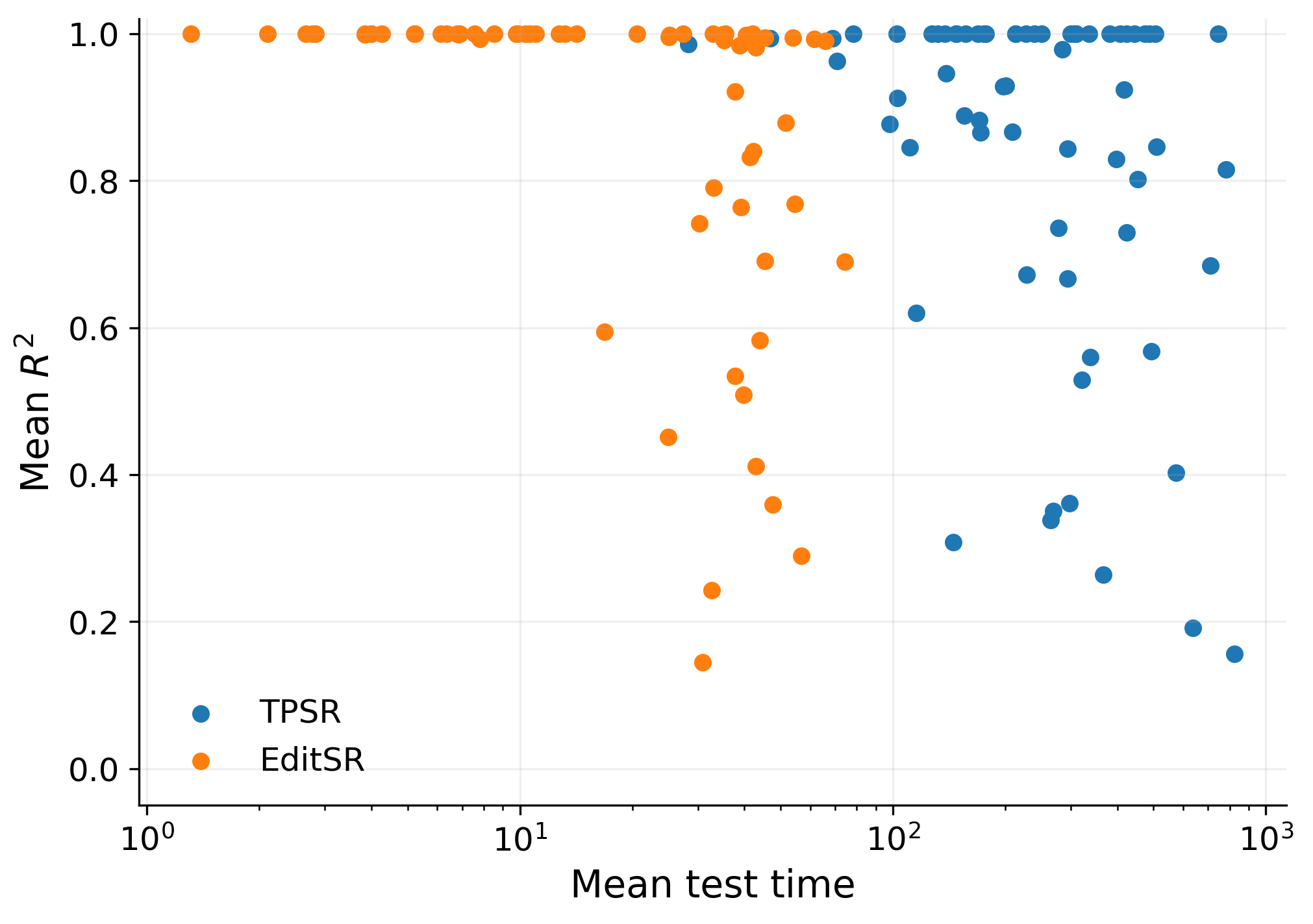}
	\caption{\textbf{$R^2$ versus test time for EditSR and TPSR.} The results represent the mean over 5 runs. Overall, EditSR occupies one of the strongest trade-off regions, achieving comparable or better performance on most problems while requiring substantially less time.}
	\label{fig:ablation_tpsr_vs_editsr_time}
\end{figure}

As shown in Fig.~\ref{fig:ablation_tpsr_vs_editsr_time}, TPSR can rectify part of the errors left by NeSymReS by reopening the search process, but it usually requires much higher computational cost. EditSR achieves higher $R^2$ results across more problems while requiring substantially less time than TPSR. This result supports our motivation: once NeSymReS has produced a plausible prediction, a short learned sequence of syntax-constrained local edits can be more efficient than reopening an online search. 

\section{Conclusion}
\label{sec:conclusion}
In this paper, we introduce EditSR to alleviate error accumulation in neural symbolic regression models. Instead of treating an incorrect prediction as a failed endpoint that requires restarting the search, EditSR regards post-hoc rectification as an edit-based state-transition chain, in which an incorrect prediction can be step-by-step rectified toward the target expression. By shifting the major burden of rectification from inference to pretraining, EditSR preserves the efficiency advantage of neural models and avoids repeatedly reopening global search after an incorrect prediction has been generated. In addition, because each rectification decision is conditioned on the current state rather than the history, errors introduced at earlier steps are not irrevocably inherited and can be rectified by subsequent edits, thereby helping suppress error accumulation. Extensive evaluations on mainstream benchmarks show that the advantage of EditSR is more significant in long expression generation, where one-pass autoregressive decoding is more susceptible to error accumulation. Moreover, EditSR maintains a competitive balance among $R^2$, Symbolic solution rate, and Complexity, as its rectification objective always moves toward the correct symbolic structure rather than merely satisfying a numerical error threshold.

\textbf{Limitations:} Shifting computationally intensive components to pretraining is one of the Rectifier's primary advantages. Therefore, the robustness of the Rectifier is grounded in training on artificially constructed rectification chains whose initial states are obtained by applying random corruptions to target expressions. Although random corruptions broaden the coverage of error patterns during training, the rectification effect under more severe out-of-distribution shifts remains to be validated. In addition, the Rectifier is currently limited by the finite exploration capabilities of the first layer, so some extremely complex problems may remain unresolved when relying on the Rectifier alone (as discussed in Section~\ref{case_analysis}). Enhancing the global exploration capability of the first layer, or coupling the Rectifier with stronger search-based models, is a promising direction for future work.

\textbf{Future work:} One worthwhile direction is to develop a tighter collaboration scheme between the Rectifier and models with stronger global search capabilities, such as genetic programming methods or deep learning-based symbolic regression models. Such exploratory models can generate diverse candidate expressions across a wider search space, and the Rectifier then refines promising candidates through local edits. This strategy may alleviate the dependence of the Rectifier on the exploration capabilities of the first layer and improve convergence on extremely complex problems that are difficult to resolve by rectification alone. More importantly, it may extend the role of post-hoc rectification from a follow-up module to a more general component within hybrid symbolic regression pipelines.

\section*{Acknowledgments}
This work was supported by Major Program of the National Key R\&D Program of China (No.2023YFA1009002, No.2023YFA1009000, No.2023YFA1009004), and the National Natural Science Foundation of China (No.12292980 and No.12292984).

\bibliographystyle{elsarticle-num}
\bibliography{references}

\clearpage
\onecolumn 
\appendix
\section{Details of Tagger and Editor}
\label{app:architecture_of_tagger_editor}

\begin{figure*}[h]
	\centering
	\includegraphics[width=0.9\linewidth]{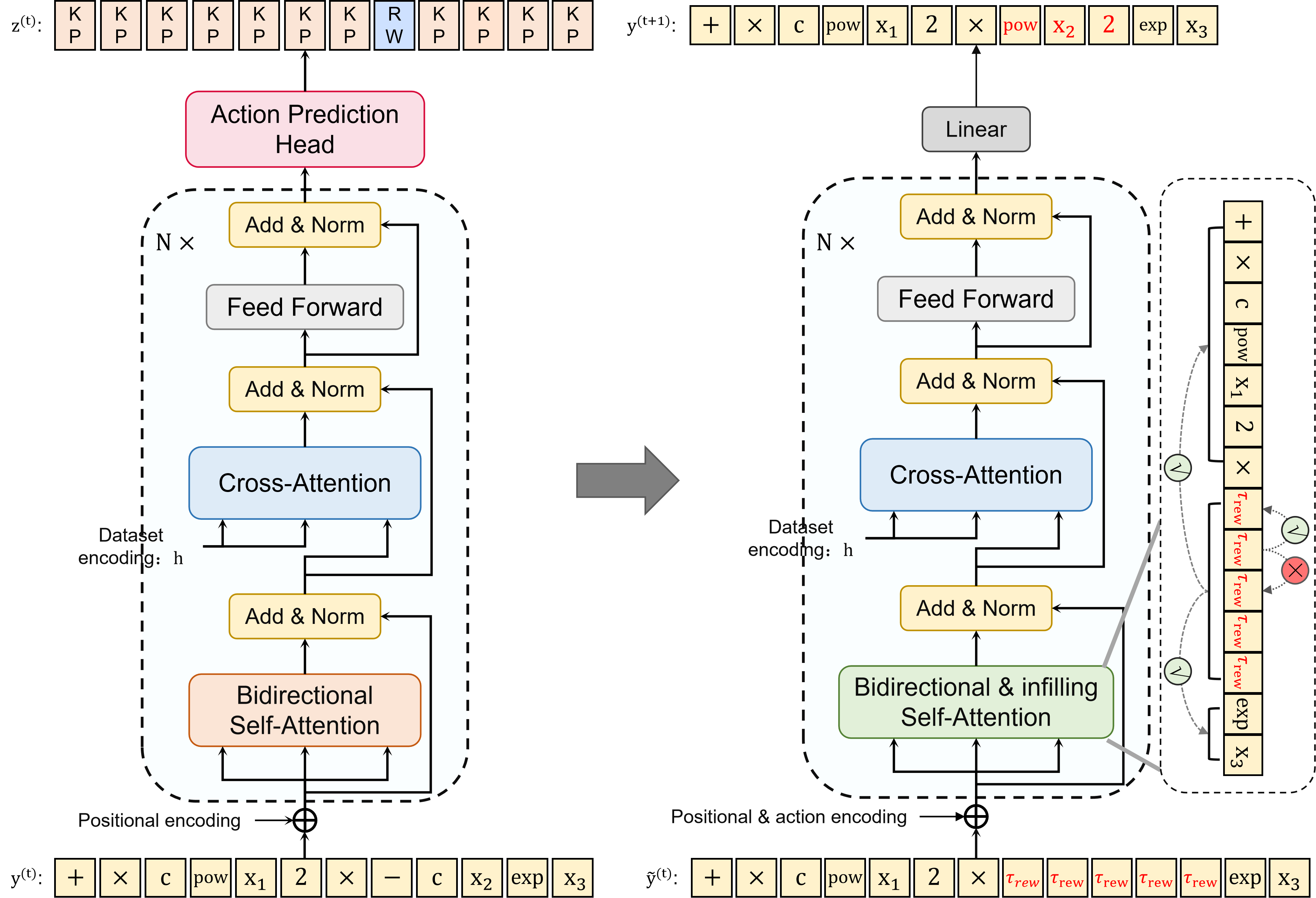}
	\caption{\textbf{Architecture illustration of Tagger and Editor.}  The example illustrates a \textsc{Rewrite} step. Left: the Tagger takes the current state $f^{(t)}$ as input and processes it with Transformer blocks using bidirectional self-attention and cross-attention to the dataset encoding $h$. An action-prediction head then outputs a position-wise distribution over admissible edit actions. The position whose best non-\textsc{Keep} action has the highest confidence is selected for editing. Right: the Editor receives the context state $\tilde f^{(t)}$, where the positions requiring edit are replaced by action-specific placeholders. It uses bidirectional and infilling self-attention, together with cross-attention to $h$, to generate the edit content.}
	\label{fig:architecture}
\end{figure*}

The Tagger and Editor share the same dataset encoding $h$, but they play different roles in the rectification loop. The Tagger is responsible for predicting the edit position and action. Given the current state $f^{(t)}$, it encodes the whole sequence with bidirectional self-attention, fuses $h$ through cross-attention, and predicts a position-wise distribution over the admissible actions in $\mathcal{Z}=\{\textsc{Keep},\textsc{Replace},\textsc{Delete},\textsc{Rewrite},\textsc{Insert}\}$. After masking out inadmissible actions according to syntactic constraints, the highest-confidence non-\textsc{Keep} position--action pair is selected and passed to the Editor. The Editor adopts bidirectional and infilling self-attention: visible context tokens attend to each other bidirectionally, while tokens inside the hole are generated autoregressively with access to both left and right context. For \textsc{Replace} and \textsc{Delete}, the prediction degenerates to a single token. Fig.~\ref{fig:architecture} illustrates both modules and their interaction in one-step rectification. 

\section{Rectifier adaptation assumptions}
\label{app:compatibility}

In the proposed framework, the Rectifier is decoupled from the first layer. When an incorrect expression has been predicted by the first layer, the rectification loop only requires the current parseable expression state $f^{(t)}$ and a dataset encoding $h$. It does not depend on the internal hidden states, decoding trajectory, or search history of the first-layer neural model. This decoupled design is an advantage of EditSR, because replacing the first layer does not require redesigning the Tagger and Editor. However, in practice, this adaptation route relies on several assumptions. First, the neural model should output expressions in a representation that can be aligned with the syntax constraints used by the Rectifier, such as the prefix-style tree representation. Second, the vocabularies and constant conventions should either match directly or be convertible to a common representation, so that the edit actions remain well defined. Provided that the above conditions are met, the Rectifier may be deployed.            

Once a new neural symbolic regression model has converged, the practical adaptation step is to construct rectification chains from its predictions and fine-tune only the Rectifier, while keeping the neural model itself unchanged. Therefore, the main additional cost lies in learning the model-specific error distribution, rather than retraining the full symbolic regression system. This property is particularly attractive in scenarios where the pretrained neural model has already absorbed substantial pretraining cost and only incremental performance gains are still desired. Our ablation studies show that, after fine-tuning, the Rectifier remains stable across NeSymReS variants trained under different dropout configurations, indicating that it can accommodate related shifts in first-layer error patterns. Therefore, the current evidence suggests the practical feasibility of adapting the Rectifier across widely compatible first-layer error patterns after interface alignment and fine-tuning. Broader cross-architecture validation remains a meaningful direction for future work.

\section{Why rectification decisions are conditioned on the current state}
\label{app:current_state}

In our framework, the supervised signal is defined on one-step state transitions. For each sampled intermediate state $f^{(t)}$, the state-transition algorithm constructs a target triplet $(p^{*(t)}, z^{*(t)}, u^{*(t)})$ that specifies how to move from $f^{(t)}$ to $f^{(t+1)}$ toward $f^*$. Therefore, the learning objective of the Rectifier is not to imitate a particular historical trajectory, but to learn how to act on the expression that is currently observed. Furthermore, since the initial states of the artificially constructed rectification chains are obtained by applying random corruptions to the target expression, the Rectifier is exposed to a wide variety of local and global error patterns. Conditioning on the current state allows these heterogeneous error patterns to be mapped into a unified training format, i.e., regardless of how a state was reached, the Rectifier only needs to decide what edit should be executed. In this sense, the current state-based modeling approach enlarges the effective coverage of the rectification chains, because different edit histories that lead to comparable states can share the same type of supervision.

From the inference perspective, conditioning on the current state makes the rectification loop more flexible. After each edit, the Tagger and the Editor re-evaluate the updated parseable expression, rather than rigidly adhering to a fixed path determined by earlier decisions. Therefore, previously generated content is not irrevocable. A subtree introduced earlier can still be deleted, replaced, or rewritten later if it proves inconsistent with the target structure. Moreover, because the state-transition chain always remains in the space of syntactically valid expressions, later edits are performed on a stable structural object rather than on a partially broken one. Therefore, EditSR is less vulnerable to error accumulation than standard autoregressive generation.

\section{Illustration of supervised rectification-chain construction}
\label{app:worked_chain_example}

To further clarify how the state-transition algorithm constructs the supervised rectification chain, we provide a concrete example under the cost function in Eq.~\eqref{eq:frontier_cost_34}. Throughout this example, all edit positions $p$ refer to the prefix positions in the current state $f^{(t)}$. We use $\lambda=0.1$ and the generation budget $5$ of the Editor.

Consider the current state
\[
f^{(0)} = [\texttt{sub}, \texttt{mul}, x_1, x_2, \texttt{mul}, \texttt{mul}, x_3, x_5, x_6],
\]
and the target expression
\[
f^{*} = [\texttt{add}, \texttt{mul}, x_1, x_2, \texttt{mul}, \texttt{mul}, x_3, x_4, \texttt{mul}, x_5, x_6].
\]
In tree form, these correspond to
\[
f^{(0)}
=
\texttt{sub}\!\Big(
\texttt{mul}(x_1,x_2),
\texttt{mul}\big(\texttt{mul}(x_3,x_5),x_6\big)
\Big),
\]
and
\[
f^{*}
=
\texttt{add}\!\Big(
\texttt{mul}(x_1,x_2),
\texttt{mul}\big(\texttt{mul}(x_3,x_4),\texttt{mul}(x_5,x_6)\big)
\Big).
\]

At the root, the current subtree and the target subtree have the same arity. Since the current root is an internal node, \textsc{Insert} is not executable. Since the aligned target at the root is also an internal subtree rather than a leaf, \textsc{Delete} is not admissible. Since the current subtree does not yet match the target subtree, \textsc{Keep} is excluded. Therefore, the admissible candidates at $p=1$ reduce to \textsc{Replace} and \textsc{Rewrite}. Since the current node has child nodes, the deferred-edit plan is available. Therefore, the state-transition algorithm compares a direct-edit plan with a deferred-edit plan.

One possible plan is to apply a root-level \textsc{Rewrite}. However, the full target subtree rooted at $p'=1$ has prefix length $11$, which exceeds the generation budget $S$. The algorithm first constructs a budget-limited intermediate subtree
\[
\tilde I^{*}(1)=[\texttt{add}, \texttt{mul}, x_1, x_2, x_3],
\]
which corresponds to $\texttt{add}(\texttt{mul}(x_1,x_2),x_3)$. According to Eq.~\eqref{eq:frontier_cost_34}, the corresponding root-level \textsc{Rewrite} has cost
\[
1+\lambda |\tilde I^{*}(1)| = 1+0.1\times 5 = 1.5.
\]
After this \textsc{Rewrite}, the state becomes
\[
[\texttt{add}, \texttt{mul}, x_1, x_2, x_3].
\]
The remaining mismatch is then concentrated at the second child. At $p=5$, the current node is the leaf $x_3$, while the aligned target is the subtree
\[
\texttt{mul}\big(\texttt{mul}(x_3,x_4),\texttt{mul}(x_5,x_6)\big),
\]
whose prefix length is $7>S$. Since the current node is a leaf, the deferred-edit plan is not available because there are no child nodes to recurse into. We therefore only consider direct-edit actions at $p=5$. Since the current node is a leaf, \textsc{Delete} and \textsc{Rewrite} are not executable. Since the aligned target is a subtree rather than a leaf, \textsc{Replace} is not admissible. Since the current token does not match the target subtree, \textsc{Keep} is excluded. Therefore, the only admissible action at $p=5$ is \textsc{Insert}. The algorithm then inserts the budget-limited intermediate subtree
\[
[\texttt{mul}, \texttt{mul}, x_3, x_4, x_5],
\]
with cost
\[
1+\lambda\times 5 = 1.5.
\]
A further insertion is still required at the leaf aligned with $[\texttt{mul},x_5,x_6]$, with cost
\[
1+\lambda\times 3 = 1.3.
\]
Therefore, the total cost of this direct-edit plan is
\[
1.5+1.5+1.3=4.3.
\]

Alternatively, the algorithm may defer the repair to descendant nodes. Since the root symbols differ while the arities agree, a root-level \textsc{Replace} at $p=1$, namely $\texttt{sub}\rightarrow\texttt{add}$, is admissible and has cost $1$. The left subtree $\texttt{mul}(x_1,x_2)$ already matches the target subtree and therefore contributes zero cost. The remaining mismatch is entirely located in the right subtree,
\[
\texttt{mul}\big(\texttt{mul}(x_3,x_5),x_6\big)
\quad\text{vs.}\quad
\texttt{mul}\big(\texttt{mul}(x_3,x_4),\texttt{mul}(x_5,x_6)\big).
\]
At the root of this right subtree, the current node and the aligned target have the same token and the same arity. Since this node has child nodes, the deferred-edit plan remains available. Therefore, the algorithm continues to recurse into its descendants rather than editing this matched internal node directly.

The first mismatch is then reached at $p=8$, where the current token is the leaf $x_5$ and the aligned target is the leaf $x_4$. Since the current node is a leaf, the deferred-edit plan is not available because there are no child nodes to recurse into. We therefore only consider direct-edit actions at $p=8$. Since the current node is a leaf, \textsc{Delete} and \textsc{Rewrite} are not executable. Since the aligned target is also a leaf rather than a subtree, \textsc{Insert} is not admissible. Since the current token does not match the target token, \textsc{Keep} is excluded. Therefore, the only admissible action at $p=8$ is \textsc{Replace}, whose cost is $1$.

The second mismatch is reached at $p=9$, where the current token is the leaf $x_6$ and the aligned target is the subtree $[\texttt{mul},x_5,x_6]$. Since this node is also a leaf, the deferred-edit plan is no longer available. According to the same feasibility analysis, only \textsc{Insert} is admissible at this position, with cost
\[
1+\lambda\times 3 = 1.3.
\]
Therefore, the total cost of this deferred-edit plan is
\[
1+1+1.3=3.3.
\]
Since this cost is lower than $4.3$, the state-transition algorithm selects the first edit of this deferred-edit plan. The first one-step supervision label is
\[
\bigl(p^{*(0)}, z^{*(0)}, u^{*(0)}\bigr)
=
\bigl(1, \textsc{Replace}, [\texttt{add}]\bigr).
\]

After applying this edit, the next state becomes
\[
f^{(1)} =
[\texttt{add}, \texttt{mul}, x_1, x_2, \texttt{mul}, \texttt{mul}, x_3, x_5, x_6].
\]
When the state-transition algorithm is invoked again on $f^{(1)}$, the first unresolved mismatch is reached at $p=8$. Since the current node is a leaf and has no child nodes, the deferred-edit plan is unavailable. According to the same feasibility analysis as above, only \textsc{Replace} is admissible at this position. Therefore, the state-transition algorithm emits
\[
\bigl(p^{*(1)}, z^{*(1)}, u^{*(1)}\bigr)
=
\bigl(8, \textsc{Replace}, [x_4]\bigr).
\]
Applying this edit gives
\[
f^{(2)} =
[\texttt{add}, \texttt{mul}, x_1, x_2, \texttt{mul}, \texttt{mul}, x_3, x_4, x_6].
\]

When the state-transition algorithm is invoked on $f^{(2)}$, the only remaining mismatch is reached at $p=9$. Since the current node is a leaf and has no child nodes, the deferred-edit plan is unavailable. According to the same feasibility analysis as above, only \textsc{Insert} is admissible at this position. Therefore, the state-transition algorithm emits
\[
\bigl(p^{*(2)}, z^{*(2)}, u^{*(2)}\bigr)
=
\bigl(9, \textsc{Insert}, [\texttt{mul}, x_5, x_6]\bigr).
\]
Applying this edit gives
\[
f^{(3)}
=
[\texttt{add}, \texttt{mul}, x_1, x_2, \texttt{mul}, \texttt{mul}, x_3, x_4, \texttt{mul}, x_5, x_6]
=
f^{*}.
\]

The complete supervised rectification chain is therefore
\[
f^{(0)}
\;\xrightarrow[p^{*(0)}=1]{\textsc{Replace}}\;
f^{(1)}
\;\xrightarrow[p^{*(1)}=8]{\textsc{Replace}}\;
f^{(2)}
\;\xrightarrow[p^{*(2)}=9]{\textsc{Insert}}\;
f^{(3)}=f^{*}.
\]

This example shows that the one-step supervision label at each non-terminal state is determined by minimizing the total cost over admissible executable plans. Although the state-transition algorithm is recursive, its practical overhead remains limited because the action admissibility constraints exclude most infeasible actions at each state, while the deferred-edit plan becomes unavailable immediately once the current node is a leaf.

\section{Statistics of the benchmarks}
Table~\ref{tab:benchmark_statistics} summarizes the benchmark statistics, including the number of problems, the number of samples, and the split schemes. For all officially released datasets, we adopt the training-test split schemes specified in their original publications.

\newcolumntype{C}[1]{>{\centering\arraybackslash}m{#1}}
\newcolumntype{L}[1]{>{\raggedright\arraybackslash}m{#1}}
\begin{table*}[t]
	\centering
	\caption{Benchmark statistics, where the number of samples and the complexity are averages over problems.}
	\label{tab:benchmark_statistics}
	\renewcommand{\arraystretch}{1.3}
	
	\begin{tabular*}{\textwidth}{@{\extracolsep{\fill}}@{}
			L{5cm}
			C{1.5cm}
			C{1.8cm}
			C{1.5cm}
			C{3cm}
			C{1.5cm}
			C{1.1cm}
			@{}}
		\toprule
		Benchmark
		&No. Problems
		&Max. Dimension
		&No. Samples
		&Training split
		&Complexity
		&Natural noise
		\\
		\midrule
		
		Standard benchmarks~\citep{uy2011semantically,keijzer2003improving,korns2011accuracy,petersen2019deep,jin2019bayesian}
		& 91 & 3 & 200 & Independent sampling for training and testing  & 11.49 & No
		\\
		
		Feynman~\citep{la2021contemporary}
		& 120 & 9 & 10{,}000 & 75\%  & 16.57 & No
		\\
		
		ODE-Strogatz~\citep{la2021contemporary}
		& 14 & 2 & 400 & 75\%  & 13.36 & No
		\\

		Phenomenological \& first-principles~\citep{aldeia2025call}
		& 12 & 4 & 48 & 75\%  & 9.17 & Yes
		\\
		
		Black-box~\citep{aldeia2025call}
		& 12 & 51 & 4387 & 75\%  & $\sim$ & No
		\\
		
		\bottomrule
	\end{tabular*}
\end{table*}

\section{Main parameters}
Table~\ref{tab:equation_generator} describes the parameters used by the skeleton generator. Note that even if constants or exponents are sampled from a single distribution, they induce a broader distribution after composition, such as $x^3 \times x^2 = x^6$. In Table~\ref{tab:rectifier_hparams}, we list the key architectural and inference parameters used in the current implementation.

\renewcommand{\tabularxcolumn}[1]{m{#1}} 
\begin{table}[H]
	\centering
	\renewcommand{\arraystretch}{1.3}
	\caption{Hyperparameters of the skeleton generator and operator sampling weights.}
	\label{tab:equation_generator}
	\begin{tabularx}{\linewidth}{>{\raggedright\arraybackslash}X | >{\centering\arraybackslash}X}
		\hline
		\textbf{Description} & \textbf{Value} \\ \hline
		Variable set & $\{x_1,\cdots,x_{10}\}$ \\
		Max number of unary operators & $5$ \\
		Max number of binary operators & $d + 5$ \\
		Max number of constants & $3$ \\
		Max length & $30$ \\
		Constant distribution & $\mathcal{U}(-10,10)$ \\
		Sampling weight of binary operators & \texttt{add}: 1,\ \texttt{sub}: 0.5,\ \texttt{mul}: 1,\ \texttt{div}: 0.5 \\
		Sampling weight of unary operators &
		\texttt{abs}: 0.1,\ \texttt{pow2}: 1,\ \texttt{pow3}: 1,\ \texttt{pow5}: 0.1,\ \texttt{sqrt}: 1, \\
		& \texttt{sin}: 0.5,\ \texttt{cos}: 0.5,\ \texttt{tan}: 0.1,\ \texttt{arcsin}: 0.1,\ \texttt{log}: 0.5,\ \texttt{exp}: 0.5 \\ \hline
	\end{tabularx}
\end{table}

\renewcommand{\tabularxcolumn}[1]{>{\raggedright\arraybackslash}p{#1}}
\begin{table}[h]
	\centering
	\renewcommand{\arraystretch}{1.3}
	\caption{Key rectifier hyperparameters in the current implementation.}
	\label{tab:rectifier_hparams}
	\begin{tabularx}{\linewidth}{>{\raggedright\arraybackslash}p{0.22\linewidth} X >{\raggedleft\arraybackslash}p{0.20\linewidth}}
		\hline
		\textbf{Hyperparameter} & \textbf{Description} & \textbf{Value} \\
		\hline
		\multicolumn{3}{l}{\textit{\textbf{Train}}} \\ \hline
		Training and validation set sizes & $-$ &100,000,000 and 1,000  \\
		Epochs & $-$ & 30 for training NeSymReS, 30 for training the Rectifier, and 5 for fine-tuning\\
		Model dimensions & $-$ & 512 \\
		Heads & $-$ & 8 \\
		Learning rate & $-$ & 0.0001 with cosine decay \\
		Batch size & $-$ & 200 \\
		Set encoder layers & $-$ & 4 \\
		Decoder layers & $-$ & 8 \\
		Tagger / Editor layers & $-$ & 4 \\
		$T'$ & Number of corruption steps  & $\mathcal{U}\{1,2,...,20\}$\\
		$L_{\max}$ & Maximum prefix expression length & 50 \\
		$S$ & Placeholder budget of the Editor & 5 \\
		$\lambda$ & Subtree length penalty in Eq.~\ref{eq:frontier_cost_34} &0.1 \\
		\hline
		\multicolumn{3}{l}{\textit{\textbf{Inference}}} \\ \hline
		B & Beam size of NeSymReS & 30 \\
		$T_{\max}$ & Maximum number of rectification iterations & 10 \\
		$MSE_{\mathrm{stop}}$ & Error threshold for early stopping & 1e-5 \\
		Initial sampling range of constant for BFGS & $-$ & $\mathcal{U}(0,10)$ \\
		Number of BFGS restarts & $-$ & 10 \\ 
		\hline
	\end{tabularx}
\end{table}

\section{Main parameters for baselines}
\label{parameters_for_baselines}
The main hyperparameters of the baseline models used for comparison in our experiments, including ParFam, uDSR, TPSR, SR4MDL, and RILS-ROLS, are summarized in Table~\ref{tab:baselines_parameters}. For all other parameters, we use the default values provided in the public implementations.
\begin{table}[H]
	\centering
	\caption{The main parameters for baselines.}
	
	\renewcommand{\arraystretch}{1.3}
	\label{tab:baselines_parameters}
	\begin{tabular}{l|p{14cm}}
		\hline
		\textbf{Baselines} & \textbf{Hyperparameter sets} \\
		\hline
		\hline
		ParFam & $\{ 
		\textit{iterate}: \text{True},\ 
		\textit{time\_limit}: 1000,\ 
		\textit{time\_mode}: \text{auto},$ \newline
		$\textit{functions}: [\text{'sin'}, \text{'cos'}, \text{'exp'}, \text{'log'}, \text{'sqrt'}, \text{'tan'}, \text{'abs'}, \text{'asin'}, \text{'acos'}],\ 
		\textit{pass\_functions}: \text{True},$ \newline$ \}$\\
		\hline
		uDSR & 
		$\{ \textit{function\_set}: [\text{'add'}, \text{'sub'}, \text{'mul'}, \text{'div'}, \text{'sin'}, \text{'cos'}, \text{'exp'}, \text{'log'}, \text{'poly'}] $ \newline
		$\textit{batch\_size}: 500, \textit{n\_samples}: 200,000 \}$  \\ \hline
		TPSR &
		$\{ \textit{backbone\_model}: \text{e2e in benchmarking and NeSymReS in ablation studies }, \textit{search\_algorithm}: \text{MCTS}, \textit{tree\_search\_mode}: \text{best},$ \newline
		$\textit{width}: 3, \textit{horizon}: 200, \textit{rollout}: 3$\}
		 \\ \hline
		SR4MDL & 
		$\{ \textit{search\_method}: \text{MCTS}, \textit{n\_iter}: 10,000,$ \newline
		$\textit{tokenizer\_range}: [-100, 100], \textit{tokenizer\_precision}: 4,$ \newline
		$\textit{binary\_operators}: [\text{'mul'}, \text{'div'}, \text{'add'}, \text{'sub'}],$ \newline
		$\textit{unary\_operators}: [\text{'sqrt'}, \text{'cos'}, \text{'sin'}, \text{'pow2'}, \text{'pow3'}, \text{'exp'}, \text{'inv'},$ \newline
		$\text{'arcsin'}, \text{'arccos'}, \text{'log'}$], \newline
		$\textit{leaf}: [\text{'1'}, \text{'2'}, \pi] \}$ \\
		\hline
		
		RILS-ROLS & $\textit{max\_fit\_calls}: 10,000, \textit{max\_time}: 5000, \textit{complexity\_penalty}: 0.001, \textit{verbose}: \text{False} $ \\ \hline
		
	\end{tabular}
\end{table}

\section{Additional successful repair cases}
\label{app:successful_rectification_cases}
Table~\ref{tab:successful_rectification_cases} lists representative cases where the initial prediction from NeSymReS fails to recover the exact symbolic structure, but EditSR successfully rectifies it.

{\footnotesize
	\setlength{\tabcolsep}{3pt}
	\renewcommand{\arraystretch}{3}
	\begin{longtable}{@{}p{0.10\linewidth}p{0.20\linewidth}p{0.20\linewidth}p{0.08\linewidth}p{0.20\linewidth}p{0.08\linewidth}@{}}
		\caption{Representative cases of successful rectification.}\label{tab:successful_rectification_cases}\\
		\toprule
		\textbf{Name} & \textbf{True expression} & \textbf{NeSymReS} & \textbf{$R^2$} & \textbf{EditSR} & \textbf{$R^2$} \\
		\midrule
		\endfirsthead
		
		\multicolumn{6}{@{}l}{\tablename\ \thetable\ -- continued from previous page}\\
		\toprule
		\textbf{Name} & \textbf{True expression} & \textbf{NeSymReS} & \textbf{$R^2$} & \textbf{EditSR} & \textbf{$R^2$} \\
		\midrule
		\endhead
		
		\midrule
		\multicolumn{6}{r@{}}{Continued on next page}\\
		\endfoot
		
		\bottomrule
		\endlastfoot
		\texttt{I.10.7} & $\displaystyle \frac{x_{1}}{\sqrt{- \frac{x_{2}^{2}}{x_{3}^{2}} + 1}}$ & $\displaystyle \frac{0.60 x_{1} x_{2}^{2}}{x_{3}^{2}} + x_{1}$ & 0.9998 & $\displaystyle \frac{x_{1}}{\sqrt{- \frac{x_{2}^{2}}{x_{3}^{2}} + 1}}$ & 0.9999 \\
		\texttt{I.11.19} & $\displaystyle x_{1} x_{4} + x_{2} x_{5} + x_{3} x_{6}$ & $\displaystyle x_{1} x_{4} + x_{2} x_{5} + x_{3} + \frac{4.10 x_{6}}{x_{5}}$ & 0.7755 & $\displaystyle x_{1} x_{4} + x_{2} x_{5} + x_{3} x_{6}$ & 0.9999 \\
		\texttt{I.13.4} & $\displaystyle \frac{x_{1} \left(x_{2}^{2} + x_{3}^{2} + x_{4}^{2}\right)}{2}$ & $\displaystyle 0.46 x_{1}^{2} x_{4} + x_{1} x_{2} x_{3} + 6.83$ & 0.9092 & $\displaystyle x_{1} \left(x_{2} x_{4} + \frac{x_{3}^{2}}{2} + \frac{\left(- x_{2} + x_{4}\right)^{2}}{2}\right)$ & 0.9999 \\
		\texttt{I.15.1} & $\displaystyle \frac{x_{1} x_{2}}{\sqrt{- \frac{x_{2}^{2}}{x_{3}^{2}} + 1}}$ & $\displaystyle \frac{0.58 x_{1} x_{2}^{3}}{x_{3}^{2}} + x_{1} x_{2}$ & 0.9997 & $\displaystyle \frac{1.00 x_{1} x_{2}}{\sqrt{- \frac{x_{2}^{2}}{x_{3}^{2}} + 1.00}}$ & 0.9999 \\
		\texttt{I.16.6} & $\displaystyle \frac{x_{2} + x_{3}}{1 + \frac{x_{2} x_{3}}{x_{1}^{2}}}$ & $\displaystyle - \frac{0.04 x_{1}^{2} x_{2}^{2}}{x_{3}^{2} \left(0.71 - x_{2}\right)^{2}} + x_{1} - 0.04$ & 0.9138 & $\displaystyle x_{1} - \frac{\left(x_{1} - x_{2}\right) \left(x_{1} - x_{3}\right)}{x_{1} + \frac{x_{2} x_{3}}{x_{1}}}$ & 0.9999 \\
		
		\texttt{I.27.6} & $\displaystyle \frac{1}{\frac{x_{3}}{x_{2}} + \frac{1}{x_{1}}}$ & $\displaystyle 0.44 \sqrt{\frac{x_{1} x_{2}}{x_{3}}}$ & 0.9113 & $\displaystyle \frac{x_{2}}{x_{3} + \frac{x_{2}}{x_{1}}}$ & 0.9999 \\
	
		\texttt{I.30.3} & $\displaystyle \frac{x_{1} \sin^{2}{\left(\frac{x_{2} x_{3}}{2} \right)}}{\sin^{2}{\left(\frac{x_{2}}{2} \right)}}$ & $\displaystyle \frac{2.26 x_{1}}{x_{2}^{2} \left(- 0.20 x_{2} + 0.02 x_{3} + 1\right)}$ & 0.4402 & $\displaystyle \frac{x_{1} \left(1 - \cos{\left(x_{2} x_{3} \right)}\right)}{1 - \cos{\left(x_{2} \right)}}$ & 0.9999 \\
		\texttt{I.32.5} & $\displaystyle \frac{x_{1}^{2} x_{2}^{2}}{6 \pi x_{3} x_{4}^{3}}$ & $\displaystyle \frac{0.12 x_{1}^{2} x_{2}^{2}}{\pi x_{3} x_{4}^{2}}$ & 0.9303 & $\displaystyle \frac{0.17 x_{1}^{2} x_{2}^{2}}{\pi x_{3} x_{4}^{3}}$ & 0.9999 \\
		\texttt{I.48.2} & $\displaystyle \frac{x_{1} x_{3}^{2}}{\sqrt{- \frac{x_{2}^{2}}{x_{3}^{2}} + 1}}$ & $\displaystyle x_{1} \left(x_{2} + x_{3}^{2}\right)$ & 0.9999 & $\displaystyle \frac{x_{1} x_{3}^{2}}{\sqrt{- \frac{x_{2}^{2}}{x_{3}^{2}} + 1}}$ & 0.9999 \\
		\texttt{I.6.2} & $\displaystyle \frac{\sqrt{2} e^{- \frac{x_{2}^{2}}{2 x_{1}^{2}}}}{2 \sqrt{\pi} x_{1}}$ & $\displaystyle 0.19 \sin{\left(\frac{2.34}{x_{1} x_{2}} \right)}$ & 0.1130 & $\displaystyle \frac{2.78 \sqrt{2} e^{- \frac{x_{2}^{2}}{2 x_{1}^{2}}}}{\pi^{2} x_{1}}$ & 0.9999 \\
		\texttt{II.11.17} & $\displaystyle x_{1} \left(1 + \frac{x_{5} x_{6} \cos{\left(x_{4} \right)}}{x_{2} x_{3}}\right)$ & $\displaystyle x_{1} + \frac{2.00 x_{5} x_{6} \cos{\left(x_{4} \right)}}{x_{2} x_{3}}$ & 0.9030 & $\displaystyle x_{1} + \frac{x_{1} x_{5} x_{6} \cos{\left(x_{4} \right)}}{x_{2} x_{3}}$ & 0.9999 \\
		\texttt{II.13.23} & $\displaystyle \frac{x_{1}}{\sqrt{- \frac{x_{2}^{2}}{x_{3}^{2}} + 1}}$ & $\displaystyle \frac{0.60 x_{1} x_{2}^{2}}{x_{3}^{2}} + x_{1}$ & 0.9998 & $\displaystyle \frac{x_{1}}{\sqrt{- \frac{x_{2}^{2}}{x_{3}^{2}} + 1}}$ & 0.9999 \\
		\texttt{II.13.34} & $\displaystyle \frac{x_{1} x_{2}}{\sqrt{- \frac{x_{2}^{2}}{x_{3}^{2}} + 1}}$ & $\displaystyle \frac{1.11 x_{1} x_{2}^{2}}{x_{3}^{2}} + x_{1} x_{2}$ & 0.9987 & $\displaystyle \frac{x_{1} x_{2}}{\sqrt{- \frac{x_{2}^{2}}{x_{3}^{2}} + 1}}$ & 0.9999 \\
		\texttt{II.36.38} & $\displaystyle \frac{x_{1} x_{2}}{x_{3} x_{4}} + \frac{x_{1} x_{5} x_{8}}{x_{3} x_{4} x_{6} x_{7}^{2}}$ & $\displaystyle \frac{1.14 x_{1} x_{2}}{x_{3} x_{4}} + \frac{0.12 x_{1} x_{5} x_{6}}{x_{3} x_{4}}$ & 0.8020 & $\displaystyle \frac{x_{1} \left(x_{2} + \frac{x_{5} x_{8}}{x_{6} x_{7}^{2}}\right)}{x_{3} x_{4}}$ & 0.9999 \\
		\texttt{II.6.11} & $\displaystyle \frac{x_{2} \cos{\left(x_{3} \right)}}{4 \pi x_{1} x_{4}^{2}}$ & $\displaystyle \frac{0.16 x_{2} \cos{\left(x_{3} \right)}}{\pi x_{1} x_{4}}$ & 0.9035 & $\displaystyle \frac{x_{2} \cos{\left(x_{3} \right)}}{4 \pi x_{1} x_{4}^{2}}$ & 0.9999 \\
		\texttt{II.6.15a} & $\displaystyle \frac{3 x_{2} x_{6} \sqrt{x_{4}^{2} + x_{5}^{2}}}{4 \pi x_{1} x_{3}^{5}}$ & $\displaystyle \frac{8.10 x_{2} x_{6}}{\pi x_{1} x_{3}^{5} \left(x_{4} + x_{5}\right)}$ & 0.7868 & $\displaystyle \frac{0.75 x_{2} x_{6} \sqrt{x_{4}^{2} + x_{5}^{2}}}{\pi x_{1} x_{3}^{5}}$ & 0.9999 \\
		\texttt{II.6.15b} & $\displaystyle \frac{3 x_{2} \sin{\left(x_{3} \right)} \cos{\left(x_{3} \right)}}{4 \pi x_{1} x_{4}^{3}}$ & $\displaystyle \frac{0.12 x_{2} \cos{\left(x_{3} \right)}}{x_{1} x_{4}^{3}}$ & 0.8019 & $\displaystyle \frac{0.24 x_{2} \sin{\left(x_{3} \right)} \cos{\left(x_{3} \right)}}{x_{1} x_{4}^{3}}$ & 0.9999 \\
		\texttt{III.4.32} & $\displaystyle \frac{1}{e^{\frac{x_{1} x_{2}}{2 \pi x_{3} x_{4}}} - 1}$ & $\displaystyle \frac{6.12 x_{3} x_{4}}{x_{1} x_{2}}$ & 0.9987 & $\displaystyle \frac{1}{e^{\frac{x_{1} x_{2}}{2 \pi x_{3} x_{4}}} - 1}$ & 0.9999 \\
		\texttt{III.8.54} & $\displaystyle \sin^{2}{\left(\frac{2 \pi x_{1} x_{2}}{x_{3}} \right)}$ & $\displaystyle \sin^{2}{\left(2 \pi x_{1} x_{2} \left(x_{3} + 10.49\right) \right)}$ & -0.9971 & $\displaystyle \sin^{2}{\left(\frac{2 \pi x_{1} x_{2}}{x_{3}} \right)}$ & 0.9999 \\
		\texttt{test\_18} & $\displaystyle \frac{\frac{3 x_{2} x_{5}^{2}}{x_{3}^{2}} + 3 x_{4}^{2}}{8 \pi x_{1}}$ & $\displaystyle \frac{x_{2} x_{4}}{x_{1} x_{3} x_{5}}$ & -0.0746 & $\displaystyle \frac{0.12 x_{2} x_{5}^{2}}{x_{1} x_{3}^{2}} + \frac{0.12 x_{4}^{2}}{x_{1}}$ & 0.9999 \\
		\texttt{test\_5} & $\displaystyle \frac{2 \pi x_{1}^{\frac{3}{2}}}{\sqrt{x_{2} \left(x_{3} + x_{4}\right)}}$ & $\displaystyle x_{1} \left(3.83 - \frac{0.46 x_{4}^{2} \left(1 - 0.87 x_{2}\right)^{2}}{x_{2} x_{3}}\right)$ & 0.7968 & $\displaystyle \frac{2 \pi x_{1}^{\frac{3}{2}}}{\sqrt{x_{2} \left(x_{3} + x_{4}\right)}}$ & 0.9999 \\
	
		\texttt{test\_9} & $\displaystyle - \frac{32 x_{1}^{4} x_{3}^{2} x_{4}^{2} \left(x_{3} + x_{4}\right)}{5 x_{2}^{5} x_{5}^{5}}$ & $\displaystyle - \frac{4.63 x_{1} x_{3}^{2} x_{4}^{2}}{x_{5}^{2} \left(0.87 - x_{2}\right)^{2}}$ & 0.4848 & $\displaystyle - \frac{6.40 x_{1}^{4} x_{3}^{2} x_{4}^{2} \left(x_{3} + x_{4}\right)}{x_{2}^{5} x_{5}^{5}}$ & 0.9999 \\
	\end{longtable}
}

\end{document}